\documentclass[10pt,twocolumn,letterpaper]{article}

\usepackage{cvpr}
\usepackage{times}
\usepackage{epsfig}
\usepackage{graphicx}
\usepackage{subcaption}
\usepackage{amsmath}
\usepackage{amssymb}
\usepackage[font={small}]{caption}
\usepackage{multirow}
\usepackage{diagbox}
\usepackage{ifthen}
\usepackage{bm}
\usepackage{bbding} 
\usepackage{lipsum}
\usepackage{url}
\usepackage{booktabs}

\newcommand{\redbf}[1]{{\textbf{\color{red}{#1}}}} 
\newcommand{\blueud}[1]{{\underline{\color{blue}{#1}}}} 
\usepackage{pifont}
\usepackage[hang,flushmargin]{footmisc} 

\usepackage[pagebackref=true,breaklinks=true,letterpaper=true,colorlinks,bookmarks=false]{hyperref}
\cvprfinalcopy 
\usepackage{cleveref}

\def\cvprPaperID{86} 

\ifcvprfinal

\begin{document}

\title{\vspace{0.2cm}EDPN: Enhanced Deep Pyramid Network for Blurry Image Restoration}

\author{
	Ruikang Xu$^*$ \quad
    Zeyu Xiao\thanks{These authors contribute equally to this work.} \quad
    Jie Huang   \quad 
    Yueyi Zhang \quad 
    Zhiwei Xiong\thanks{Correspondence should be addressed to zwxiong@ustc.edu.cn} \\
	University of Science and Technology of China \\
}

 \maketitle

\begin{abstract}
Image deblurring has seen a great improvement with the development of deep neural networks. In practice, however, blurry images often suffer from additional degradations such as downscaling and compression.
To address these challenges, we propose an \textbf{E}nhanced \textbf{D}eep \textbf{P}yramid \textbf{N}etwork (EDPN) for blurry image restoration from multiple degradations, by fully exploiting the self- and cross-scale similarities in the degraded image.
Specifically, we design two pyramid-based modules, i.e., the pyramid progressive transfer (PPT) module and the pyramid self-attention (PSA) module, as the main components of the proposed network.
By taking several replicated blurry images as inputs, the PPT module transfers both self- and cross-scale similarity information from the same degraded image in a progressive manner.
Then, the PSA module fuses the above transferred features for subsequent restoration using self- and spatial-attention mechanisms.
Experimental results demonstrate that our method significantly outperforms existing solutions for blurry image super-resolution and blurry image deblocking. 
In the NTIRE 2021 Image Deblurring Challenge, EDPN achieves the best PSNR/SSIM/LPIPS scores in Track 1 (Low Resolution) and the best SSIM/LPIPS scores in Track 2 (JPEG Artifacts).
The implementation code is available at \url{https://github.com/zeyuxiao1997/EDPN}.

\end{abstract}
\vspace{-0.5cm}

\section{Introduction}
Image deblurring has long been an important task in computer vision and image processing.
Blurry images may be caused by camera shake~\cite{su2017deep,pan2019single}, object motion~\cite{zhang2018dynamic,ramakrishnan2017deep,wieschollek2017learning} or out-of-focus~\cite{abuolaim2020defocus,lee2019deep,abuolaim2021NTIRE}, and the goal of image deblurring is to recover a sharp latent image with necessary edge structures and details. Image deblurring is a highly ill-posed task especially due to the difficulties in estimating the spatially varying blur kernel with limited information from a single observation. 

Early Bayesian-based iterative deblurring methods include the Wiener filter~\cite{weiner1949smoothing} and the
Richardson-Lucy algorithm~\cite{richardson1972bayesian}.
Later works commonly rely on developing effective image priors~\cite{krishnan2009fast,schmidt2014shrinkage,xu2010two,zoran2011learning} or sophisticated data terms~\cite{dong2018learning}.
More recently, convolutional neural networks (CNNs) have been exploited for image deblurring and produce promising results.
For example, Nah~\emph{et al.}~\cite{nah2017deep} propose a multi-scale loss function to implement a coarse-to-fine processing pipeline. Tao~\emph{et al.} ~\cite{tao2017detail} and Gao~\emph{et al.}~\cite{gao2019dynamic} improve this work by using shared network parameters at different scales, achieving state-of-the-art performance.

Despite of the encouraging performance achieved by CNN-based methods for image deblurring, they fail to reconstruct sharp results from the blurry images with multiple degradations.
In practice, however, the blurry images often suffer from additional degradations.
For example, to save the storage and transmission bandwidth, the raw images are generally downscaled and/or compressed, resulting in low resolution and/or compression artifacts when the images are received by terminal users.
Therefore, a more general blurry image restoration task should not only consider single blur degradation, but also cover more complex degradations, \eg, blurry image super-resolution (BISR) and blurry image deblocking (BID).

To address the general blurry image restoration task (\ie, BISR and BID), a straightforward strategy is to cascade deblurring and super-resolution/deblocking techniques, or vice versa.
However, there are several issues with such approaches.
First, a simple concatenation of two models is a sub-optimal solution due to error accumulation, \ie, the estimated error of the first model will be propagated and magnified in the second model.
Second, the two-stage network does not fully exploit the correlation between the two tasks. 
Third, the cascading approach is specifically designed for fixed tasks, which cannot be easily deployed in more general scenarios.
Several recent methods~\cite{quan2020collaborative,xu2017learning,yu2018super,zhang2018gated} jointly solve the image deblurring and super-resolution problems using end-to-end deep neural networks. However, these methods either focus on domain-specific applications, \eg, face and text~\cite{xu2017learning,yu2018super} images, or address the uniform Gaussian blur only~\cite{zhang2018deep}.

In this paper, we propose an \textbf{E}nhanced \textbf{D}eep \textbf{P}yramid \textbf{N}etwork (EDPN), which is extensible to various blurry image restoration tasks, including BISR and BID. 
The inputs of EDPN are several replicated blurry images, which aims to fully exploit the self-similarity contained in the degraded image.
The cores of EDPN are (1) an information transfer module named as pyramid progressive transfer (PPT) module, and (2) a feature fusion module named as pyramid self-attention (PSA) module.
Specifically, the PPT module is designed to transfer the cross-scale similarity information from the same degraded image at the feature level with a pyramid structure, which performs the deformable convolution and generates attention masks to transfer the self-similarity information in a progressive manner.
The PSA module is designed to aggregate information across the transferred features with a pyramid structure, which adopts self- and spatial-attention mechanisms to weight the multiple features.

Our contributions can be summarized as follows:
\begin{enumerate}
    \item We propose a blurry image restoration network named EDPN, which can generate sharp results from blurry images with multiple degradations.
    \item We design two core components, \ie, the PPT module and the PSA module, for fully exploiting the self- and cross-scale similarities of the same degraded image.
    \item 
    Our method significantly outperforms existing solutions for blurry image super-resolution and blurry image deblocking.
    In the NTIRE 2021 Image Deblurring Challenge, EDPN achieves the best PSNR/SSIM/LPIPS scores in Track 1 (Low Resolution) and the best SSIM/LPIPS scores in Track 2 (JPEG Artifacts).
\end{enumerate}

\section{Related Work}

\textbf{Image deblurring.}
Image deblurring is a highly ill-posed problem which aims at generating a sharp image from a blurry observation. Various natural images and kernel priors have been developed to regularize the solution space of the latent sharp image, including heavy-tailed gradient prior~\cite{shan2008high}, sparse kernel prior~\cite{fergus2006removing}, $l_0$ gradient prior~\cite{xu2013unnatural}, normalized sparsity prior~\cite{krishnan2011blind} and dark channels~\cite{pan2016blind}.
Recently, several CNN-based methods have been proposed for image deblurring.
For example, Sun~\emph{et al.}~\cite{sun2015learning} propose a CNN-based model to estimate a kernel and remove non-uniform motion blur. Chakrabarti~\cite{chakrabarti2016neural} uses a network to compute estimations of sharp images that are blurred by an unknown motion kernel. Nah~\emph{et al.}~\cite{nah2017deep} propose a multi-scale loss function to apply a coarse-to-fine strategy. Kupyn~\emph{et al.} propose DeblurGAN~\cite{kupyn2018deblurgan} and DeblurGAN-v2~\cite{kupyn2019deblurgan} to remove blur based on adversarial learning.

\textbf{Image super-resolution.}
A plenty of works have been proposed to solve image super-resolution (SR), including interpolation-based~\cite{zhang2006edge}, model-based~\cite{gu2015convolutional} and learning-based methods~\cite{xiong2009image,xiong2013example,dong2014learning,kim2016accurate,lim2017enhanced,zhang2018image,chen2019camera}.
Traditional methods are usually limited in representing the complex local-image structures, while recently developed deep CNNs have shown great advantages in image structure representation and consequently boost the SR performance~\cite{dong2014learning,dai2019second,lim2017enhanced}.
For example, Kim~\emph{et al.}~\cite{kim2016accurate} employ the residual learning strategy to design the VDSR model with 20 convolution layers.
By introducing channel attention mechanism, Zhang~\emph{et al.}~\cite{zhang2018image} propose RCAN which improves the SR performance a lot. Dai~\emph{et al.}~\cite{dai2019second} propose a second-order attention network for more powerful feature expression and feature correlation learning.

\begin{figure*}[t]
  \centering
  \includegraphics[width=1\linewidth]{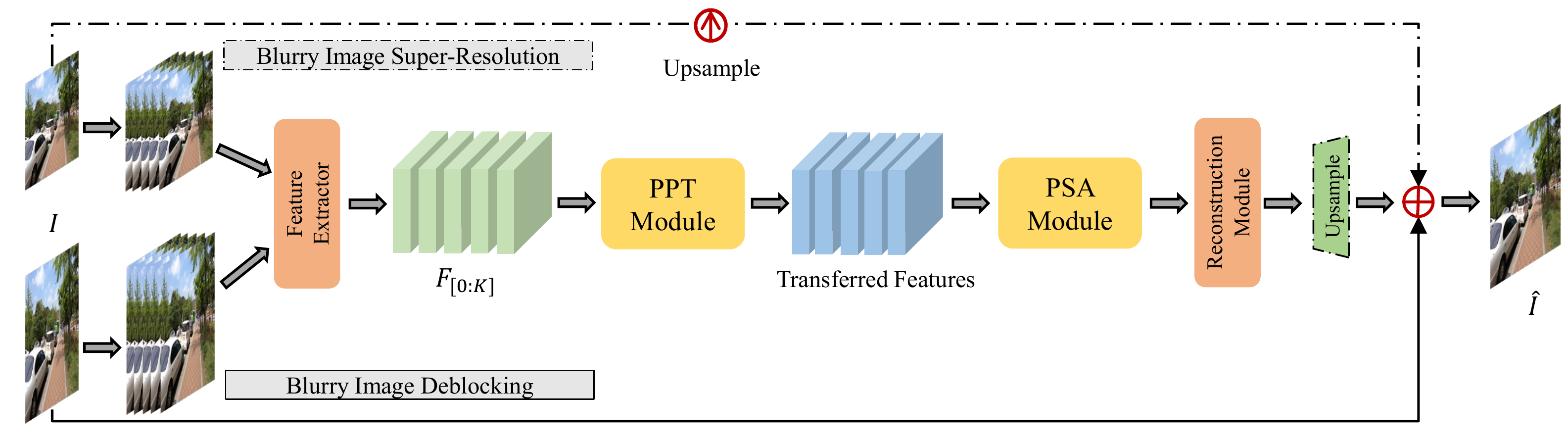}
  \vspace{-0.6cm}
  \caption{The EDPN network is suitable for various image deblurring tasks. Here, we focus on tasks of blurry image super-resolution and blurry image deblocking for example. The given blurry image is first replicated for several times, and then sent to the following components. We use five replicated images as an illustrative example.}
  \label{fig:overview}
  \vspace{-0.3cm}
\end{figure*}

\textbf{Image deblocking.}
Traditional JPEG artifacts removal methods pay attention to filter design.
For example, Foi~\emph{et al.}~\cite{foi2007pointwise} propose the shape-adaptive DCT-based filter for image denoising and de-blocking.
The method in~\cite{chang2013reducing} utilizes sparse coding to restore compressed images. 
Others treat JPEG artifacts removal as an ill-posed inverse problem and solve it by using regression trees~\cite{jancsary2012loss} and non-local self-similarity property~\cite{li2017iterative}. 
CNN-based methods learn to minimize the reconstruction error with respect to ground truth reference images, and operate in the pixel domain~\cite{fu2019jpeg,dong2015compression,svoboda2016compression,wang2020jpeg}, the DCT domain~\cite{yoo2018image}, or both domains~\cite{guo2017one,guo2016building,zhang2018dmcnn}.
For example, Fan~\emph{et al.}~\cite{fan2018decouple} propose a decoupled learning framework to combine different parameterized operators. Fu~\emph{et al.}~\cite{fu2019jpeg} introduce a more compact and explainable deep sparse coding architecture to generate high-quality deblocking results.

\textbf{General blurry image restoration.}
A typical blurry image restoration task is to super-resolve a low-resolution (LR) image and deblur a blurry image jointly~\cite{zhang2018gated,xu2017learning,zhang2018deep,zhang2020joint,yang2020deblurring}.
This joint problem is more challenging than the individual problems. 
Xu~\emph{et al.}~\cite{xu2017learning} train a generative adversarial network to super-resolve blurry face and text images.
Zhang~\emph{et al.}~\cite{zhang2018deep} propose a deep encoder-decoder network for joint image deblurring and super-resolution. However, they focus on LR images degraded by the uniform Gaussian blur.
Zhang~\emph{et al.}~\cite{zhang2018gated} propose a dual-branch network to extract features for deblurring and super-resolution and learn a gate module to adaptively fuse the features for image restoration.

\textbf{Attention mechanism.}
The attention mechanism in deep learning, which mimics the human visual attention mechanism, is originally developed in a non-local manner.
For example, the matrix multiplication in self-attention draws global dependencies of each word in a sentence~\cite{vaswani2017attention} or each pixel in an image~\cite{wang2018non}.
The squeeze-and-excitation network squeezes global spatial information into a channel descriptor to capture channel-wise dependencies~\cite{hu2018squeeze}.
To alleviate the problems arising from scale variation and small objects, Dai~\emph{et al.}~\cite{dai2021attentional} propose the multi-scale channel attention module for aggregating contextual information from different receptive fields, which can simultaneously aggregate local and global feature contexts inside the channel attention mechanism.

\section{Network Architecture}
\subsection{Overview}\label{subsec:Overview}
Given a blurry image $I$, our method aims to reconstruct a high-quality image $\hat{I}$, which should be close to the ground truth ${I}^{GT}$.
As shown in Figure~\ref{fig:overview}, our EDPN mainly consists of four parts: the feature extractor, the PPT module, the PSA module and the reconstruction module. 

Take the BISR task as an example, we first replicate the given blurry image $I$ for $K$ times as the inputs of our EDPN, which can better exploit the self-similarity in the degraded image.
Then, we extract the features from the inputs by the feature extractor.
The feature extractor consists of 18 residual blocks. 
The extracted features are denoted as $F_{[0:K]}$, which will be utilized for subsequent operations.

After feature extraction, the features are fed into the PPT module to transfer the self- and cross-scale similarity information in a progressive manner.
Then, the PSA module fuses the similarity information of the transferred features $\hat{F}_{[0:K]}$ and conducts feature aggregation.
The details of these two modules are described in Section~\ref{subsec:PPT} and Section~\ref{subsec:PSA}.
After that, the fused features are fed into the reconstruction module with an upsampling operation. 
Finally, the restored image is obtained by adding the predicted image residual to a directly upsampled image. 
The reconstruction module is composed of 120 multi-scale residual channel-attention blocks~\cite{dai2021attentional}.
For tasks with high spatial resolution inputs such as BID, the upsampling layer at the end is not necessary.

\subsection{Pyramid Progressive Transfer Module}\label{subsec:PPT}
\begin{figure}[!t]
  \centering
  \includegraphics[width=1\linewidth]{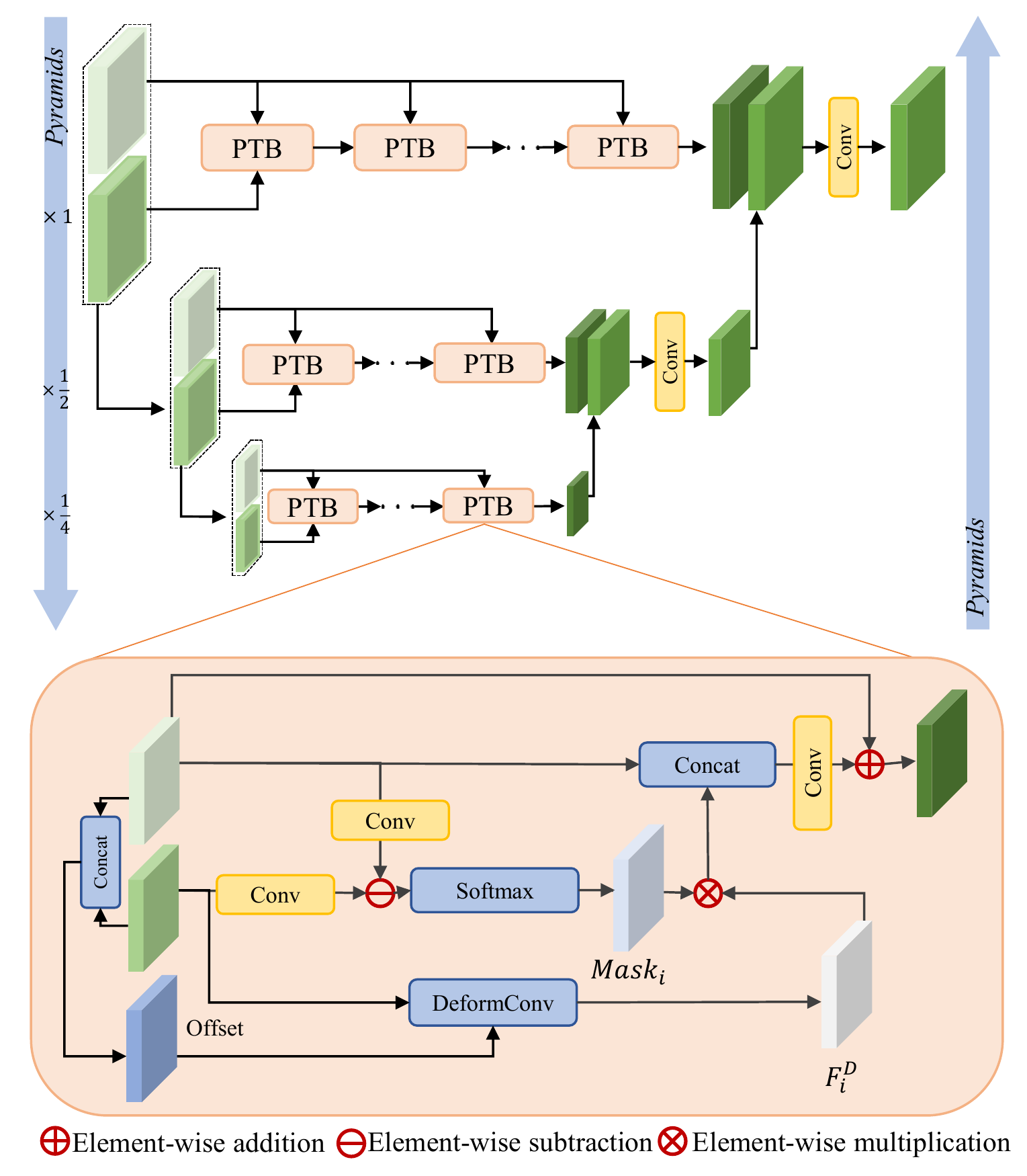}
  \vspace{-0.6cm}
  \caption{Structure of the PPT module.} 
  \label{fig:PPT}
  \vspace{-0.2cm}
\end{figure}

The detailed structure of the proposed PPT module is shown in Figure~\ref{fig:PPT}. 
The inputs of the PPT module are the features $F_{[0:K]}$. 
Given $K+1$ features, the PPT module also needs to be executed for $K+1$ times.

Inside the PPT module, we adopt the pyramid and progressive structure to learn the self- and cross-scale similarities.
For the pyramid structure, strided convolution layers are utilized to downscale the features from the upper level by a factor of 2 for obtaining features at the current level.
Assuming that the number of pyramid levels constructed in the PPT module is $M$, for each level, there are $N$ progressive transfer blocks (PTBs) to extract the self-similarity progressively. 
Then at the $m$-th level, the inputs of the $n$-th PTB are the first feature $F_{0}^{m}$ and the output of the previous block $(F_{i, PTB}^{n-1})^{m}  (i\in {[0,K]})$. 
It should be noted that the inputs of the first PTB are $F_{0}^{m}$ and $F_{i}^{m}$.
Inspired by TDAN~\cite{tian2020tdan} and EDVR~\cite{wang2019edvr}, we apply the deformable convolution~\cite{dai2017deformable} in the PTB.
This process can be denoted as
\begin{equation}
(F^D_i)^{m, n} = \mathcal{F}_{Dconv}(F_{0}^{m},(F_{i, PTB}^{n-1})^{m})),
\end{equation}
where $\mathcal{F}_{Dconv}(\cdot)$ stands for the deformable convolution and $(F^D_i)^{m, n}$ stands for the output of the deformable convolution of the $n$-th block at the $m$-th level. 
The learned offsets of the deformable convolution are predicted from the inputs, which is formulated as
\begin{equation}
(\Delta P_i)^{m, n} = \mathcal{F}_{C}(F_{0}^{m} \parallel (F_{i, PTB}^{n-1})^{m}),
\end{equation}
where $(\Delta P_i)^{m, n}$ stands for the learned offset of the $n$-th block at the $m$-th level, $\parallel$ stands for the channel-wise concatenation and $\mathcal{F}_{C}(\cdot)$ stands for the convolution operation.

Then, we generate the feature-level mask of the $n$-th block at the $m$-th level $(Mask_i)^{m,n}$, which forces the PTB to focus on the most correlated information of features.
Mathematically, the mask is calculated as
\begin{equation}\label{equ:motion atten}
(Mask_i)^{m,n} = {Softmax}(\mathcal{F}_{C}(F_{0}^{m})- \mathcal{F}_{C}((F_{i, PTB}^{n-1})^{m})).
\end{equation} 
The motion attention mask is further multiplied with the output of the deformable convolution. 

After a convolution layer, the generated feature is treated as the residual information of this block. 
The output feature of the $n$-th PTB at the $m$-th level is obtained by adding the residual information to the first feature, which can be formulated as
\begin{equation}\label{equ:spa}
\!(F_{i, PTB}^{n})^{m}\! =\! F_{0}^{m}\! +\! \mathcal{F}_{C}(F_{0}^{m} \! \parallel \!(Mask_i)^{\!m,n} \!\otimes (F_i^D)^{m, n}))\!,
\end{equation}
where $\otimes$ represents the element-wise multiplication.
Finally, the output feature of the PPT module at the $m$-th level $(F_i^{PPT})^{m}$ can be depicted as 
\begin{equation}
\left(F_{i}^{P P T}\right)^{m}=\mathcal{F}_{C}\left(U p\left(\left(F_{i}^{P P T}\right)^{m+1}\right)^{\uparrow s} \|\left(F_{i, P T B}^{N}\right)^{m}\right),
\end{equation} 
where $(F_{i, PTB}^{N})^{m}$ represents the feature generated after $N$ PTBs at the $m$-th level.
$U p(\cdot)^{\uparrow s}$ refers to upscaling by a factor of $s$, which is implemented by bilinear interpolation. 
We construct our PPT module with 3-level pyramid structure, \emph{i.e.}, $M = 3$.
The PPT module can transfer self- and cross-scale similarities in such a progressive and coarse-to-fine manner.
We demonstrate the effectiveness of the PPT module and analyze the relation between the performance and the number of PTBs in Section~\ref{sec:ablation}.

\begin{figure}[!t]
  \centering
  \includegraphics[width=1\linewidth]{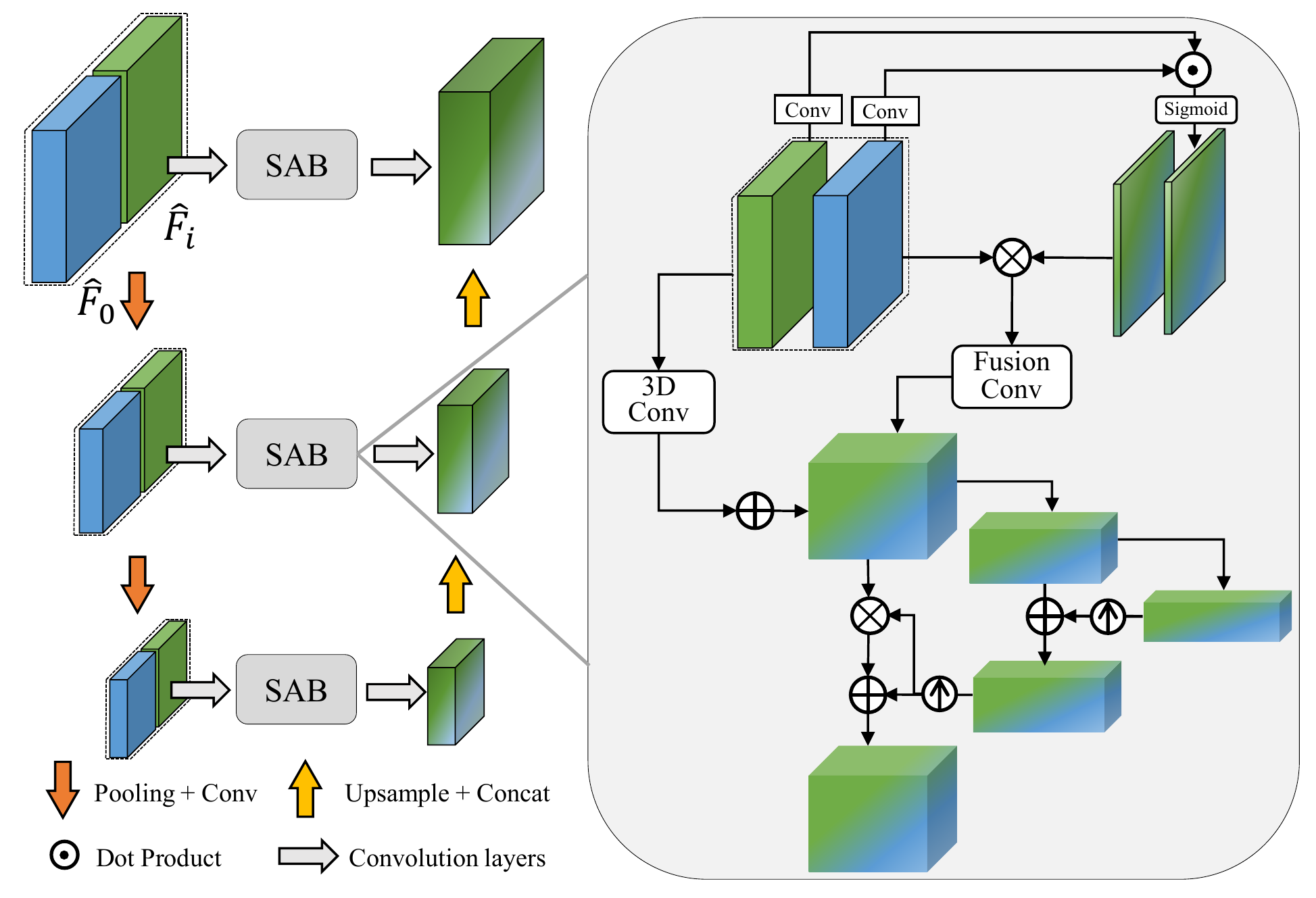}
  \vspace{-0.6cm}
  \caption{Structure of the PSA module.}
  \vspace{-0.2cm}
  \label{fig:PTSA}
\end{figure}

\subsection{Pyramid Self-Attention Module}\label{subsec:PSA}
After the PPT module, the features with self- and cross-scale similarities for fusion and reconstruction have been extracted and transferred.
Inspired by~\cite{wang2019edvr}, we propose the PSA module to assign pixel-level aggregation weights with a pyramid structure.
In addition, we adopt the 3D convolution operation to fuse the information of all features effectively as shown in Figure~\ref{fig:PTSA}.

Our PSA module also adopts the pyramid processing.
At first, we define the output feature of the self-attention block (SAB) at the $l$-th level as $F_{sa}^l$.
Then, we use strided convolution layers to downscale the features at the $l$-th pyramid level by a factor of 2, obtaining $L$-level pyramid of feature representation.
At the $l$-th level, we generate the attention maps to compute similarity in an embedding space.
For each feature, the similarity map can be calculated as
\begin{equation}\label{equ:tam}
\Theta_{i}^{l} = Sigmoid(\mathcal{F}_{C}(\hat{F}_{0}^{l})^{T} \odot \mathcal{F}_{C}(\hat{F}_{i}^{l})),
\end{equation} 
where $\odot$ stands for the dot product operation.
$\hat{F}_{i}^{l}$ stands for the transferred features at the $l$-th level. 
Specially, similarity maps are spatial-specific for each spatial location, \emph{i.e.}, the spatial size of $\Theta_{i}^{l}$ is the same as that of $\hat{F}_{i}^{l}$.
The similarity maps are then multiplied in a pixel-wise manner to the original transferred features, and an extra fusion convolution layer is adopted to aggregate these attention-modulated features $\tilde{F}_{i}^{l}$, denoted as
\begin{equation}\label{equ:tam_out}
\begin{aligned}
\tilde{F}_{i}^{l} &= \Theta_{i}^{l} \otimes \hat{F}_{i}^{l}, \\ 
F_{fusion}^{l} &= \mathcal{F}_{C}(\tilde{F}_{[0:K]}^{l}).
\end{aligned}
\end{equation}
Then, we add the original transferred features after the 3D convolution operation to the fused features.
Meanwhile, the spatial attention masks are then computed from the fused features with the pyramid structure. 
Following~\cite{wang2019edvr}, the fused features are modulated by the masks through element-wise multiplication and addition to generate the output $F_{sa}^l$ at the $l$-th level, denoted as
\begin{equation}\label{equ:tam}
F_{s a}^{l}=\mathcal{F}_{C}\left(F_{s a}^{l} \| U p\left(F_{s a}^{l+1}\right)^{\uparrow 2}\right).
\end{equation} 
Here, we use a 3-level pyramid ($L=3$). 
To reduce computational cost, we do not increase channel numbers as spatial sizes decrease.
The PSA module in such a coarse-to-fine manner improves the effectiveness of information aggregation and we demonstrate the effectiveness of the PSA module in Section~\ref{sec:ablation}.

\section{Experiments}\label{sec:ablation}
\subsection{Experimental Settings}\label{subsec:dataset}
\textbf{Dataset.} The experiments are conducted strictly following the instructions of the NTIRE 2021 Image Deblurring Challenge~\cite{Nah_2021_CVPR_Workshops_Deblur}.
There are two tracks in this challenge.
Track 1 requires to restore and upscale a blurry image by a factor of 4.
Track 2 requires to restore a blurry image with JPEG artifacts.
The dataset used for these tracks is REDS~\cite{nah2019ntire}, a real-world high-quality video dataset originally collected for video super-resolution~\cite{caballero2017real,Son_2021_CVPR_Workshops_VSR,kappeler2016video,tian2020tdan,wang2019edvr,xiao2020spacetime,xiao2021spacetime} and video deblurring~\cite{su2017deep,xue2019video}, which consists of 240 scenes for training, and 30 scenes for validation and testing.

\textbf{Loss functions.} 
To optimize EDPN, we adopt the Charbonnier loss~\cite{wang2019edvr}  defined as
\begin{equation}
\mathcal{L}_{Charb} =\sqrt{\left\|{I}^{GT}-\hat{I}\right\|^{2}+\varepsilon^{2}},
\end{equation} 
where $\varepsilon$ is set to $1 \times 10^{-3}$.
We also adpot the SSIM loss~\cite{hore2010image} defined as
\begin{equation}
\mathcal{L}_{SSIM} = 1-SSIM({I}^{GT}, \hat{I}).
\end{equation} 
The complete loss function for training EDPN is
\begin{equation}\label{lossss}
\mathcal{L} = \mathcal{L}_{Charb} + \lambda \mathcal{L}_{SSIM},
\end{equation} 
where $\lambda$ is the weighting factor.

\textbf{Training settings.}
The channel size in the feature extractor and the reconstruction module is set to 64.
We use RGB patches of size $64 \times 64$ and $160  \times 160$ as inputs for BISR and BID tasks, respectively.
The network takes five replicated images (\ie, $K=4$) as inputs.
We augment the training data with random horizontal flips and rotations.

\textbf{Implementation details.}
We utilize the Adam optimizer with parameters $\beta_1=0.9$ and $\beta_2=0.999$. The training procedure follows the mini-batch strategy and the batch size is 4.
The learning rate is initially set to $ 1 \times 10^{-4} $ and is later down-scaled by a factor of 0.8 after every 100,000 iterations till 1,000,000 iterations.
$\lambda$ is set to 0.1.
All the networks in the experiments are implemented using PyTorch 1.1 and trained with NVIDIA GeForce GTX1080Ti GPUs.

\subsection{Ablation Study}\label{sec:ablation}
In this subsection, we investigate the necessity of our proposed modules, the suitable design choices (\ie, the number of the PTBs and input images) and ensemble schemes. 
All the ablation experiments are conducted for the BISR task.

\begin{table}[!t]
  \small
  \fontsize{8}{9.6}\selectfont
  \vspace{0.1cm}
  \caption{Ablation on the PPT and PSA modules.}
  \label{table:module_ablation}
  \vspace{-0.7cm}
  \begin{center}
    \tabcolsep=0.32cm
    \resizebox{0.99\linewidth}{!}{
      \begin{tabular}{c|c|cc|cc}
        \bottomrule[1.2pt]
        \multicolumn{1}{c|}{\multirow{2}{*}{PPT}} &
  \multicolumn{1}{c|}{\multirow{2}{*}{PSA}} &
  \multicolumn{2}{c|}{RGB Channel} &
  \multicolumn{2}{c}{Y Channel} \\ \cline{3-6} 
\multicolumn{1}{c|}{} &
  \multicolumn{1}{c|}{} &
  \multicolumn{1}{c}{PSNR} &
  \multicolumn{1}{c|}{SSIM} &
  \multicolumn{1}{c}{PSNR} &
  SSIM \\ \hline
    \XSolidBrush & \XSolidBrush & 27.43 & 0.7839 &  28.80 & 0.8045 \\
    \Checkmark   & \XSolidBrush & 27.84 & 0.8010 &  29.23 & 0.8208 \\
    \XSolidBrush & \Checkmark  & 27.78 & 0.7997 &  29.18 & 0.8196 \\
    \Checkmark   & \Checkmark   & \textbf{28.01} & \textbf{0.8091} &  \textbf{29.39} & \textbf{0.8282}\\
        \toprule[1.2pt]
        \end{tabular}%
  }
\end{center}
\vspace{-0.5cm}
\end{table}%

\begin{table}[!t]
  \vspace{0.1cm}
  \caption{Ablation on the number of PTBs.}
  \label{table:ablation_block}
  \vspace{-0.7cm}
  \begin{center}
    \tabcolsep=0.32cm
    \resizebox{1\linewidth}{!}{
      \begin{tabular}{c|cc|cc}
        \bottomrule[1.2pt]
\multicolumn{1}{c|}{\multirow{2}{*}{Num. of PTB}} & \multicolumn{2}{c|}{RGB Channel}                        & \multicolumn{2}{c}{Y Channel} \\ \cline{2-5} 
\multicolumn{1}{c|}{}                      & \multicolumn{1}{c}{PSNR} & \multicolumn{1}{c|}{SSIM} & \multicolumn{1}{c}{PSNR}   & SSIM  \\ \hline 
        $N=1$&    27.91 & 0.8070 & 29.31 & 0.8264 \\
        $N=2$&    27.92 & 0.8071 & 29.32 & 0.8264 \\
        $N=3$&    28.01 & 0.8091 & 29.39 & 0.8282\\ 
        $N=4$&    28.01 & 0.8097 & 29.40 & 0.8288\\
        $N=5$&    \textbf{28.02} & \textbf{0.8102} & \textbf{29.41} & \textbf{0.8296}\\
        \toprule[1.2pt]
        \end{tabular}%
  }
\end{center}
\vspace{-0.5cm}
\end{table}%

\begin{table}[!t]
  \small
  \fontsize{8}{9.6}\selectfont
  \vspace{0.1cm}
  \caption{Ablation on the number of input images.}
  \label{table: frame_ablation}
  \vspace{-0.7cm}
  \begin{center}
    \tabcolsep=0.32cm
    \resizebox{0.99\linewidth}{!}{
      \begin{tabular}{c|cc|cc}
        \bottomrule[1.2pt]
       \multicolumn{1}{c|}{\multirow{2}{*}{Num. of Input}} & \multicolumn{2}{c|}{RGB Channel}  & \multicolumn{2}{c}{Y Channel} \\ \cline{2-5} 
\multicolumn{1}{c|}{}                      & \multicolumn{1}{c}{PSNR} & \multicolumn{1}{c|}{SSIM} & \multicolumn{1}{c}{PSNR}   & SSIM  \\ \hline 
        \# 1 ($K = 0$) &    27.89 & 0.8012   &  29.21 & 0.8244  \\
        \# 3 ($K = 2$) &    27.95 & \textbf{0.8102}   &  29.35 & \textbf{0.8292} \\
        \# 5 ($K = 4$) &    \textbf{28.01} & {0.8091}   &  \textbf{29.39} & {0.8282} \\
        \# 7 ($K = 6$) &    27.91 & 0.8069   &  29.31 & 0.8263 \\
        \toprule[1.2pt]
        \end{tabular}%
  }
\end{center}
\vspace{-0.5cm}
\end{table}%

\begin{table}[!t]
  \small
  \fontsize{8}{9.6}\selectfont
  \vspace{0.1cm}
  \caption{Ablation on the loss functions.}
  \label{table:loss}
  \vspace{-0.7cm}
  \begin{center}
    \tabcolsep=0.3cm
    \resizebox{0.99\linewidth}{!}{
      \begin{tabular}{c|c|cc|cc}
        \bottomrule[1.2pt]
        \multicolumn{1}{c|}{\multirow{2}{*}{$\mathcal{L}_{Charb}$}} &
  \multicolumn{1}{c|}{\multirow{2}{*}{$\mathcal{L}_{SSIM}$}} &
  \multicolumn{2}{c|}{RGB Channel} &
  \multicolumn{2}{c}{Y Channel} \\ \cline{3-6} 
\multicolumn{1}{c|}{} &
  \multicolumn{1}{c|}{} &
  \multicolumn{1}{c}{PSNR} &
  \multicolumn{1}{c|}{SSIM} &
  \multicolumn{1}{c}{PSNR} &
  SSIM \\ \hline
    \Checkmark & \XSolidBrush   & \textbf{28.04} & 0.8070 &  \textbf{29.44} & 0.8247\\
    \Checkmark   & \Checkmark   & 28.01 & \textbf{0.8091} &  29.39 & \textbf{0.8282}\\
        \toprule[1.2pt]
        \end{tabular}%
  }
\end{center}
\vspace{-0.5cm}
\end{table}%

\begin{table}[!t]
  \small
  \fontsize{8}{9.6}\selectfont
  \vspace{0.1cm}
  \caption{Ablation on the ensemble schemes.}
  \label{table: ensemble_ablation}
  \vspace{-0.7cm}
  \begin{center}
    \tabcolsep=0.25cm
    \resizebox{0.99\linewidth}{!}{
    \begin{tabular}{c|cc|cc}
        \bottomrule[1.2pt]
       \multicolumn{1}{c|}{\multirow{2}{*}{Ensemble scheme}} & \multicolumn{2}{c|}{RGB Channel}                        & \multicolumn{2}{c}{Y Channel} \\ \cline{2-5} 
\multicolumn{1}{c|}{}                      & \multicolumn{1}{c}{PSNR} & \multicolumn{1}{c|}{SSIM} & \multicolumn{1}{c}{PSNR}   & SSIM  \\ \hline 
        Original Model &    28.01 & 0.8091 &  29.39 & 0.8282   \\
        +Self-ensemble &    28.16 & 0.8172   &  29.53 & 0.8433   \\
        +Model-ensemble &    \textbf{28.32} & \textbf{0.8197}   &  \textbf{29.71} & \textbf{0.8452} \\
        \toprule[1.2pt]
        \end{tabular}%
  }
\end{center}
\vspace{-0.5cm}
\end{table}%

\begin{table*}[!h]
    \fontsize{9}{9.6}\selectfont
    \caption{Quantitative comparisons between EDPN and existing methods on the REDS validation set. \textbf{Top}: $4\times$  BISR; \textbf{Bottom}: BID. \redbf{Red} and \blueud{blue} indicate the best and the second best performance, respectively.}
    \label{table_comparison}
    \vspace{-0.65cm}
    \begin{center}
      \tabcolsep=0.5cm
      \resizebox{0.9\linewidth}{!}{
\begin{tabular}{c||c||c|c|c|c|c}
\bottomrule[1.2pt]

Task               & Method        & PSNR$\uparrow$      & SSIM$\uparrow$ & LILPS$\downarrow$ & \#Param (M)& Running time (s)          \\ \hline
\multirow{5}{*}{BISR} 
                      & MSRN          &  26.65   & 0.7576   & 0.1147     & 5.80     & 0.0427 \\ 
                      &  GFN         &   26.91  &  0.7647  &   0.1139    &12.21 & 0.0479 \\ 
                      & RCAN           & {27.15}   &   {0.7740}  &  {0.1093}     &14.87 &0.0982 \\ 
                      & EDVR           & \blueud{27.63}   &   \blueud{0.8033}  &  \blueud{0.1002}     & 12.38  & 0.1426\\ 
                      & \textbf{EDPN} &   \redbf{28.01}  &   \redbf{0.8091}  &  \redbf{0.0819}     &  13.34   & 0.2224  \\ \hline
\multirow{5}{*}{BID}  & RNAN          &  26.73 & 0.7672  &  0.1071     & 8.54 & 0.1052 \\ 
                      & MPRNet         &   27.52  &  0.7896  &    0.1002   & 19.19 & 0.0652 \\ 
                      & SRN          &  {27.71}  & {0.7950}   &  {0.0984}     &   9.76     & 0.0507 \\ 
                      & EDVR          &  \blueud{28.19}   &  \blueud{0.8156}  &   \blueud{0.0882}    &  12.38 & 0.1553  \\ 
                      & \textbf{EDPN} & \redbf{28.96} &  \redbf{0.8203}  &  \redbf{0.0767}  &  13.34 & 0.2224   \\ 
                      
\toprule[1.2pt]
\end{tabular}}

\end{center}
\vspace{-0.4cm}
\end{table*}%

\begin{table*}[!t]
	\small
	\caption{Challenge results on the test set of REDS. \textbf{Left}: Track 1 (Low Resolution); \textbf{Right}: Track 2 (JPEG Artifacts). \redbf{Red} and \blueud{blue} indicate the best and the second best performance, respectively.}
	\label{tab:reds}
	\vspace{-0.7cm}
    \begin{center}

		\tabcolsep=0.5cm
		\scalebox{0.9}{
			\begin{subtable}[t]{0.5\textwidth}
				\begin{tabular}{l|ccc}
\bottomrule[1.2pt]
Track 1       & \multicolumn{1}{l}{PSNR $\uparrow$} & \multicolumn{1}{l}{SSIM $\uparrow$} & \multicolumn{1}{l}{LPIPS $\downarrow$} \\ \hline
\textbf{EDPN}        & \redbf{29.04}                    & \redbf{0.8416}                   & \redbf{0.2397}                    \\
2nd   & \blueud{28.91}                   & \blueud{0.8246}                   & {0.2569}                   \\
3rd   & 28.51                    & 0.8172                   & 0.2547                    \\
4th   & 28.44                    & 0.8158                   & \blueud{0.2531}                    \\
5th   & 28.44                    & 0.8135                   & 0.2704                    \\
6th   & 28.42                    & 0.8132                   & 0.2685                    \\
7th   & 28.36                    & 0.8130                   & 0.2666                    \\
8th   & 28.33                    & 0.8132                   & 0.2606                    \\
9th   & 28.28                    & 0.8110                   & 0.2651                    \\
10th  & 28.25                    & 0.8108                   & 0.2636                    \\
\toprule[1.2pt]
\end{tabular}
\end{subtable}

\begin{subtable}[!t]{0.5\textwidth}
\begin{tabular}{l|ccc}
\bottomrule[1.2pt]
Track 2   & \multicolumn{1}{l}{PSNR $\uparrow$} & \multicolumn{1}{l}{SSIM $\uparrow$} & \multicolumn{1}{l}{LPIPS $\downarrow$} \\ \hline
1st  & \redbf{29.70}                    & \blueud{0.8403}                   & 0.2319                    \\
2nd & \blueud{29.62}                   & 0.8397                   & 0.2304                    \\
3rd & 29.60                    & 0.8398                   & \blueud{0.2302}                    \\
4th  & 29.59                    & 0.8381                   & {0.2340}                    \\
5th   & 29.56                    & 0.8385                   & 0.2322                    \\
6th  & 29.34                    & 0.8355                   & 0.2546                    \\
\textbf{EDPN}      & 29.33                    & \redbf{0.8565}                   & \redbf{0.2222}                    \\
8th   & 29.17                    & 0.8325                   & 0.2411                    \\
9th  & 29.11                    & 0.8292                   & 0.2449                    \\
10th & 29.07                    & 0.8286                   & 0.2499                    \\
\toprule[1.2pt]
\end{tabular}
			\end{subtable}
		}
		
	\end{center}
	\vspace{-0.7cm}
\end{table*}

\textbf{Network architecture.}\label{subsec:abl_net}
We perform an ablation experiment to demonstrate the effectiveness of the PPT and PSA modules. 
First, we construct a basic model for comparison.
In the basic model, we remove the pyramid structure and substitute our proposed PTBs by cascading several residual blocks in the PPT module. 
The PSA module is replaced by a cascade of several residual blocks and convolution layers in the basic model. 
Then we recover our design by adding the PPT and PSA modules step by step.
The comparison results are shown in Table~\ref{table:module_ablation}. 
When the PPT module is added, the PSNR value in RGB channel is improved from 27.43 dB to 27.84 dB. 
When the PSA module is added, the PSNR value in RGB channel is improved from 27.43 dB to 27.78 dB. 
It demonstrates that the PPT and PSA modules are proved to be highly effective for the blurry image restoration tasks.
After adding these two modules, the PSNR value is further improved to  28.01 dB.
Similar phenomenon also appears on other metrics.
This ablation experiment demonstrates that our proposed modules are effective on the general blurring image restoration task.

\textbf{PTB number.}\label{subsec:abl_PTblock}
To determine the number of PTBs used at each level of the PPT module, we compare the quantitative performance under different settings with $N=1, 2, 3, 4, 5$. 
The comparison results are shown in Table~\ref{table:ablation_block}. 
It can be observed that better performance can be achieved with the increase of the number $N$, which indicates that using more PTBs is more effective in extracting the self-similarity.
To balance the computing efficiency and the performance, we deploy 3 PTBs at each level of the PPT module.

\textbf{Input image number.}\label{subsec:abl_frame}
We train our EDPN with different numbers of replicated images $(K = 0, 2, 4, 6)$ as inputs and compare the performance in Table~\ref{table: frame_ablation}.
It can be observed that using 4 replicated images as inputs generates the best performance in terms of PSNR.
It is worth noting that, the PSNR value in RGB channel is improved from 27.89 dB to 28.01 dB when the number of input images increases from 1 to 5.
In the experiments, we always adopt 4 times replication for the inputs.

\textbf{Loss function.}\label{subsec:abl_loss}
We investigate the contribution of different loss terms by adjusting the weighting factors in Equation~\ref{lossss}, and the results are shown in Table~\ref{table:loss}.
Since $\mathcal{L}_{Charb}$ is optimized at the pixel level, the best result can be achieved in terms of PSNR.
When $\mathcal{L}_{SSIM}$ is adopted together with $\mathcal{L}_{Charb}$, we can obtain a higher SSIM value.
In the experiments, we train EDPN with both $\mathcal{L}_{Charb}$ and $\mathcal{L}_{SSIM}$ for a better tradeoff between PSNR and SSIM.

\begin{figure*}[!t]
    \vspace{-0.3cm}
    \fontsize{7}{9.6}\selectfont
    \begin{center}
    \begin{minipage}{0.98\linewidth}

    \begin{minipage}{0.152\linewidth}
      \centerline{\includegraphics[width=1.1\linewidth]{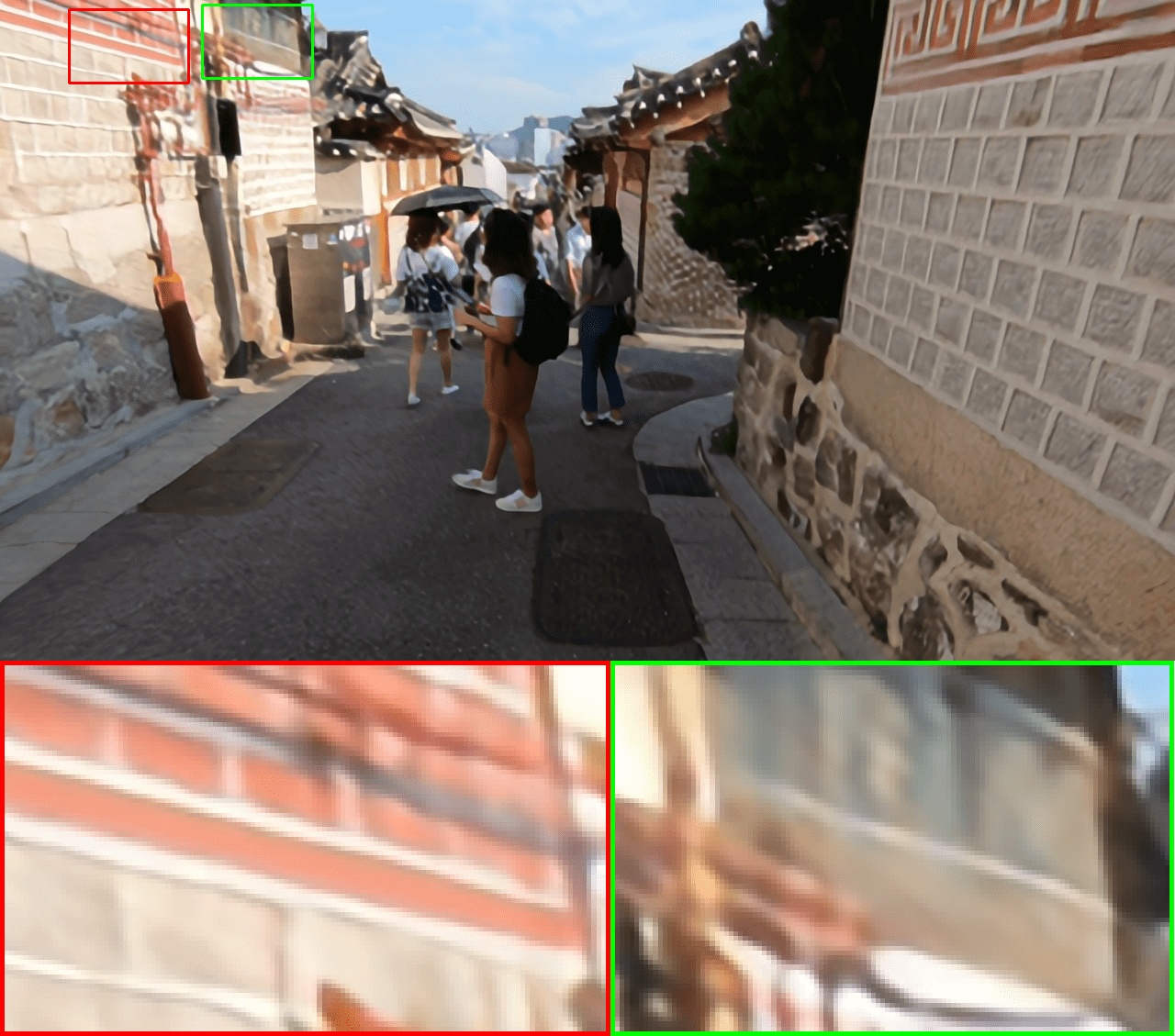}}
    \end{minipage}
  \hfill
    \begin{minipage}{0.152\linewidth}
      \centerline{\includegraphics[width=1.1\linewidth]{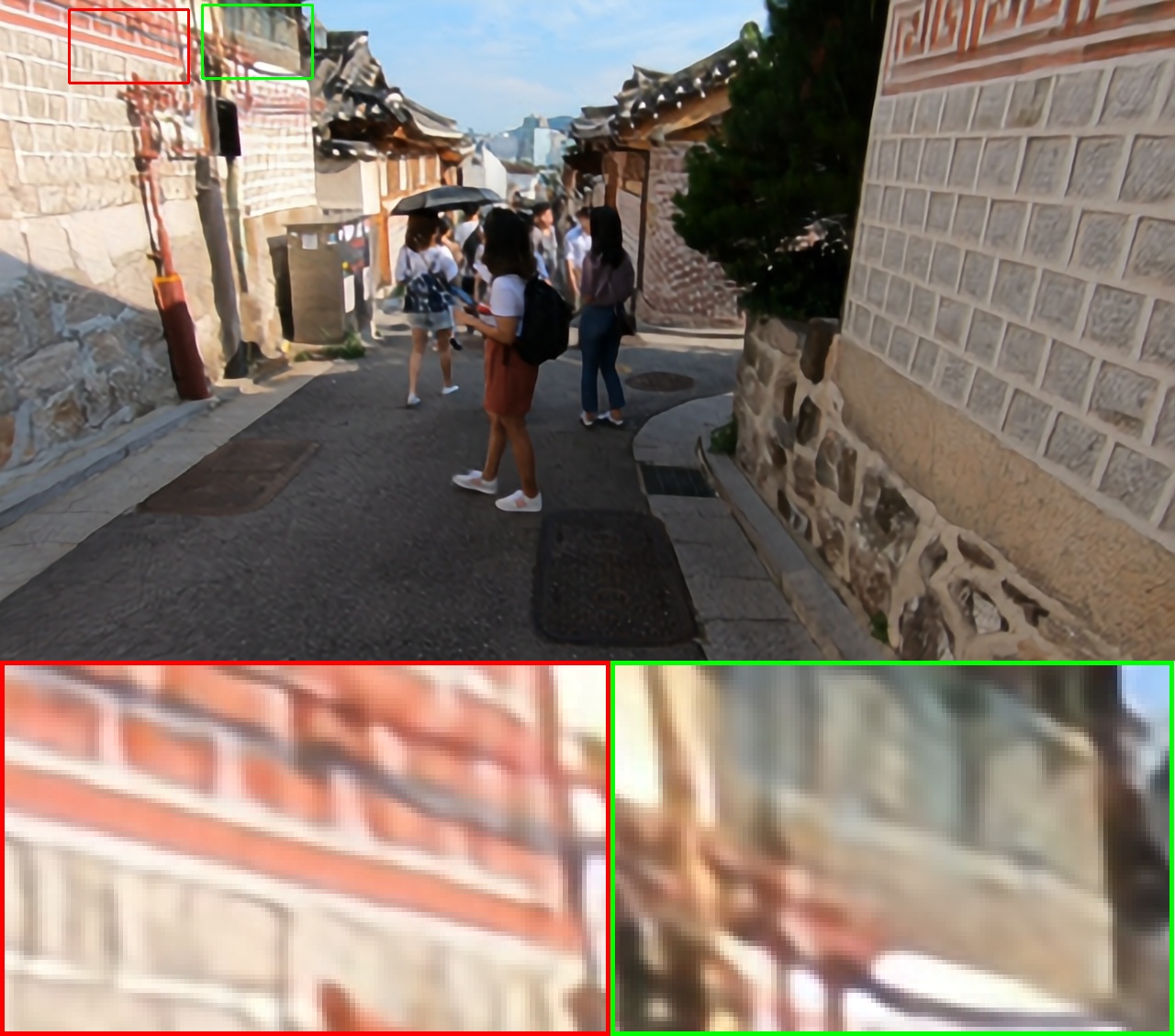}}
    \end{minipage}
  \hfill
    \begin{minipage}{0.152\linewidth}
      \centerline{\includegraphics[width=1.1\linewidth]{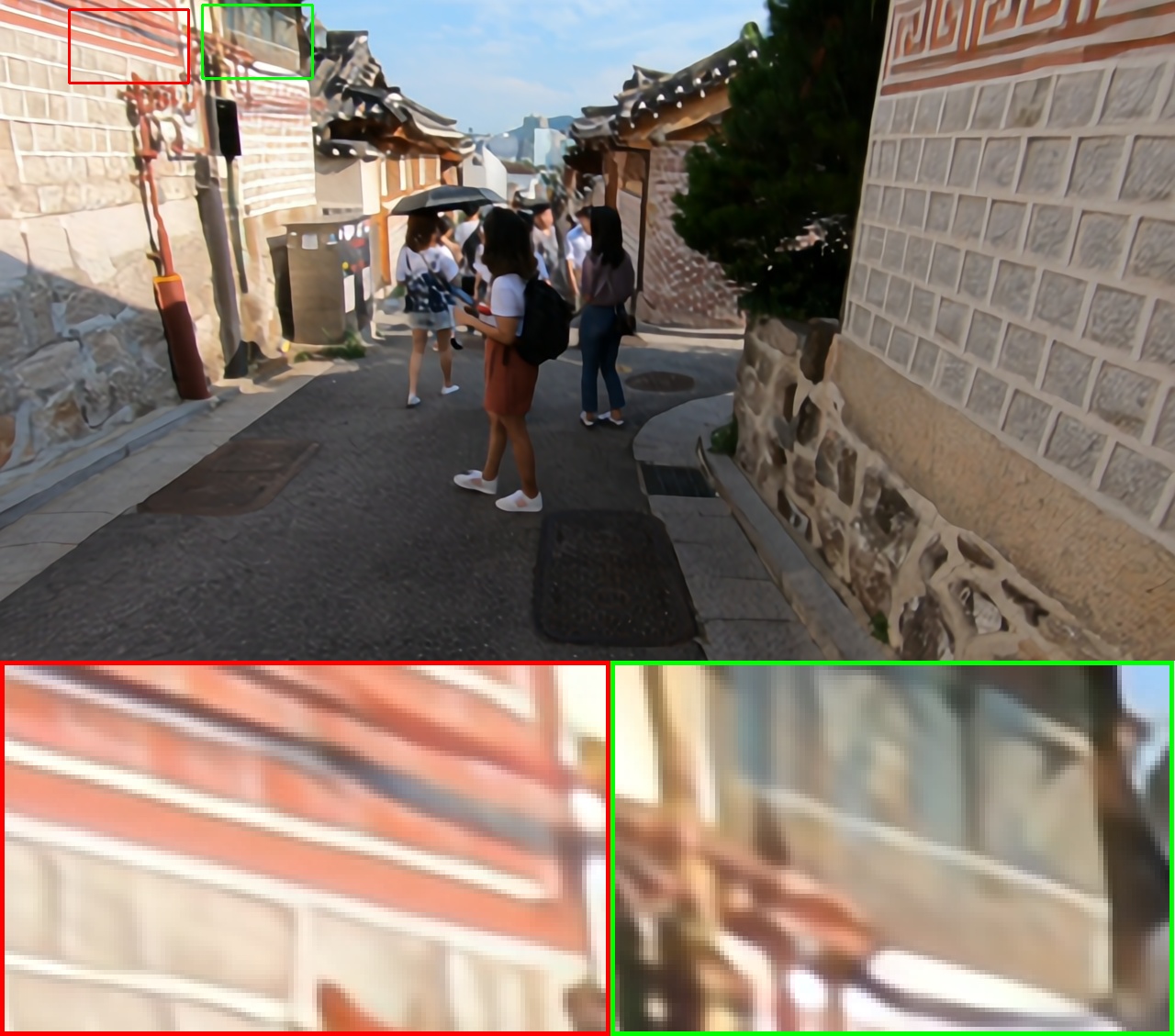}}
    \end{minipage}
  \hfill
  \begin{minipage}{0.152\linewidth}
      \centerline{\includegraphics[width=1.1\linewidth]{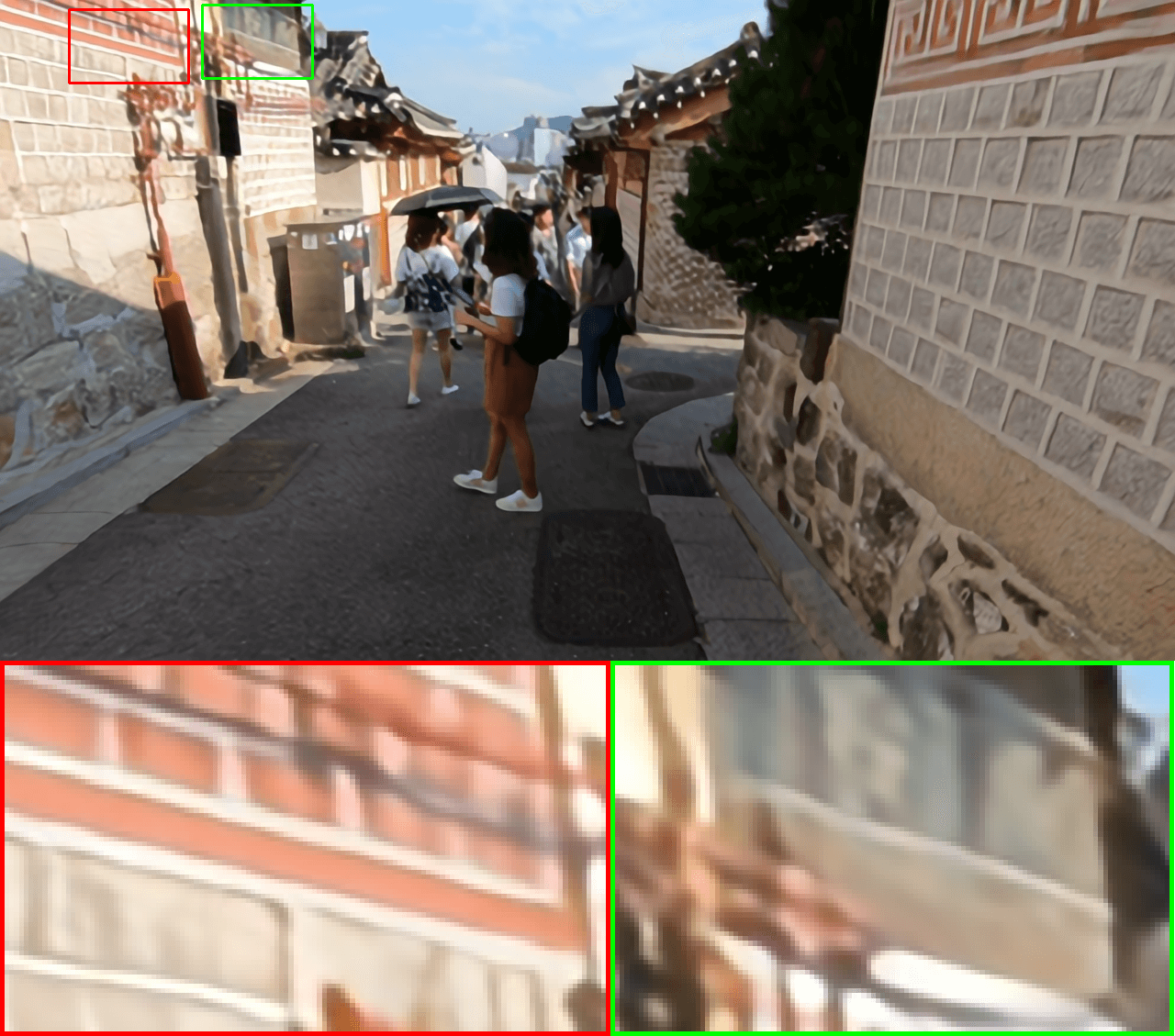}}
    \end{minipage}
  \hfill
    \begin{minipage}{0.152\linewidth}
      \centerline{\includegraphics[width=1.1\linewidth]{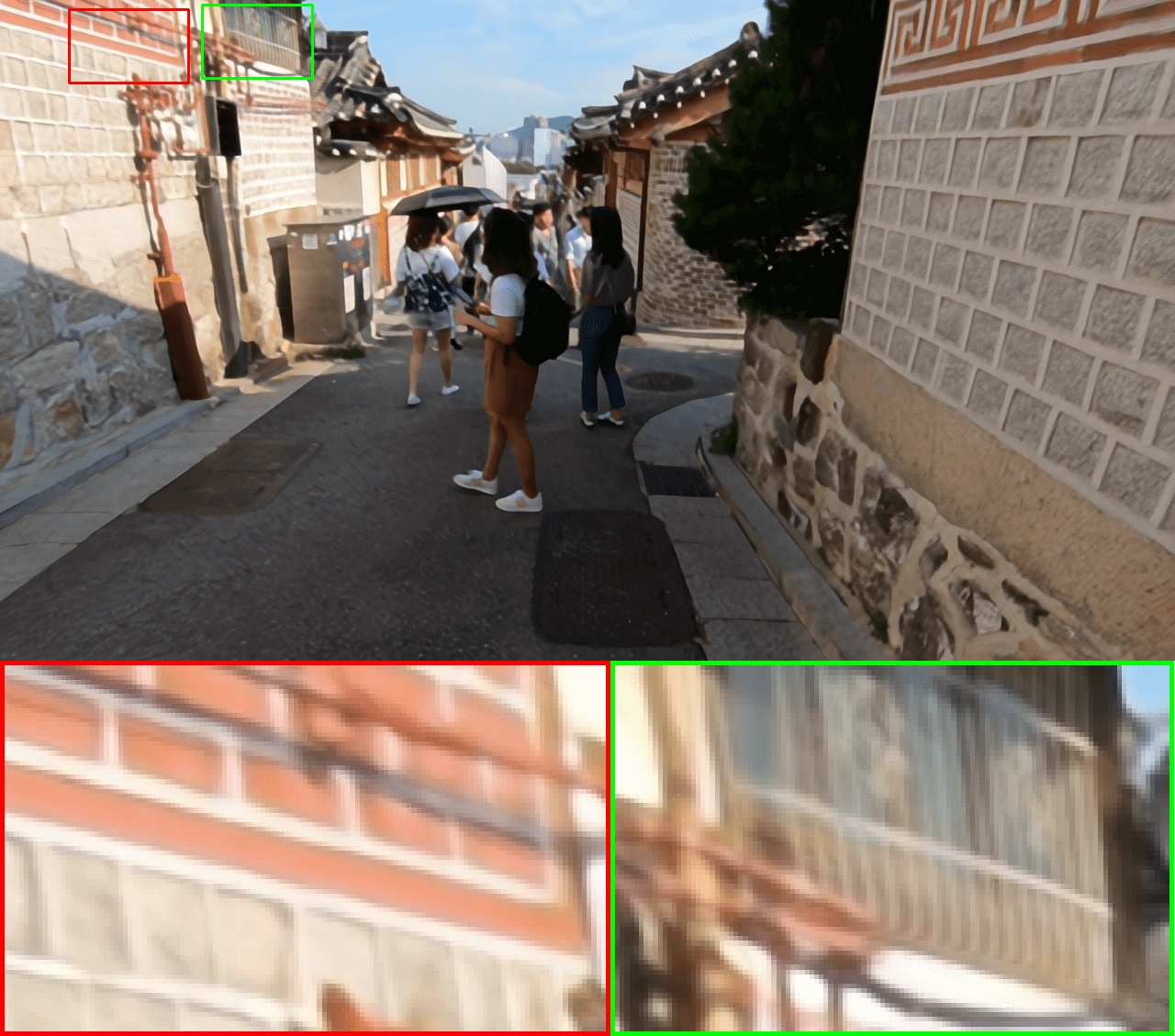}}
    \end{minipage}
\hfill
      \begin{minipage}{0.152\linewidth}
      \centerline{\includegraphics[width=1.1\linewidth]{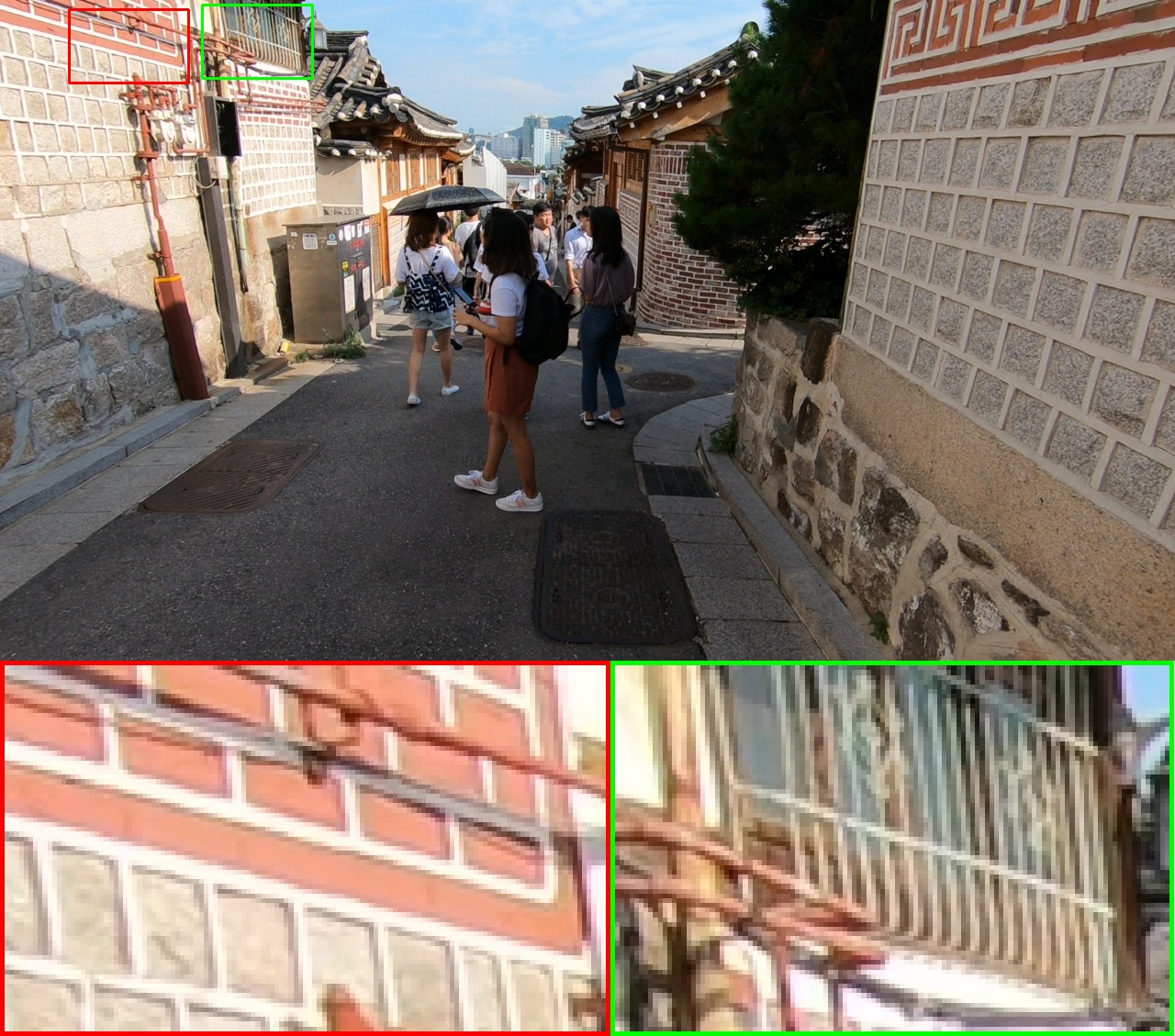}}
    \end{minipage}

  \vfill

    \begin{minipage}{0.152\linewidth}
      \centerline{\includegraphics[width=1.1\linewidth]{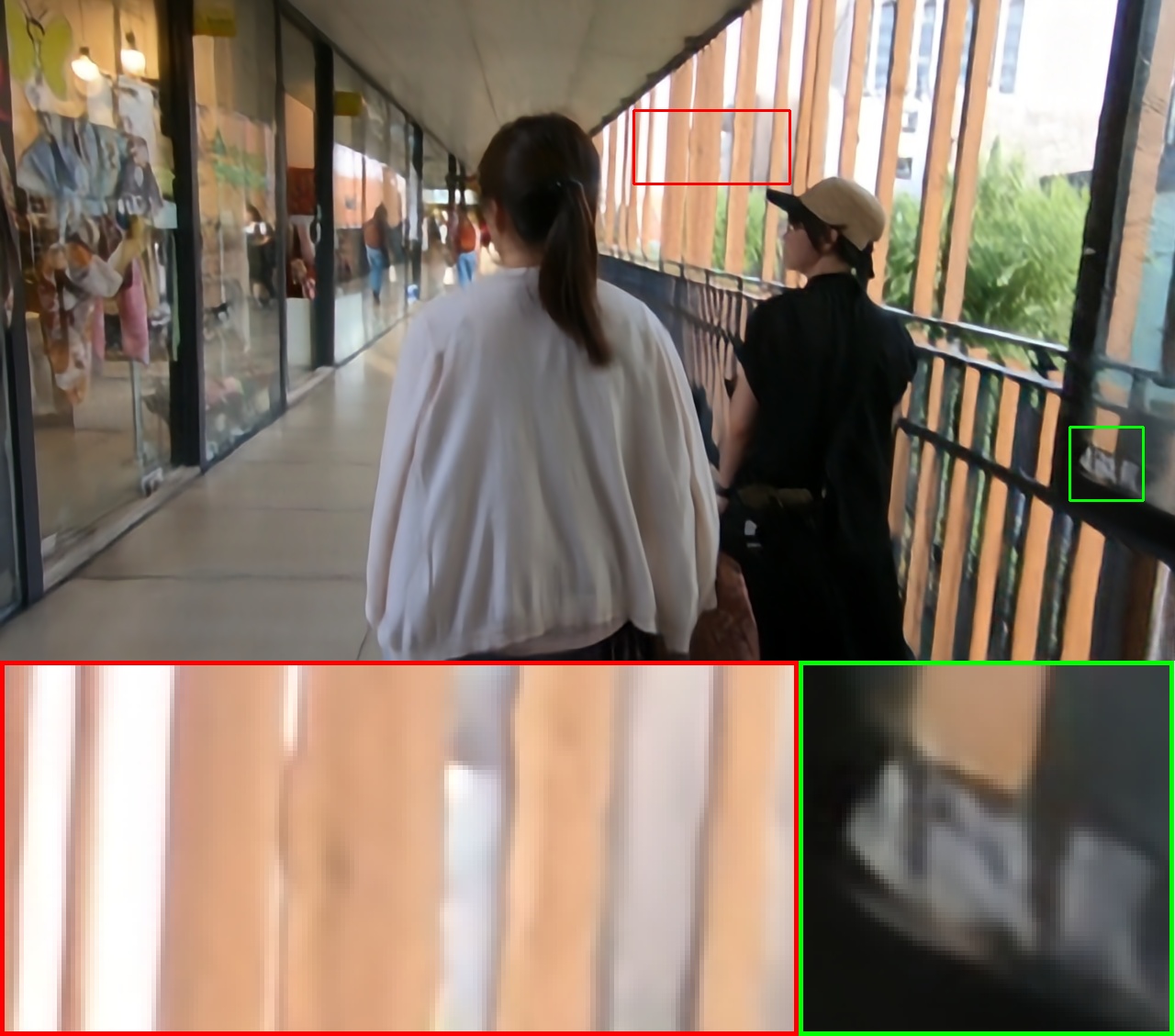}}
      \centerline{MSRN}
    \end{minipage}
  \hfill
    \begin{minipage}{0.152\linewidth}
      \centerline{\includegraphics[width=1.1\linewidth]{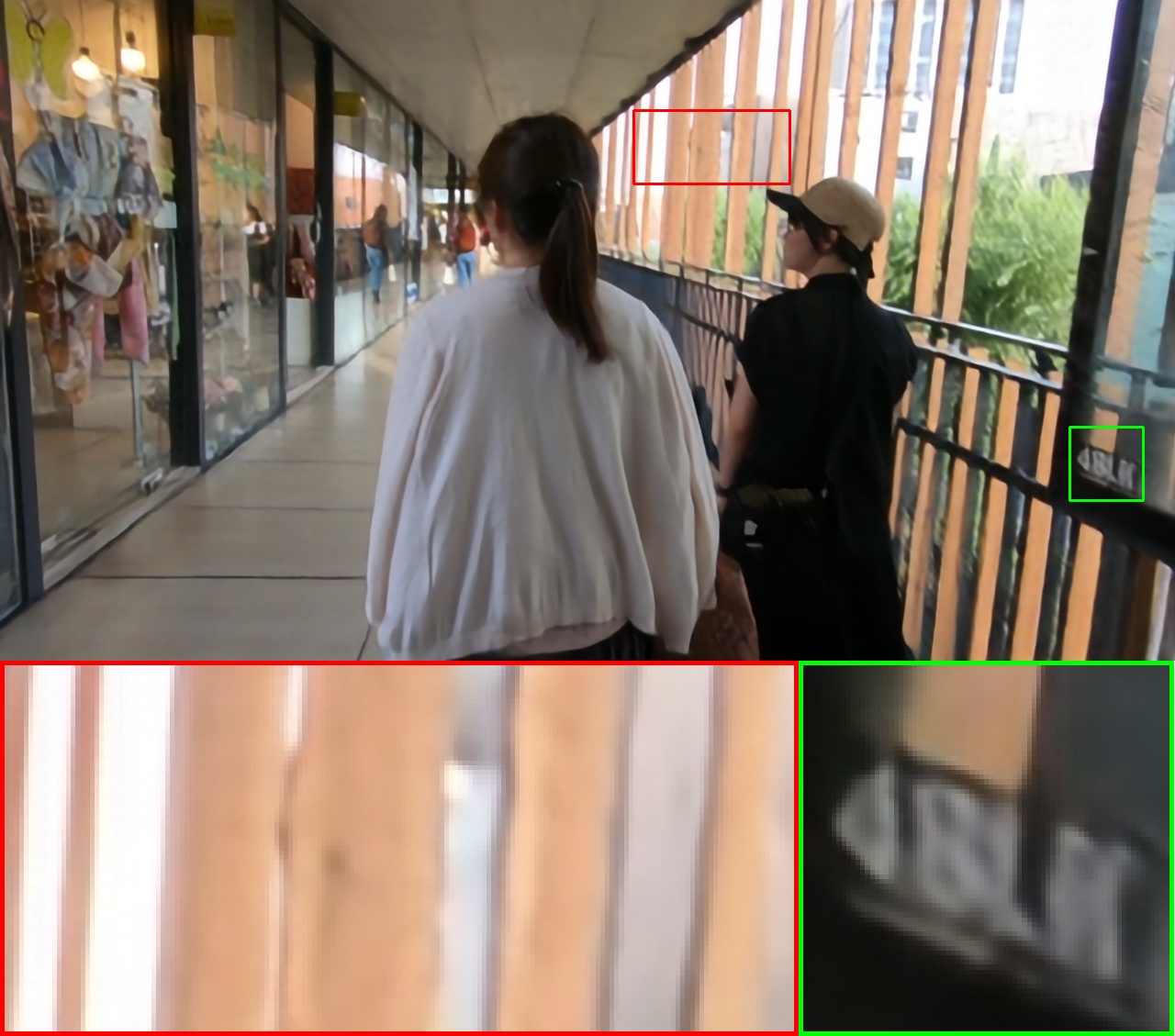}}
      \centerline{GFN}
    \end{minipage}
  \hfill
    \begin{minipage}{0.152\linewidth}
      \centerline{\includegraphics[width=1.1\linewidth]{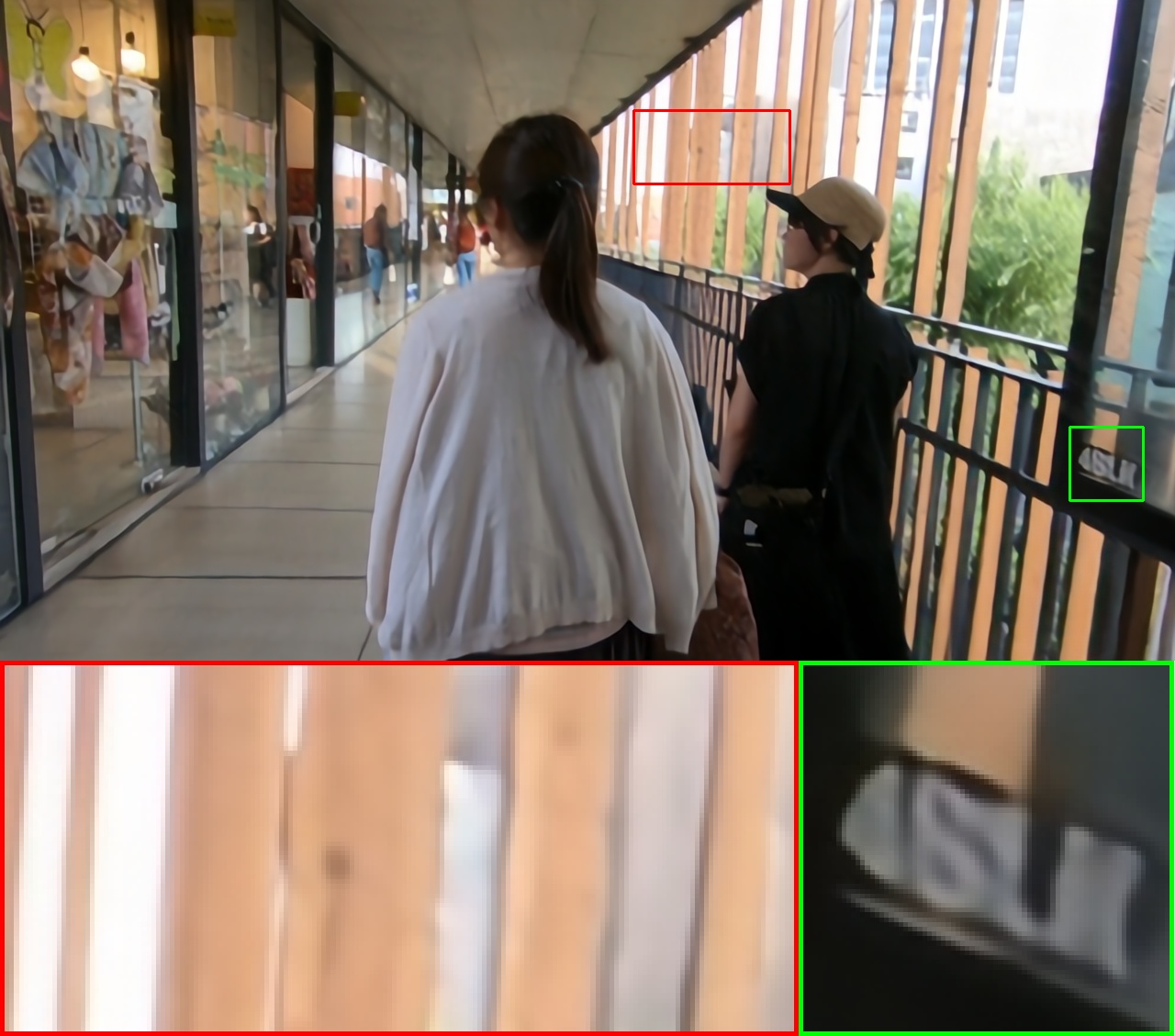}}
      \centerline{RCAN}
    \end{minipage}
  \hfill
  \begin{minipage}{0.152\linewidth}
      \centerline{\includegraphics[width=1.1\linewidth]{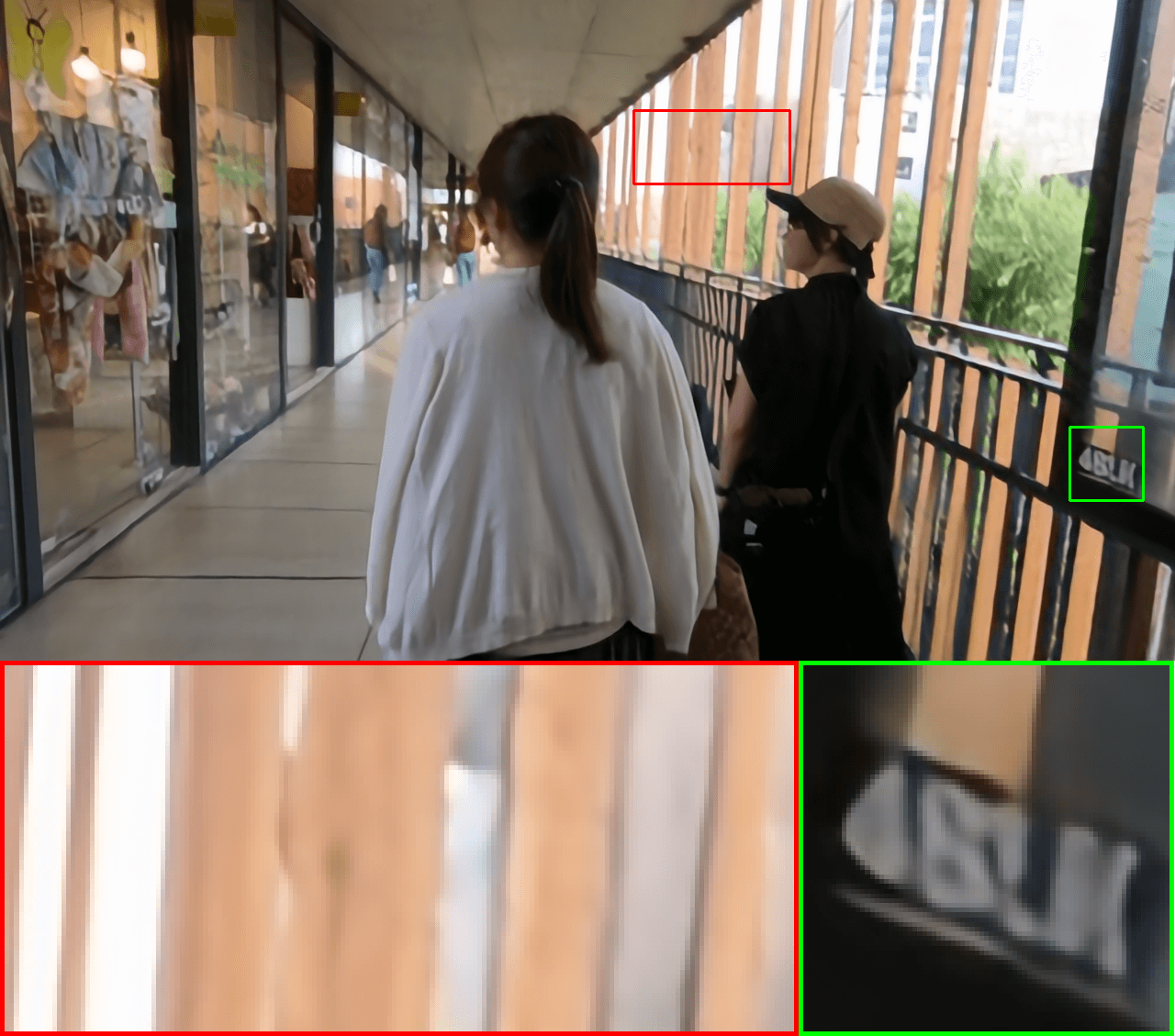}}
      \centerline{EDVR}
    \end{minipage}
  \hfill
    \begin{minipage}{0.152\linewidth}
      \centerline{\includegraphics[width=1.1\linewidth]{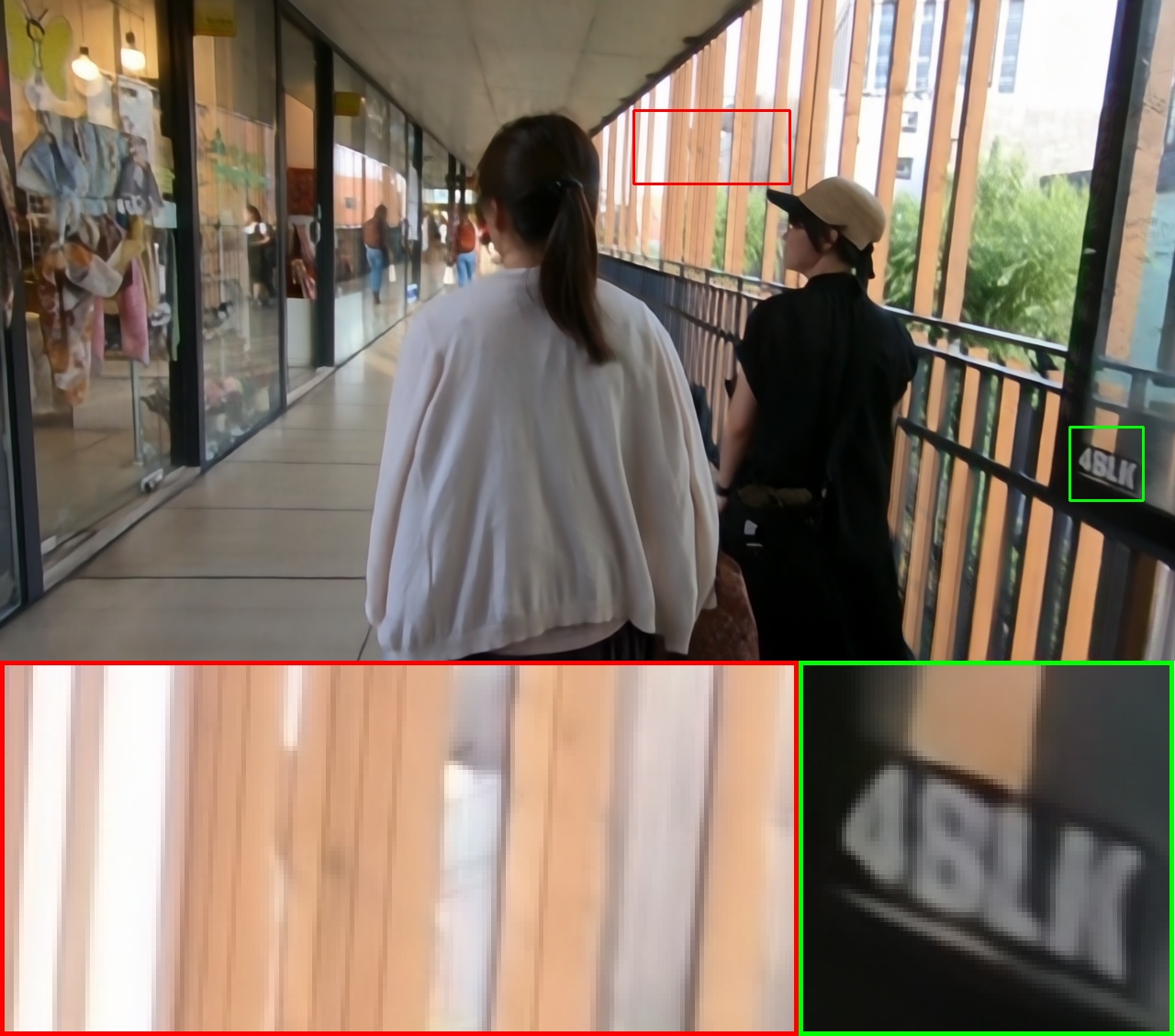}}
      \centerline{\textbf{EDPN (Ours)}}
    \end{minipage}
     \hfill
        \begin{minipage}{0.152\linewidth}
      \centerline{\includegraphics[width=1.1\linewidth]{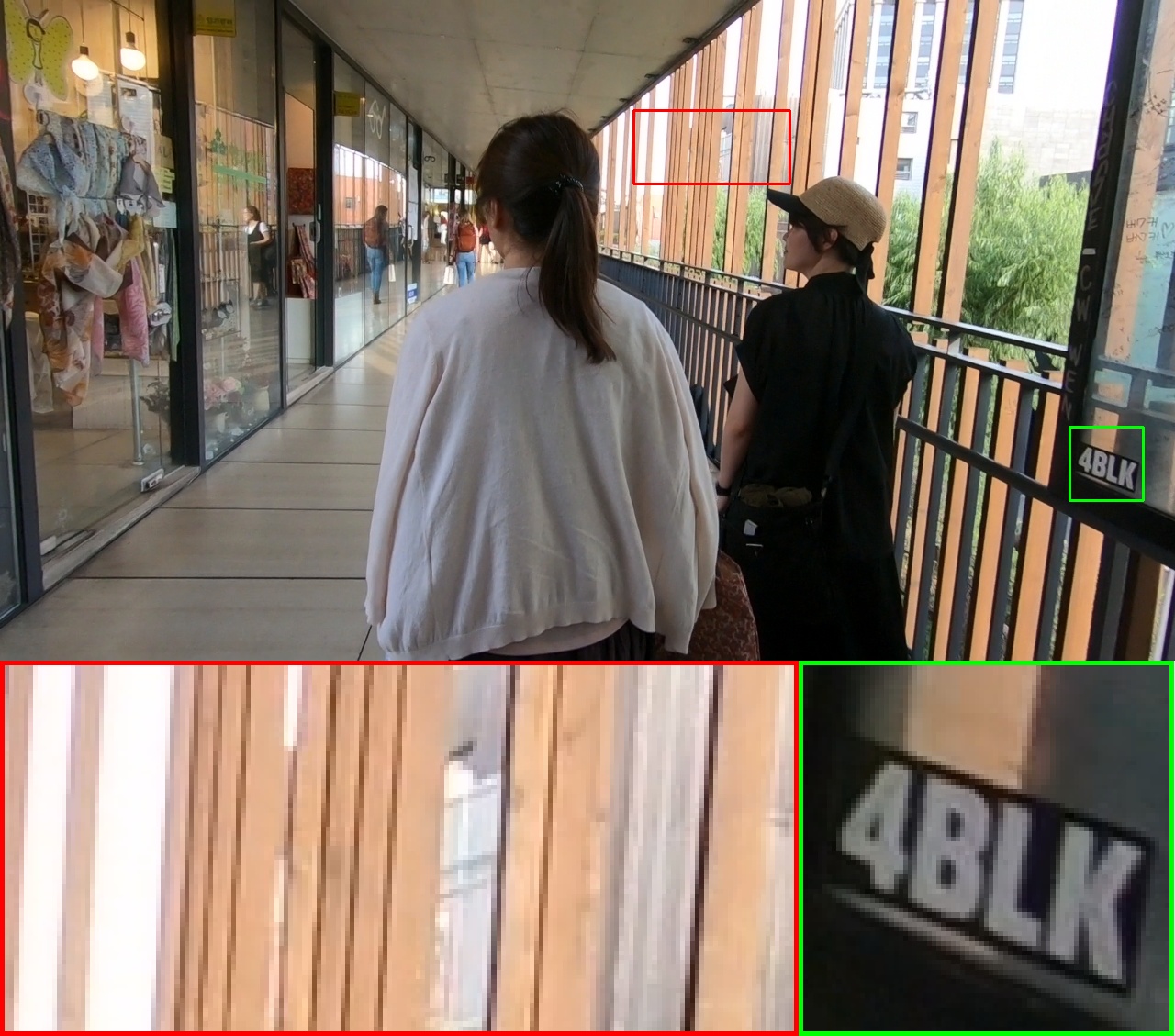}}
      \centerline{Ground truth}
    \end{minipage}

    \end{minipage}
    \end{center}
    \vspace{-0.6cm}
    \caption{Qualitative comparisons on the validation set of REDS for BISR. Please zoom in for better visualization.}
    \vspace{0.2cm}
    \label{fig_compare1}
  \end{figure*}

\begin{figure*}[!t]
    \fontsize{7}{9.6}\selectfont
    \begin{center}
    \begin{minipage}{0.98\linewidth}

    \begin{minipage}{0.152\linewidth}
      \centerline{\includegraphics[width=1.1\linewidth]{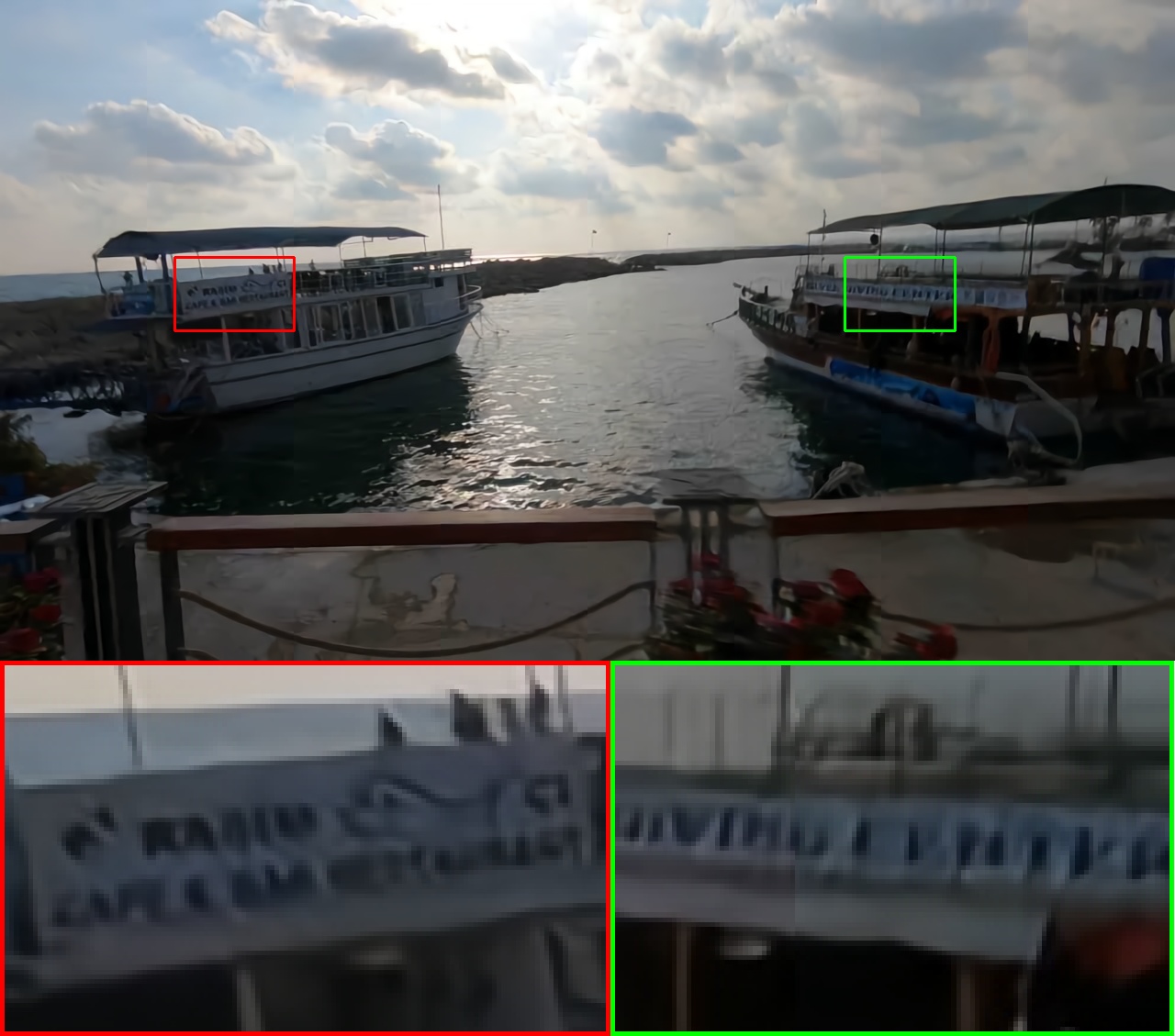}}
    \end{minipage}
  \hfill
    \begin{minipage}{0.152\linewidth}
      \centerline{\includegraphics[width=1.1\linewidth]{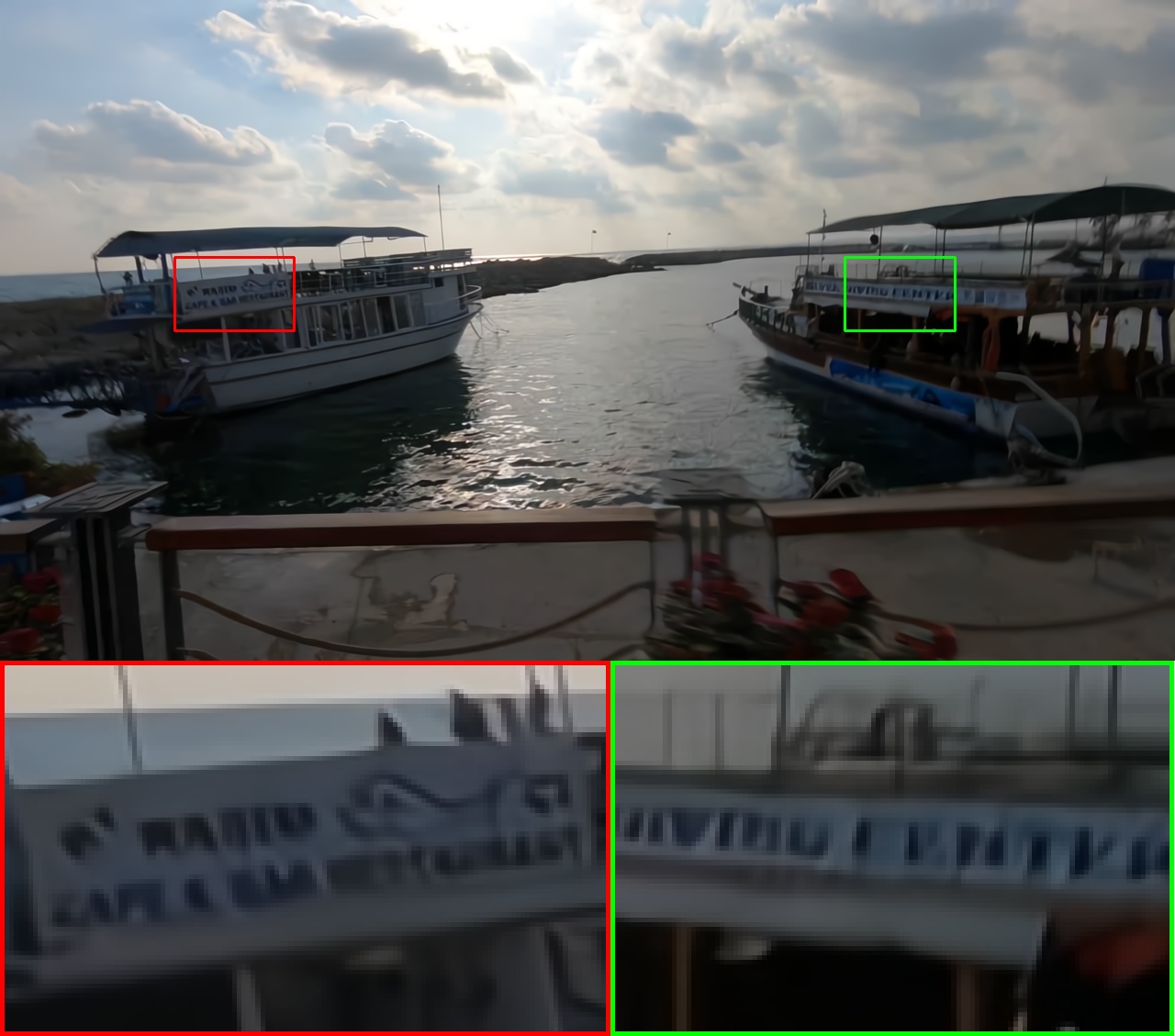}}
    \end{minipage}
  \hfill
    \begin{minipage}{0.152\linewidth}
      \centerline{\includegraphics[width=1.1\linewidth]{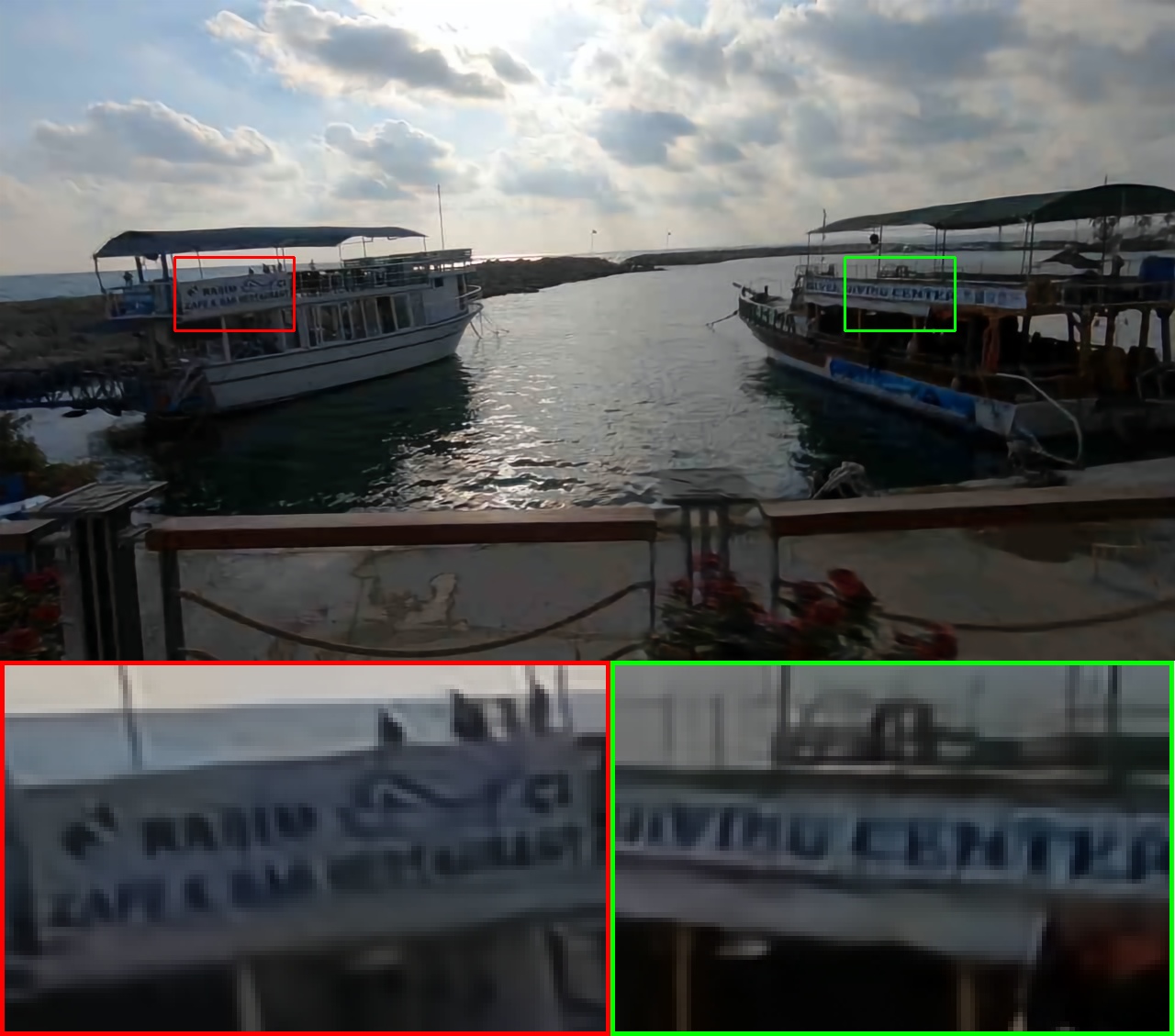}}
    \end{minipage}
  \hfill
   \begin{minipage}{0.152\linewidth}
      \centerline{\includegraphics[width=1.1\linewidth]{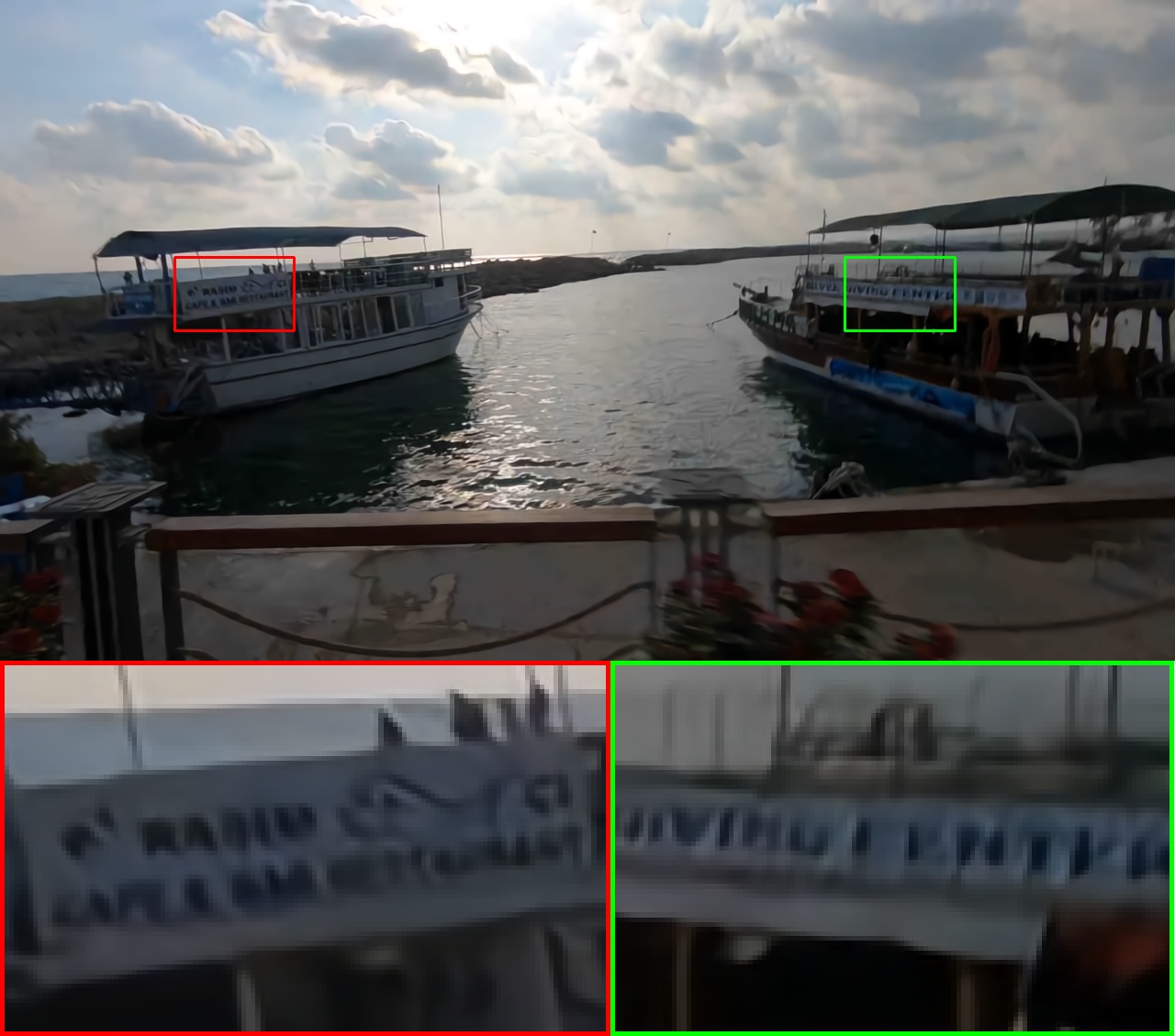}}
    \end{minipage}
  \hfill
    \begin{minipage}{0.152\linewidth}
      \centerline{\includegraphics[width=1.1\linewidth]{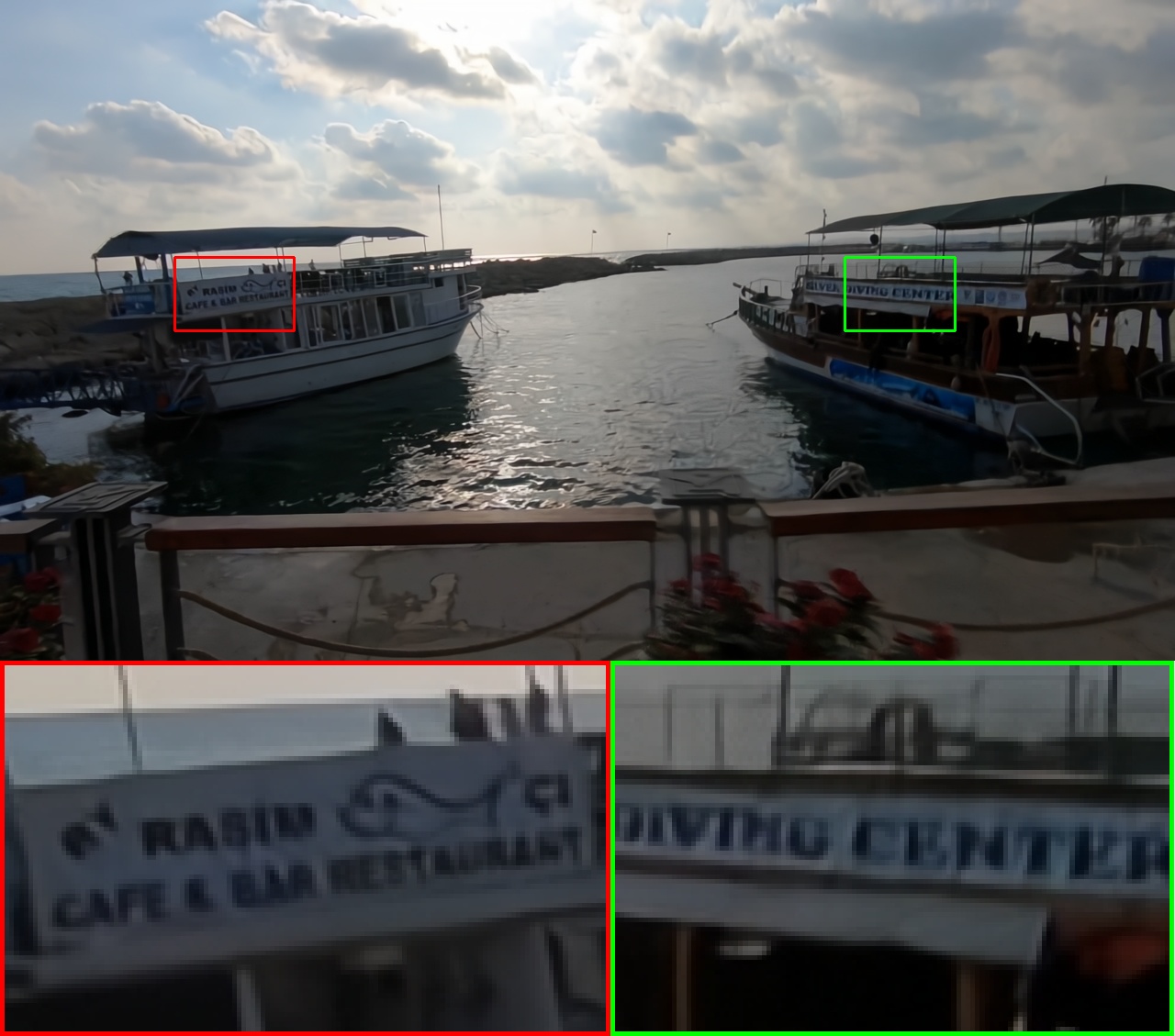}}
    \end{minipage}
    \hfill
  \begin{minipage}{0.152\linewidth}
      \centerline{\includegraphics[width=1.1\linewidth]{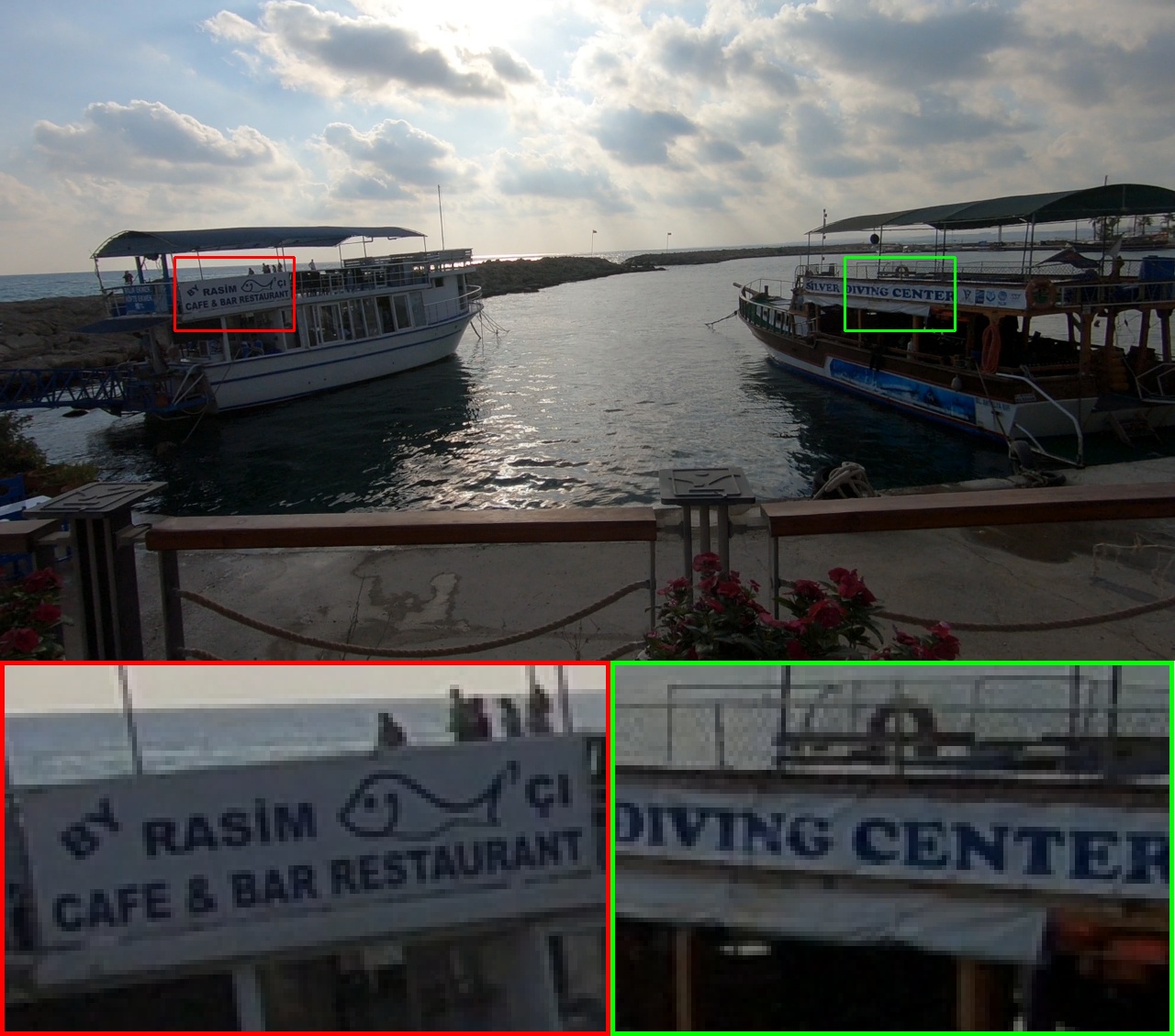}}
    \end{minipage}

  \vfill

    \begin{minipage}{0.152\linewidth}
      \centerline{\includegraphics[width=1.1\linewidth]{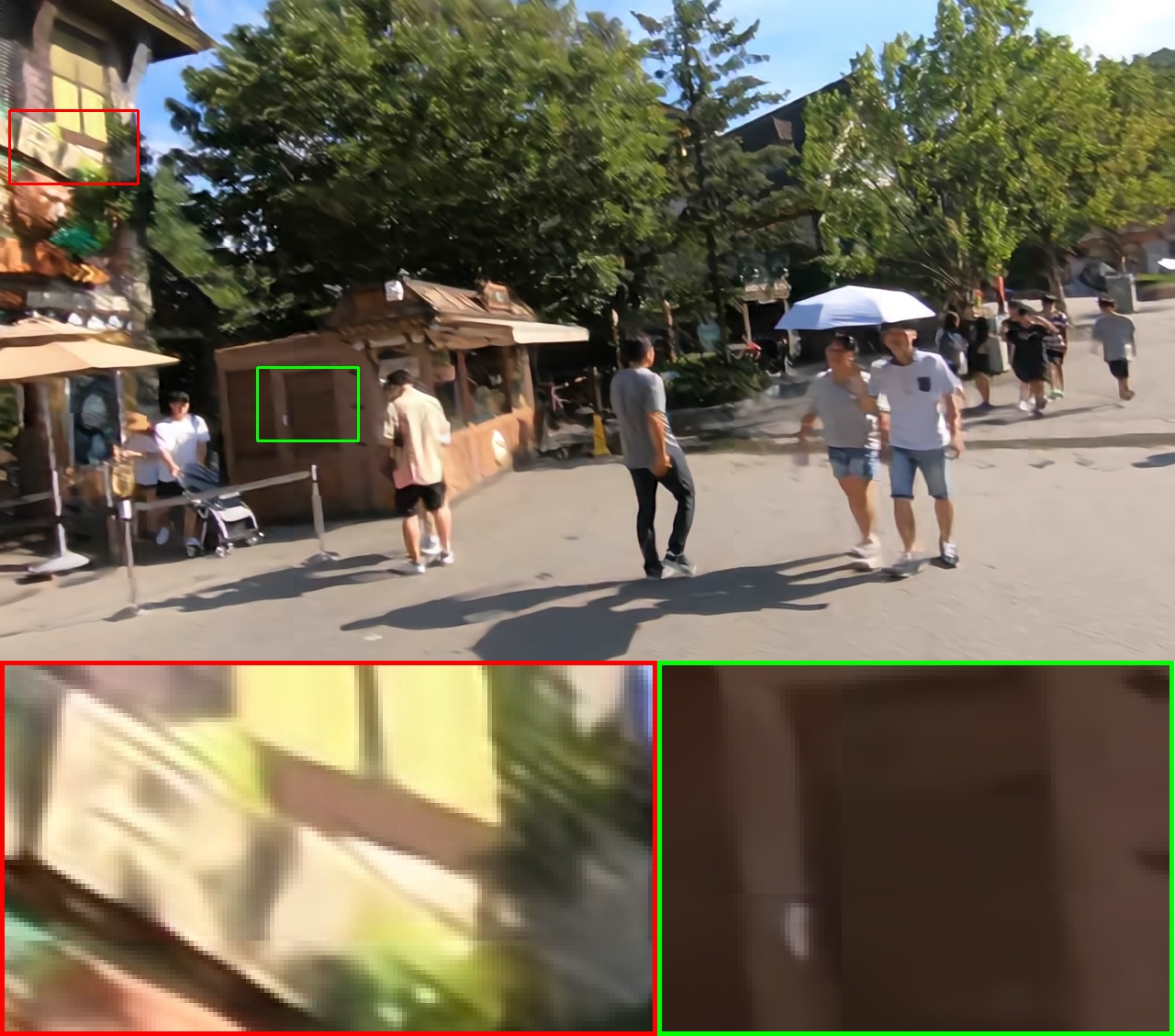}}
      \centerline{RNAN}
    \end{minipage}
  \hfill
    \begin{minipage}{0.152\linewidth}
      \centerline{\includegraphics[width=1.1\linewidth]{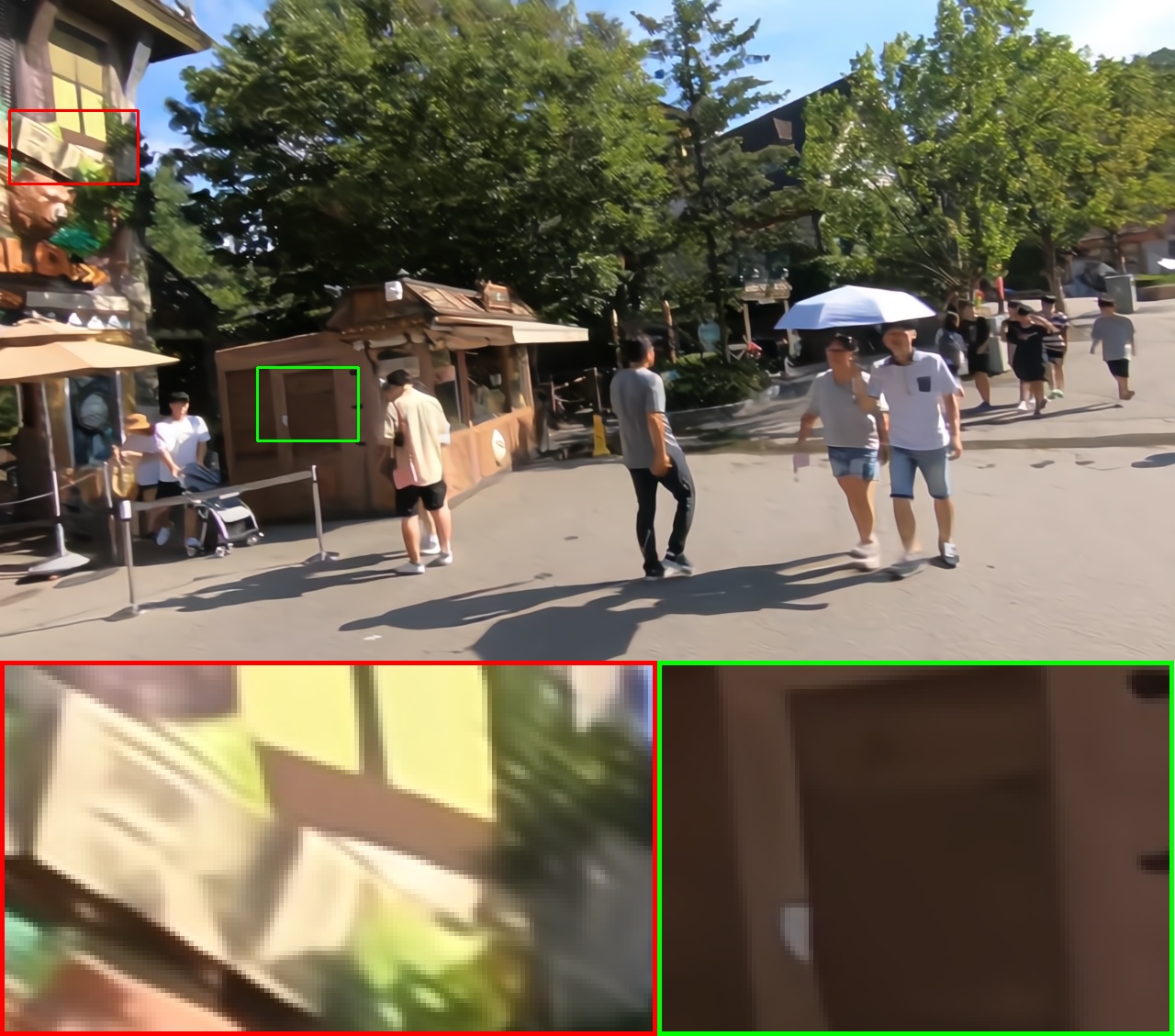}}
      \centerline{MPRNet}
    \end{minipage}
  \hfill
    \begin{minipage}{0.152\linewidth}
      \centerline{\includegraphics[width=1.1\linewidth]{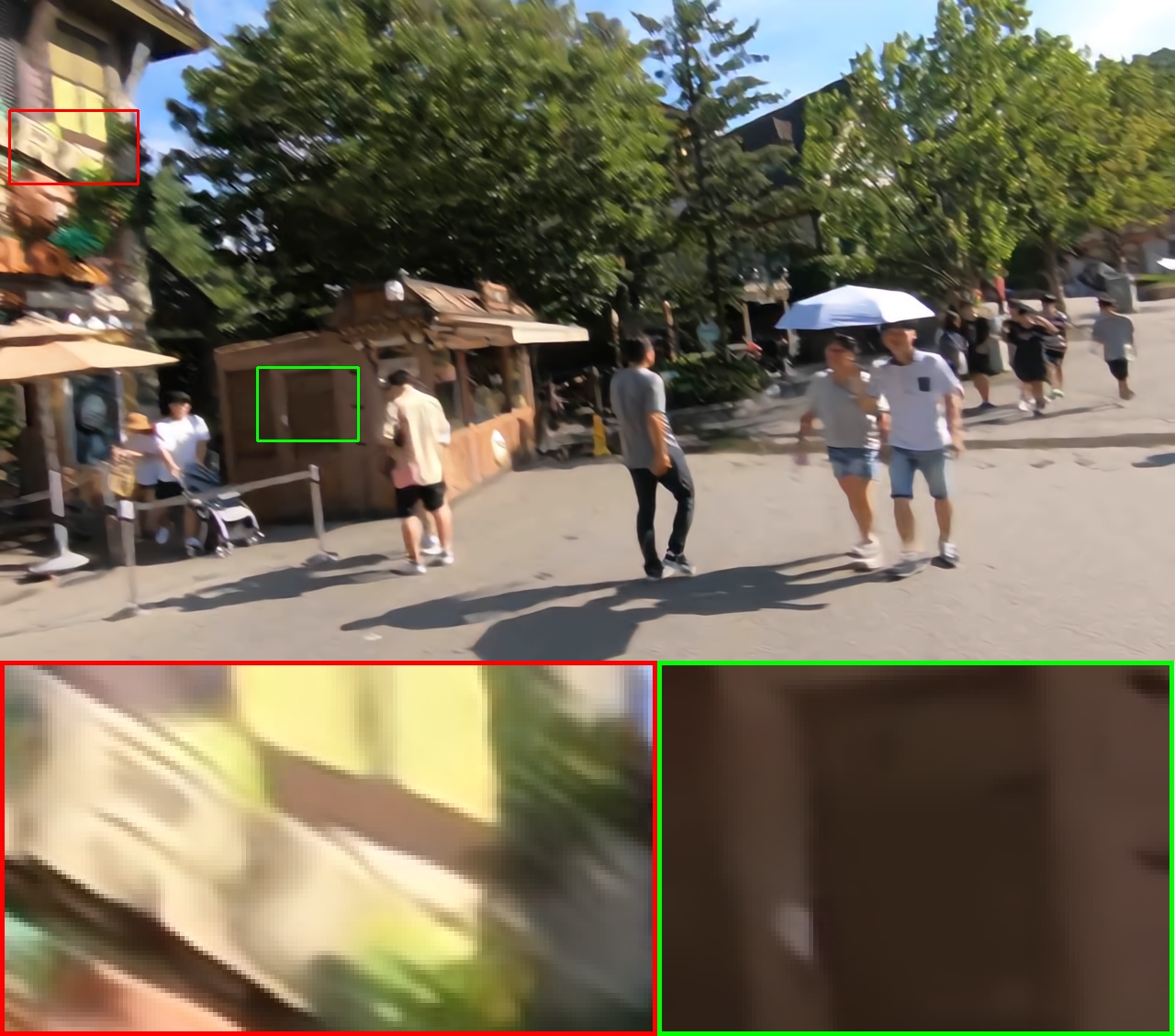}}
      \centerline{SRN}
    \end{minipage}
  \hfill
  \begin{minipage}{0.152\linewidth}
      \centerline{\includegraphics[width=1.1\linewidth]{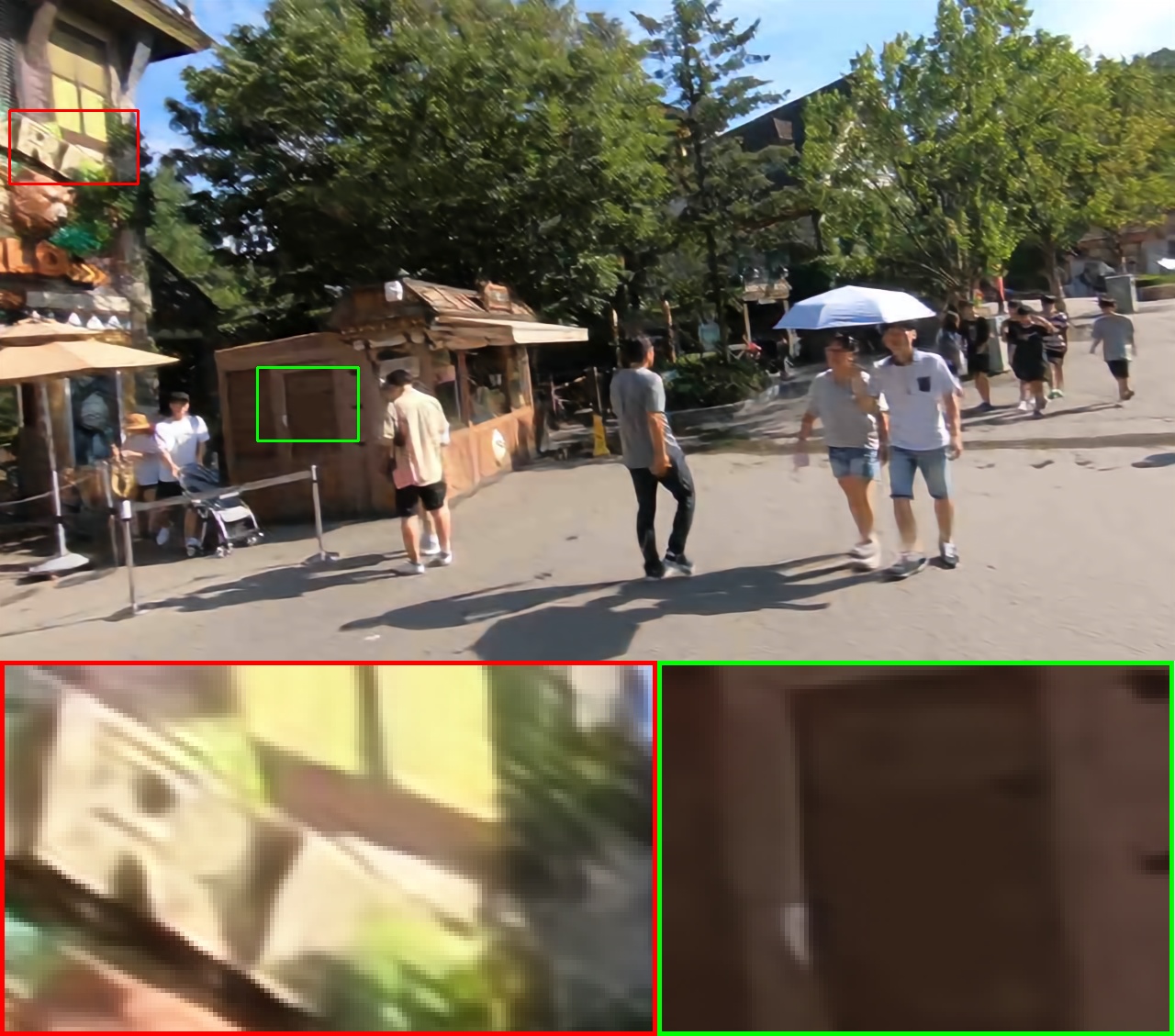}}
      \centerline{EDVR}
    \end{minipage}
  \hfill
    \begin{minipage}{0.152\linewidth}
      \centerline{\includegraphics[width=1.1\linewidth]{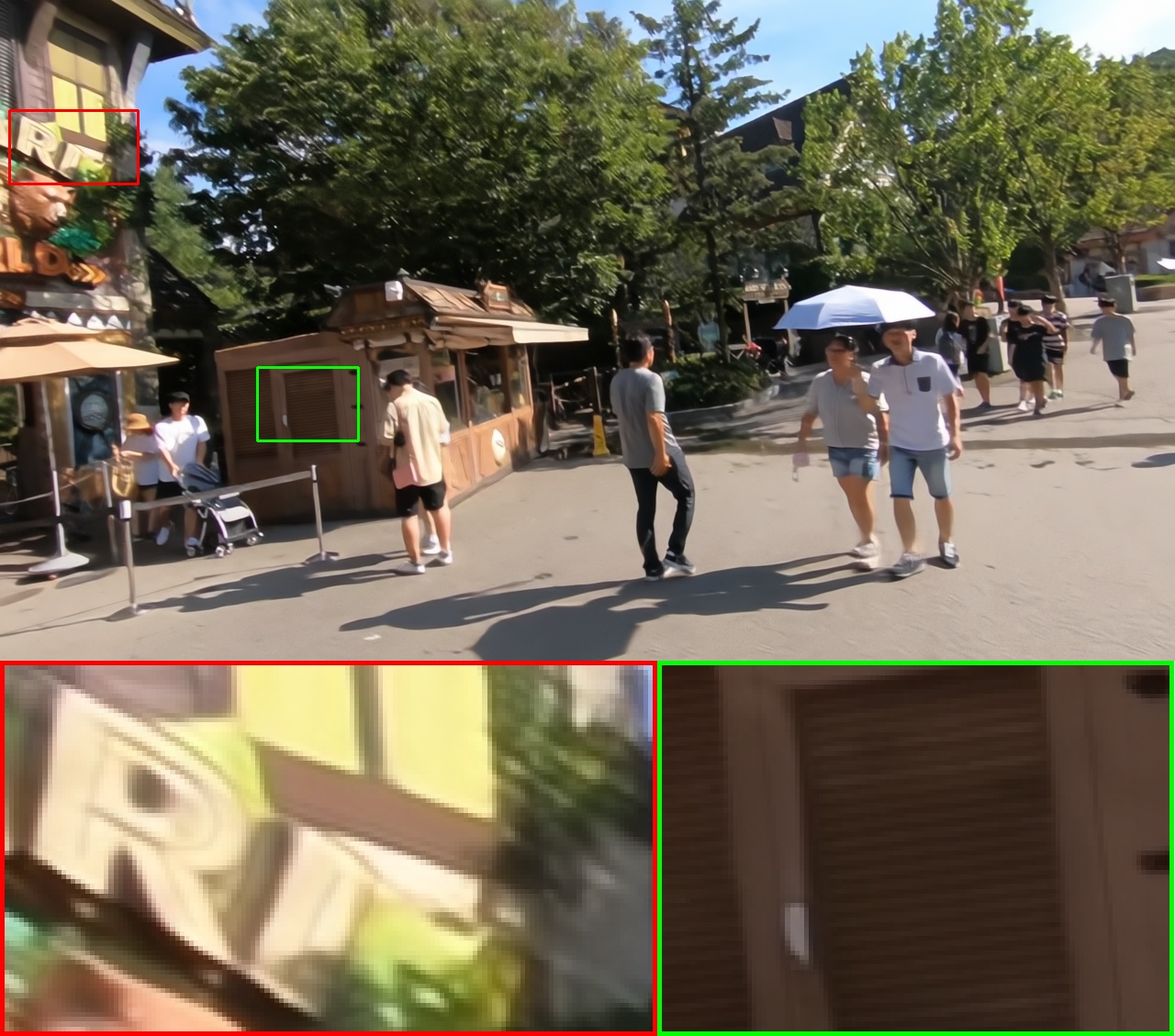}}
      \centerline{\textbf{EDPN (Ours)}}
    \end{minipage}
 \hfill
   \begin{minipage}{0.152\linewidth}
      \centerline{\includegraphics[width=1.1\linewidth]{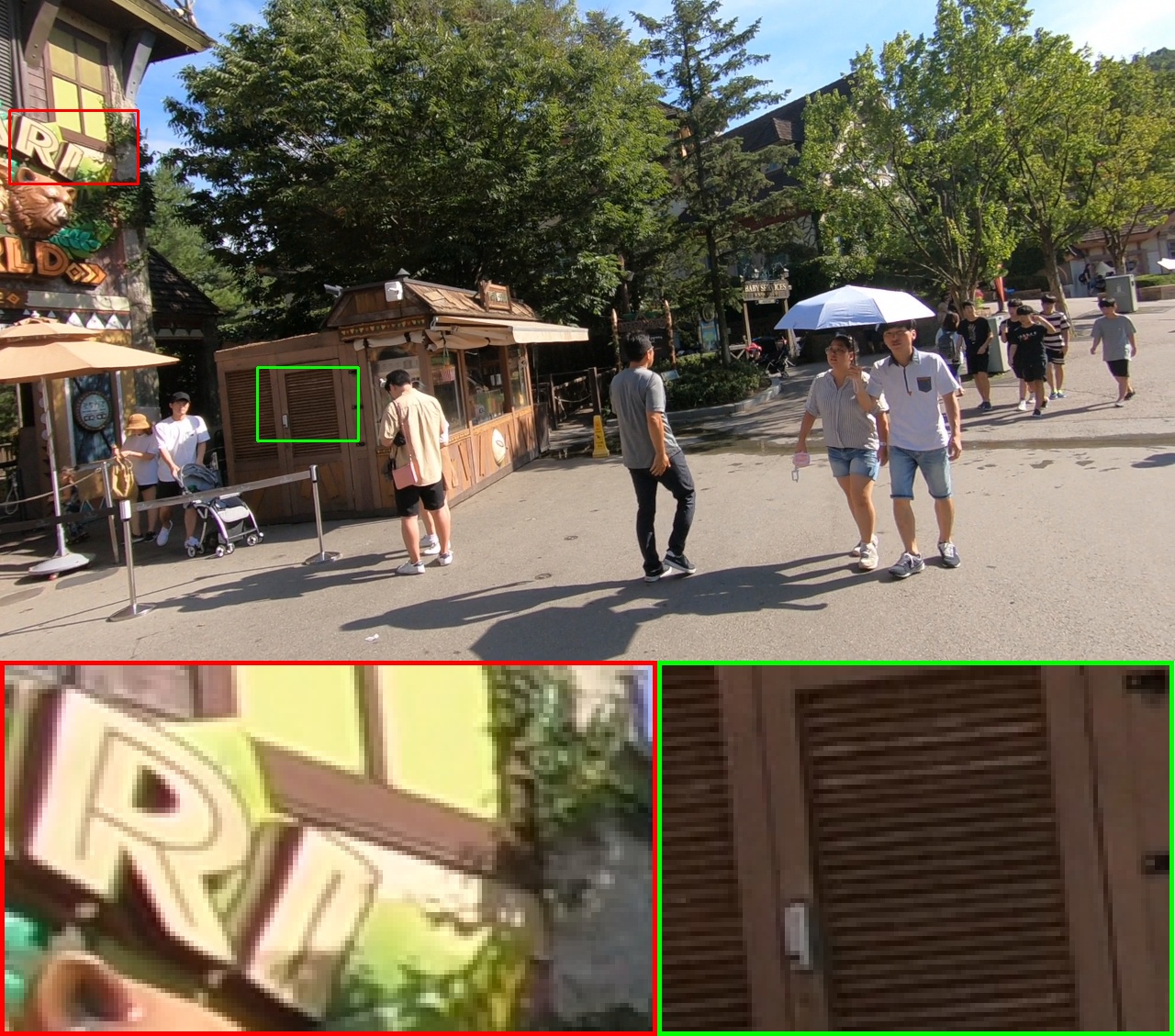}}
      \centerline{Ground Truth}
    \end{minipage}

    \end{minipage}
    \end{center}
    \vspace{-0.6cm}
    \caption{Qualitative comparisons on the validation set of REDS for BID. Please zoom in for better visualization.}
   \vspace{-0.25cm}
    \label{fig_compare2}
  \end{figure*}

\textbf{Ensemble scheme.}\label{subsec:abl_ensemble}
We adopt two types of ensemble schemes to further enhance the performance of EDPN. 
The first one is self-ensemble.
We rotate the input image $90^{\circ}$, $180^{\circ}$ and $270^{\circ}$, and feed them into the network to obtain corresponding outputs. 
Then we average these outputs and the original output as the final result.
The second one is model-ensemble, whose result is the linear combination of several models with different training iterations and loss functions.
Experimental results listed in Table~\ref{table: ensemble_ablation} demonstrate the performance improvement achieved using these two ensemble schemes.

\subsection{Comparisons with Existing Methods}

To validate the effectiveness of the proposed method, we compare our EDPN with the existing methods that can be directly applied to the BISR and BID tasks.
For BISR, we compare with RCAN~\cite{zhang2018image}, MSRN~\cite{li2018multi}, GFN~\cite{zhang2018gated} and EDVR~\cite{wang2019edvr}. 
For BID, we compare with SRN~\cite{tao2018scale}, RNAN~\cite{zhang2019residual}, MPRNet~\cite{zamir2021multi} and EDVR~\cite{wang2019edvr}. 
All these methods are trained using the whole training set of the challenge, and the official validation set is adopted for evaluation.
For EDVR, it takes the same number of input images as EDPN.
The quantitative results are listed in Table~\ref{table_comparison}.
As can be seen, our EDPN significantly outperforms previous methods in PSNR, SSIM and LPIPS metrics.
Compared with EDVR, our EDPN achieves 0.38 dB and 0.77 dB gain in PSNR for BISR and BID, respectively.
The comparison results demonstrate that our EDPN can effectively exploit self- and cross-scale similarities and boost the performance of blurry image restoration from multiple degradations.
In addition, we calculate the number of parameters and the average running time of different methods when the size of input images is $128 \times 128$ using a 1080Ti GPU as shown in Table~\ref{table_comparison}.

\begin{figure*}[!t]
    \fontsize{7}{9.6}\selectfont
    \begin{center}
    \begin{minipage}{0.98\linewidth}

    \begin{minipage}{0.152\linewidth}
      \centerline{\includegraphics[width=1.1\linewidth]{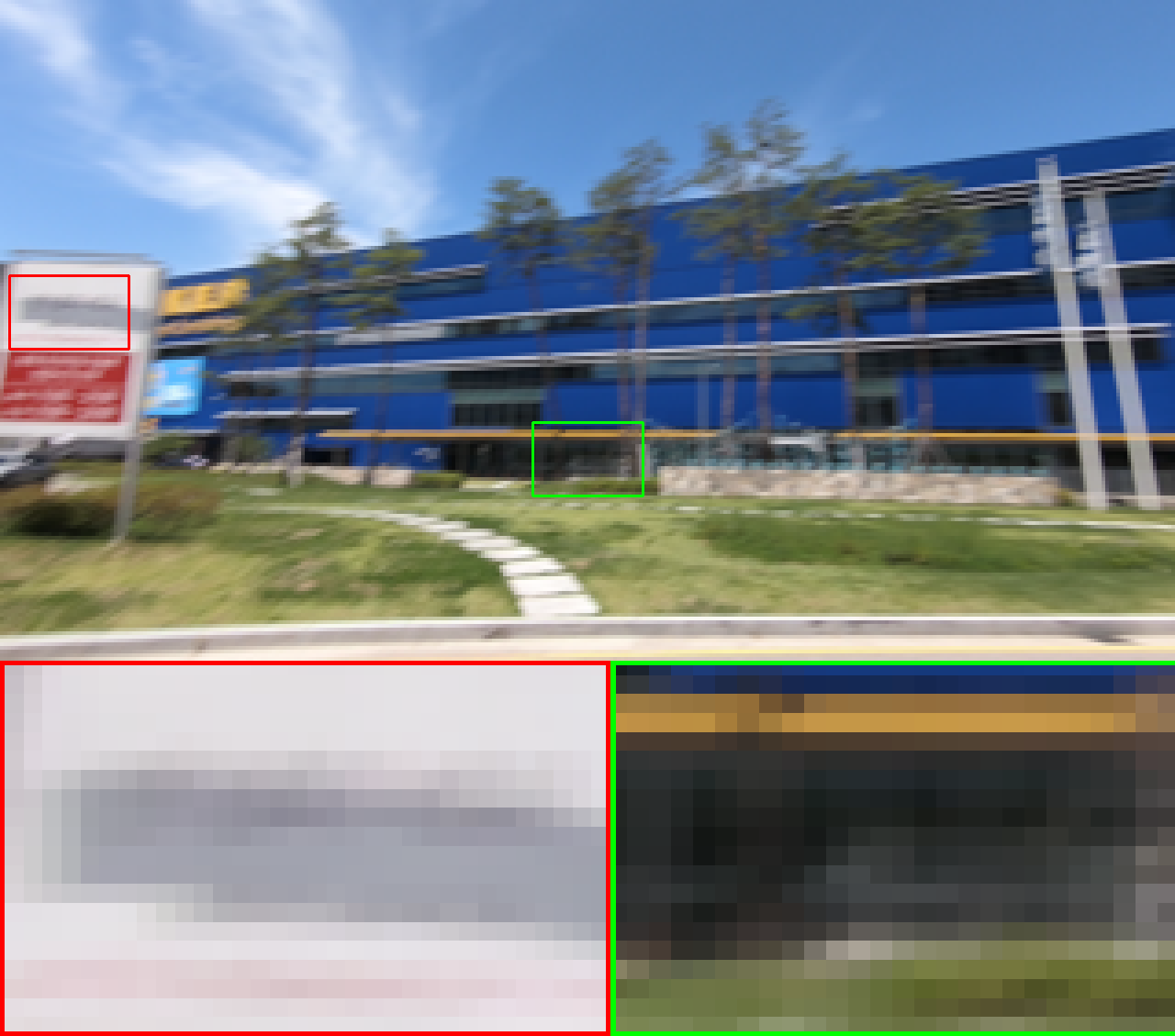}}
    \end{minipage}
  \hfill
   \begin{minipage}{0.152\linewidth}
      \centerline{\includegraphics[width=1.1\linewidth]{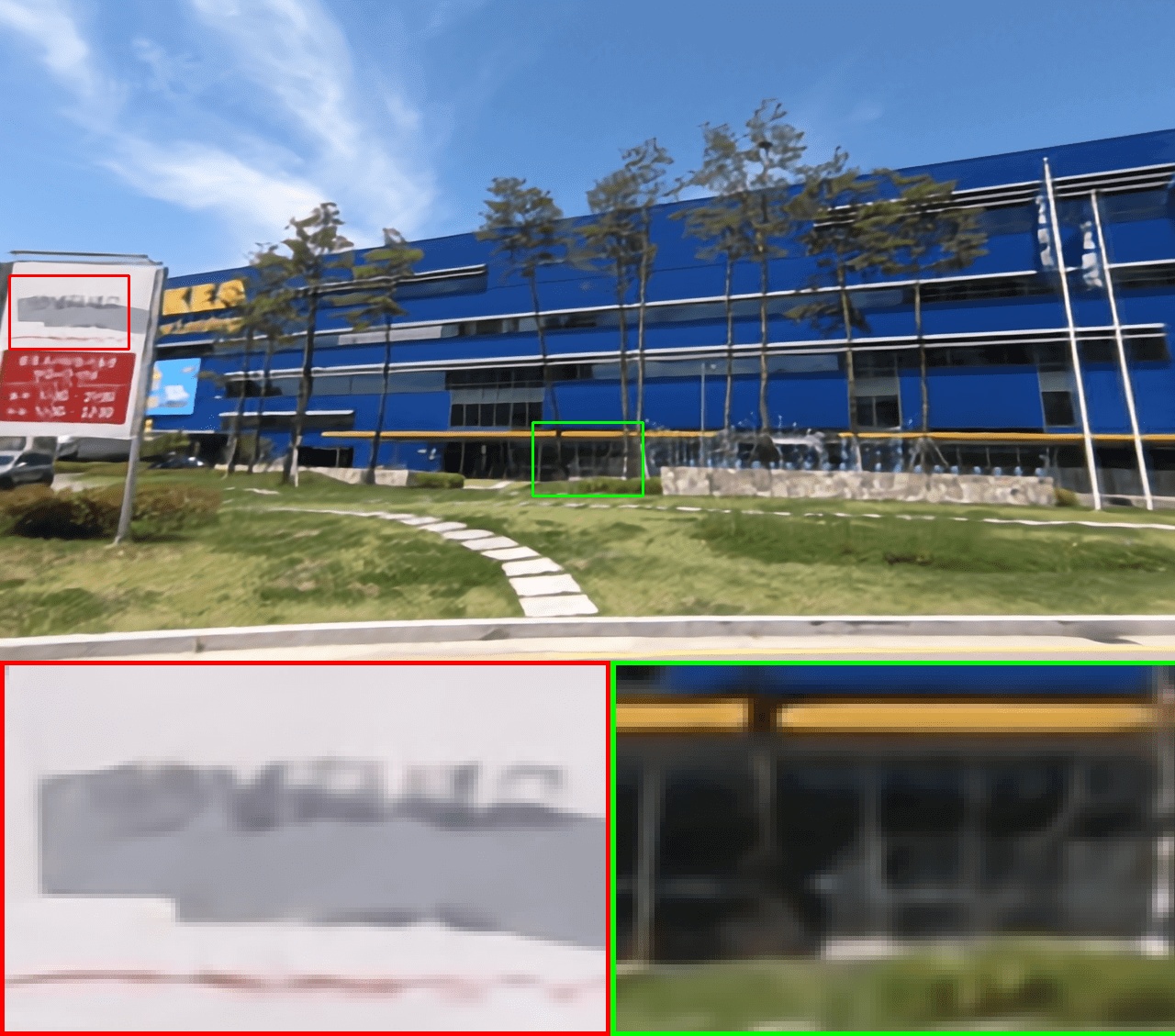}}
    \end{minipage}
  \hfill
    \begin{minipage}{0.152\linewidth}
      \centerline{\includegraphics[width=1.1\linewidth]{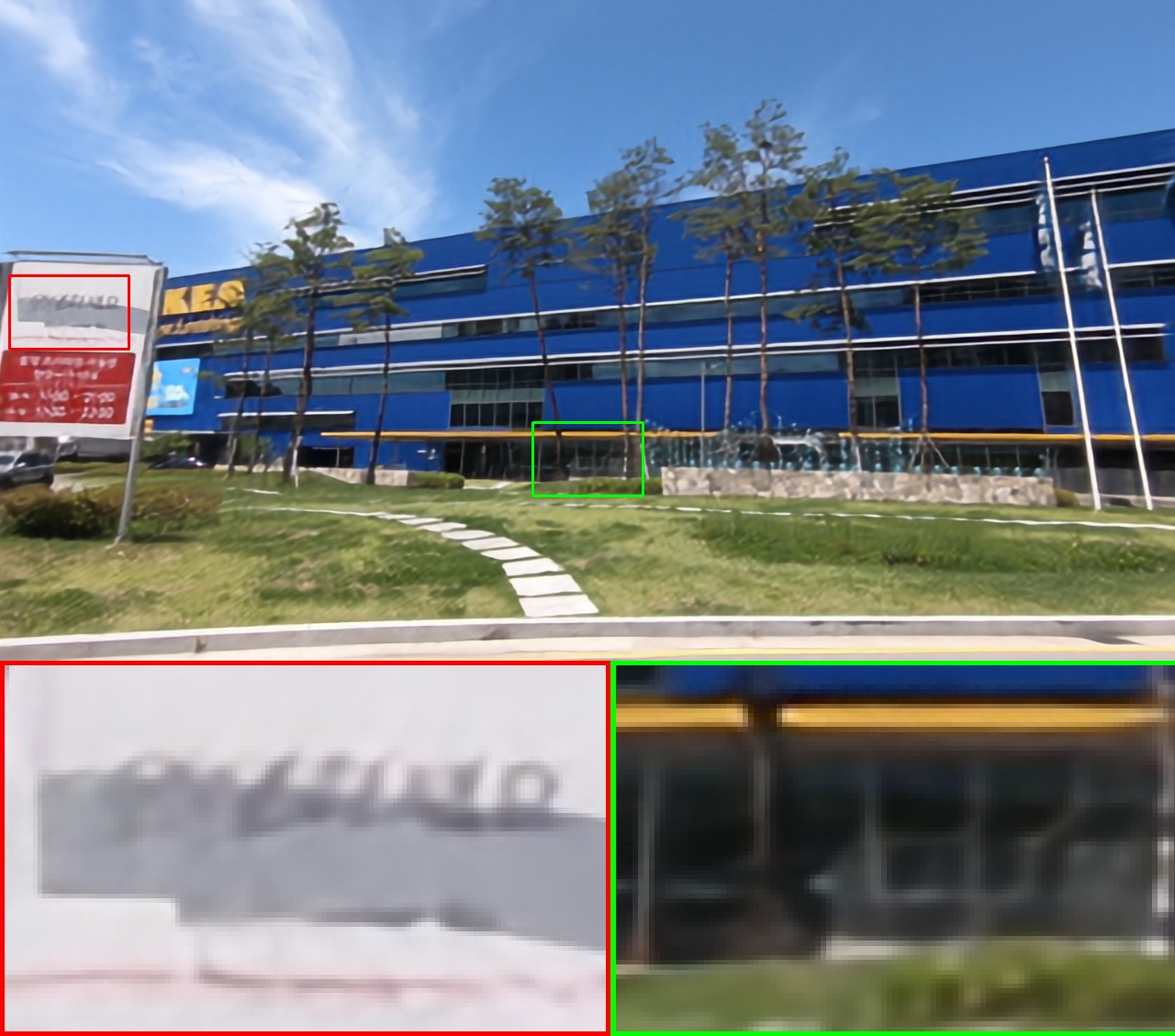}}
    \end{minipage}
  \hfill
    \begin{minipage}{0.152\linewidth}
      \centerline{\includegraphics[width=1.1\linewidth]{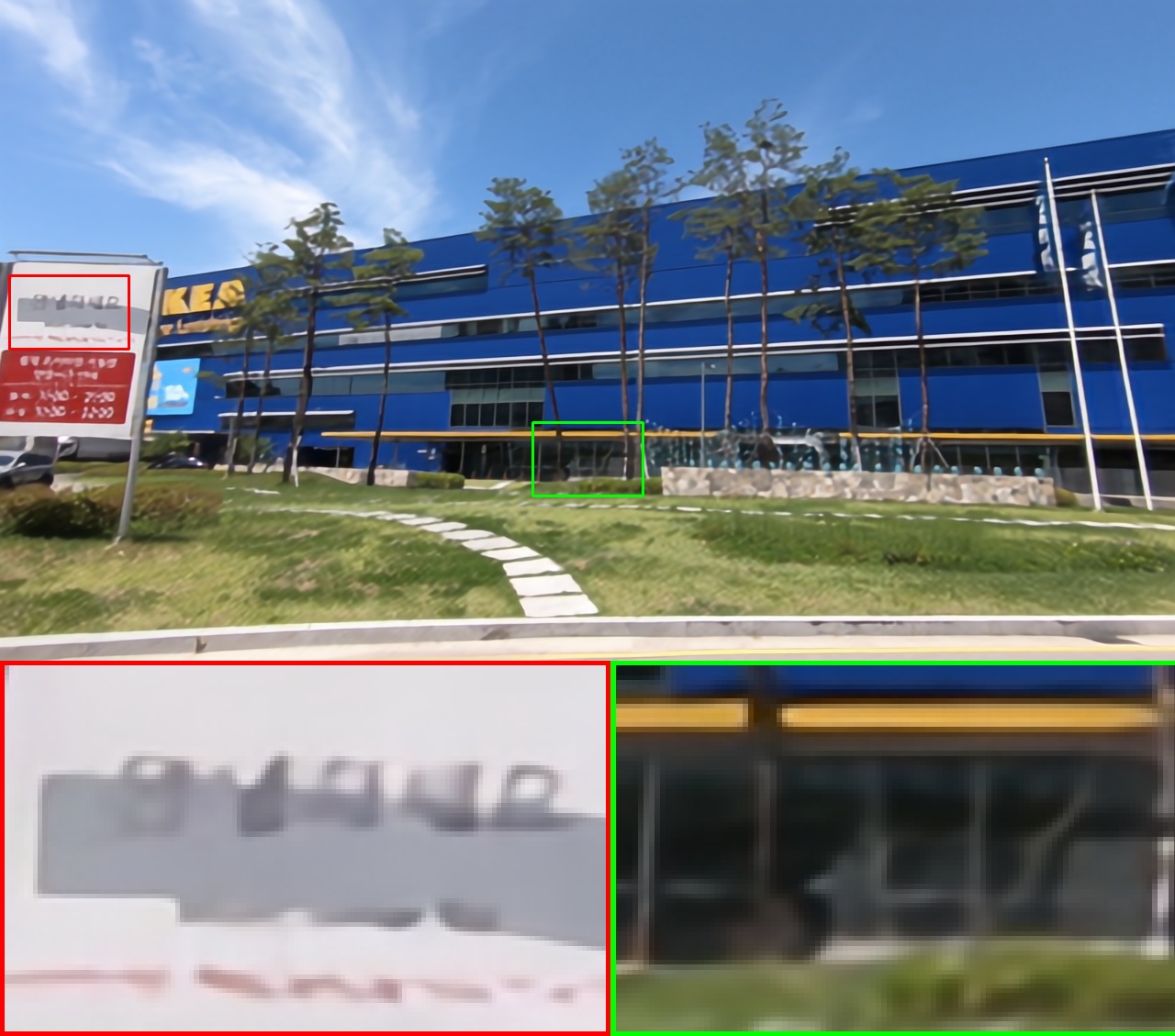}}
    \end{minipage}
  \hfill
  \begin{minipage}{0.152\linewidth}
      \centerline{\includegraphics[width=1.1\linewidth]{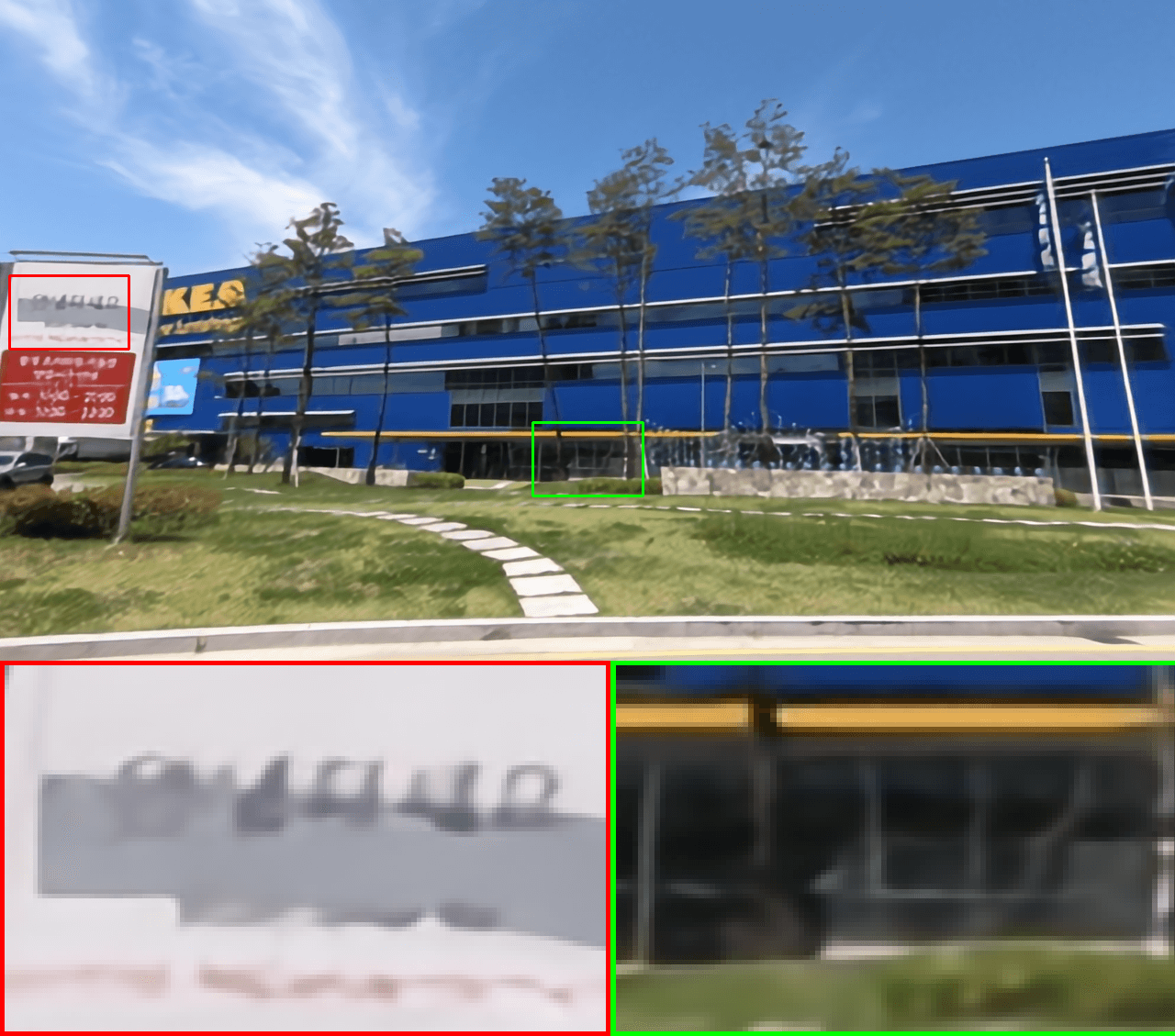}}
    \end{minipage}
  \hfill
    \begin{minipage}{0.152\linewidth}
      \centerline{\includegraphics[width=1.1\linewidth]{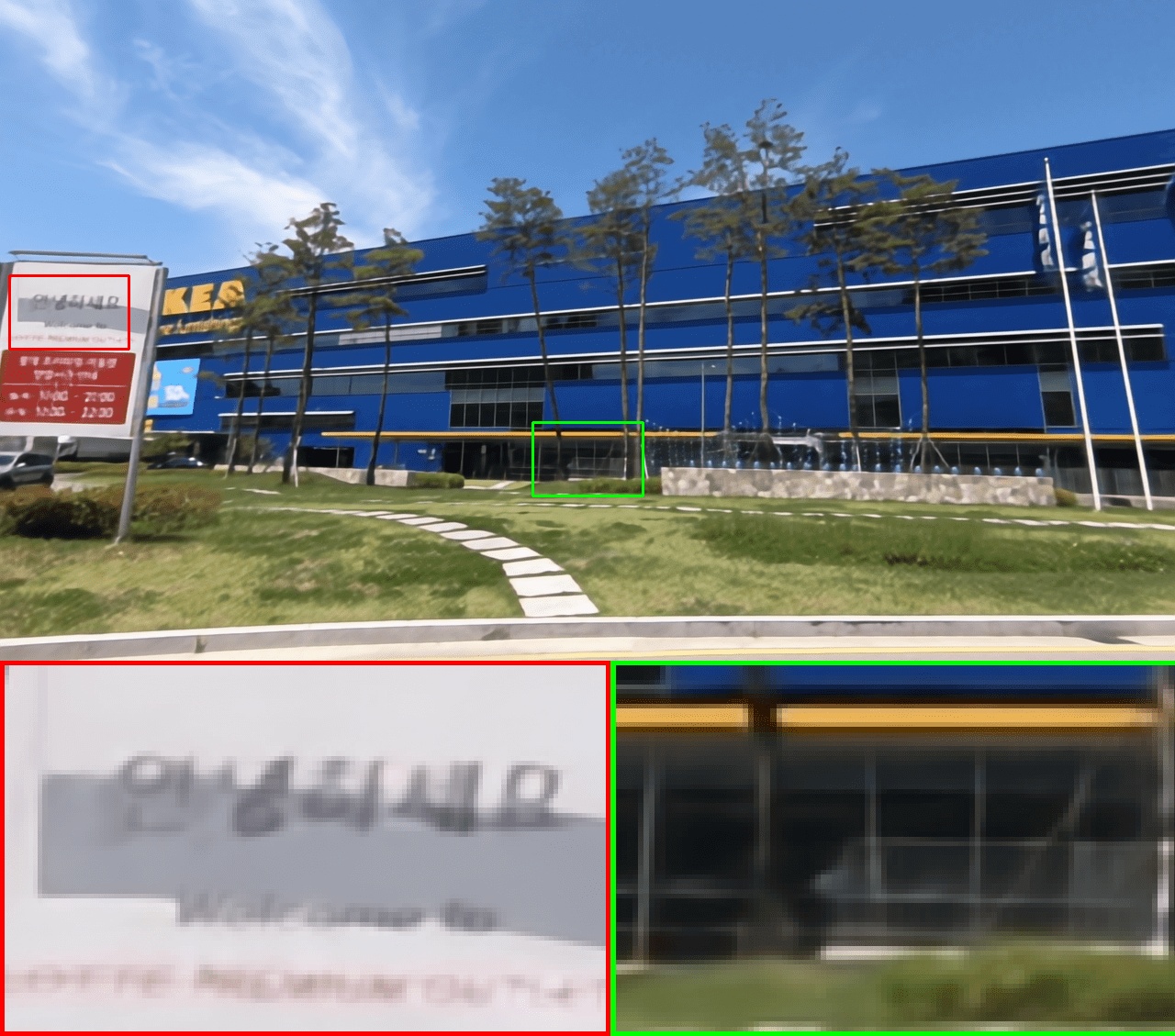}}
    \end{minipage}

  \vfill

    \begin{minipage}{0.152\linewidth}
      \centerline{\includegraphics[width=1.1\linewidth]{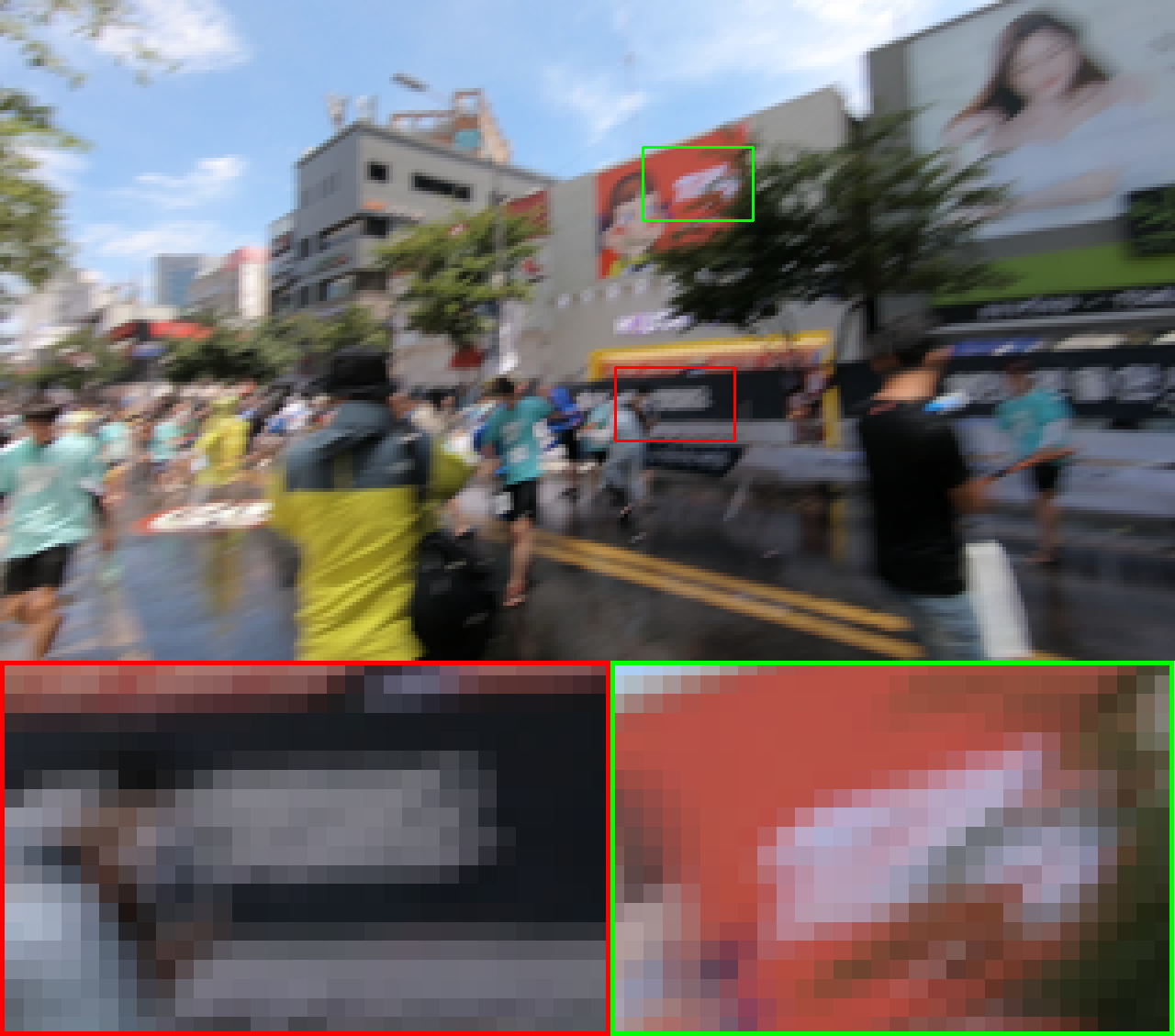}}
      \centerline{Input}
    \end{minipage}
  \hfill 
  \begin{minipage}{0.152\linewidth}
      \centerline{\includegraphics[width=1.1\linewidth]{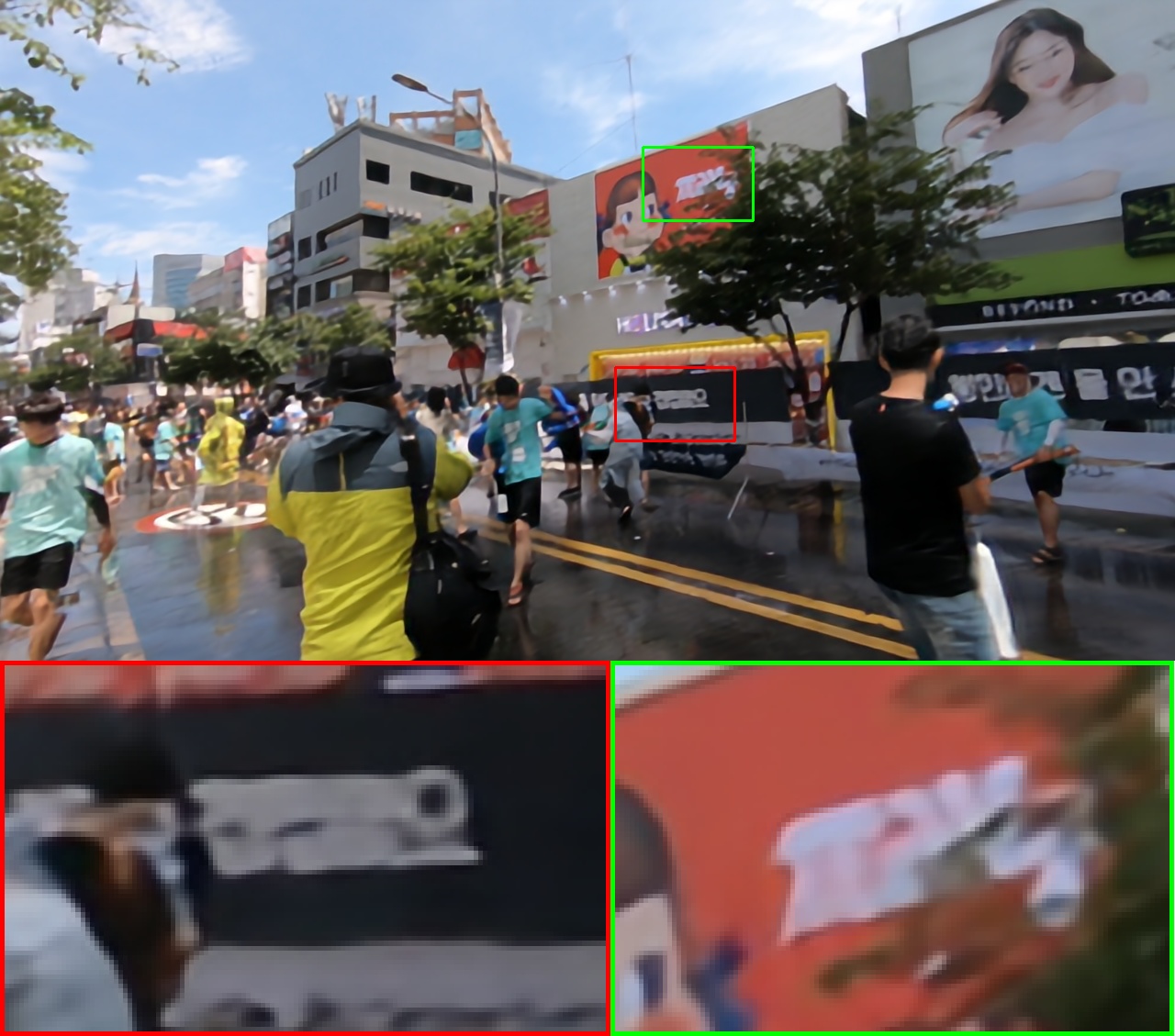}}
      \centerline{MSRN}
    \end{minipage}
  \hfill
    \begin{minipage}{0.152\linewidth}
      \centerline{\includegraphics[width=1.1\linewidth]{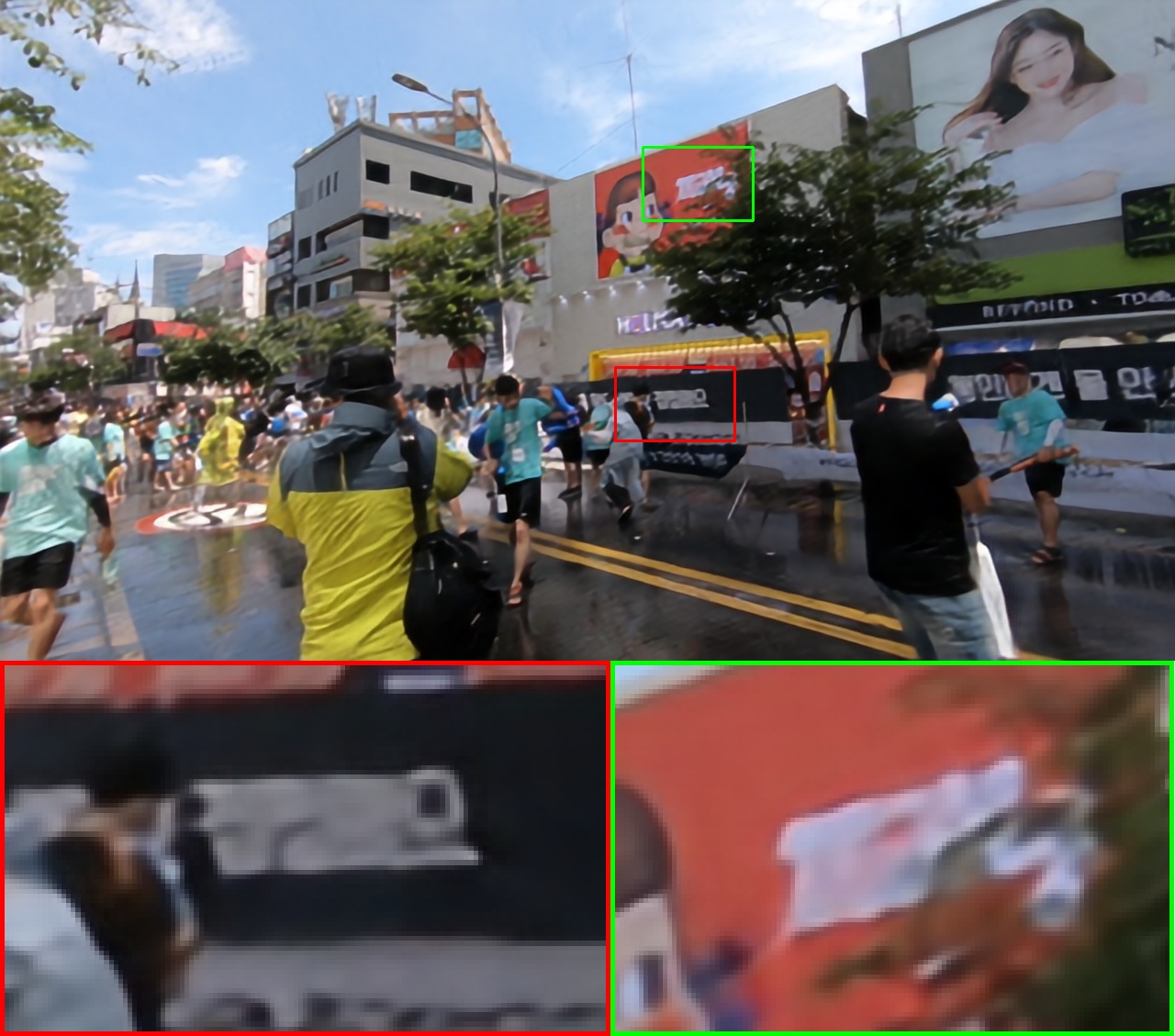}}
      \centerline{GFN}
    \end{minipage}
  \hfill
    \begin{minipage}{0.152\linewidth}
      \centerline{\includegraphics[width=1.1\linewidth]{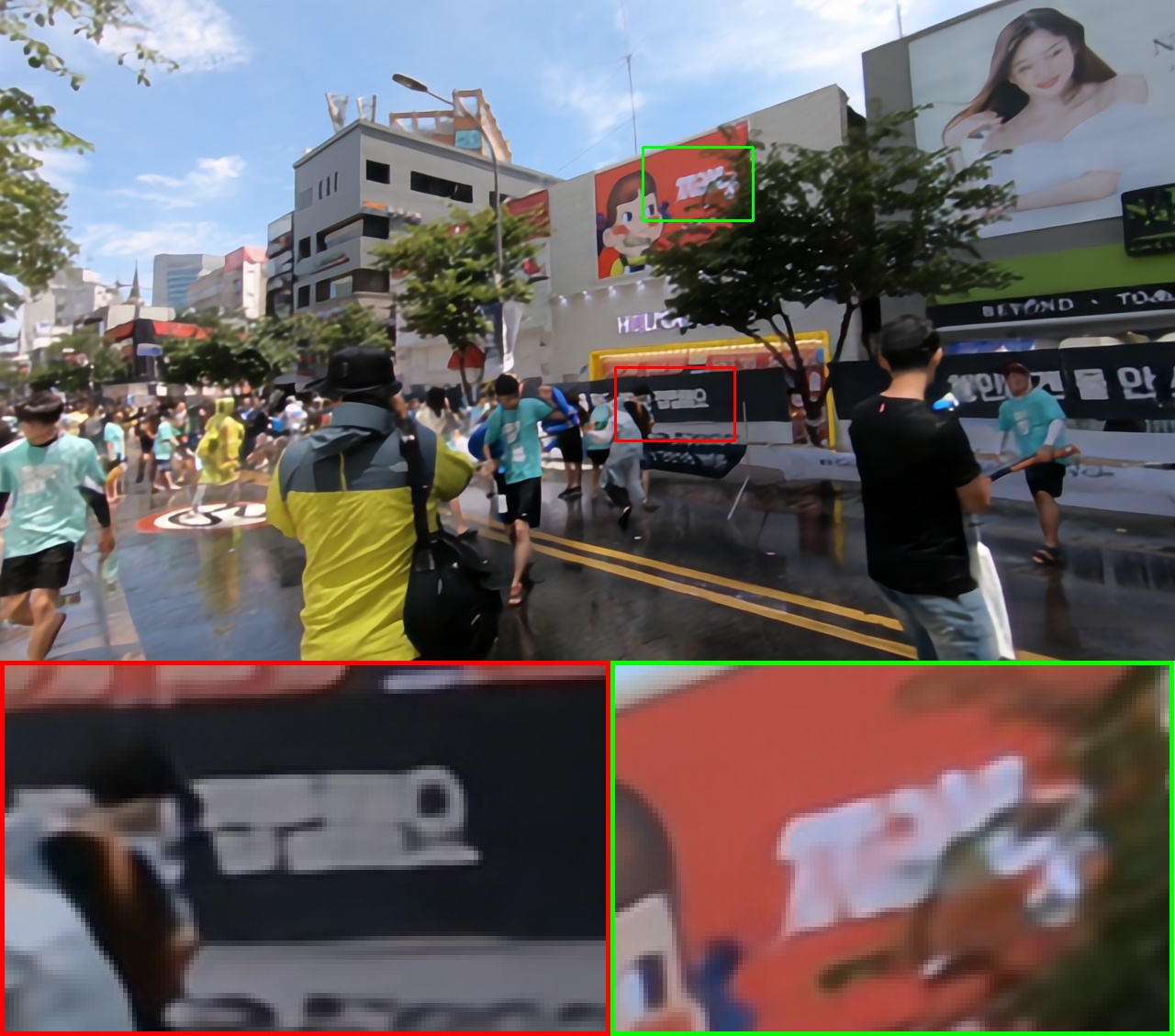}}
      \centerline{RCAN}
    \end{minipage}
  \hfill
  \begin{minipage}{0.152\linewidth}
      \centerline{\includegraphics[width=1.1\linewidth]{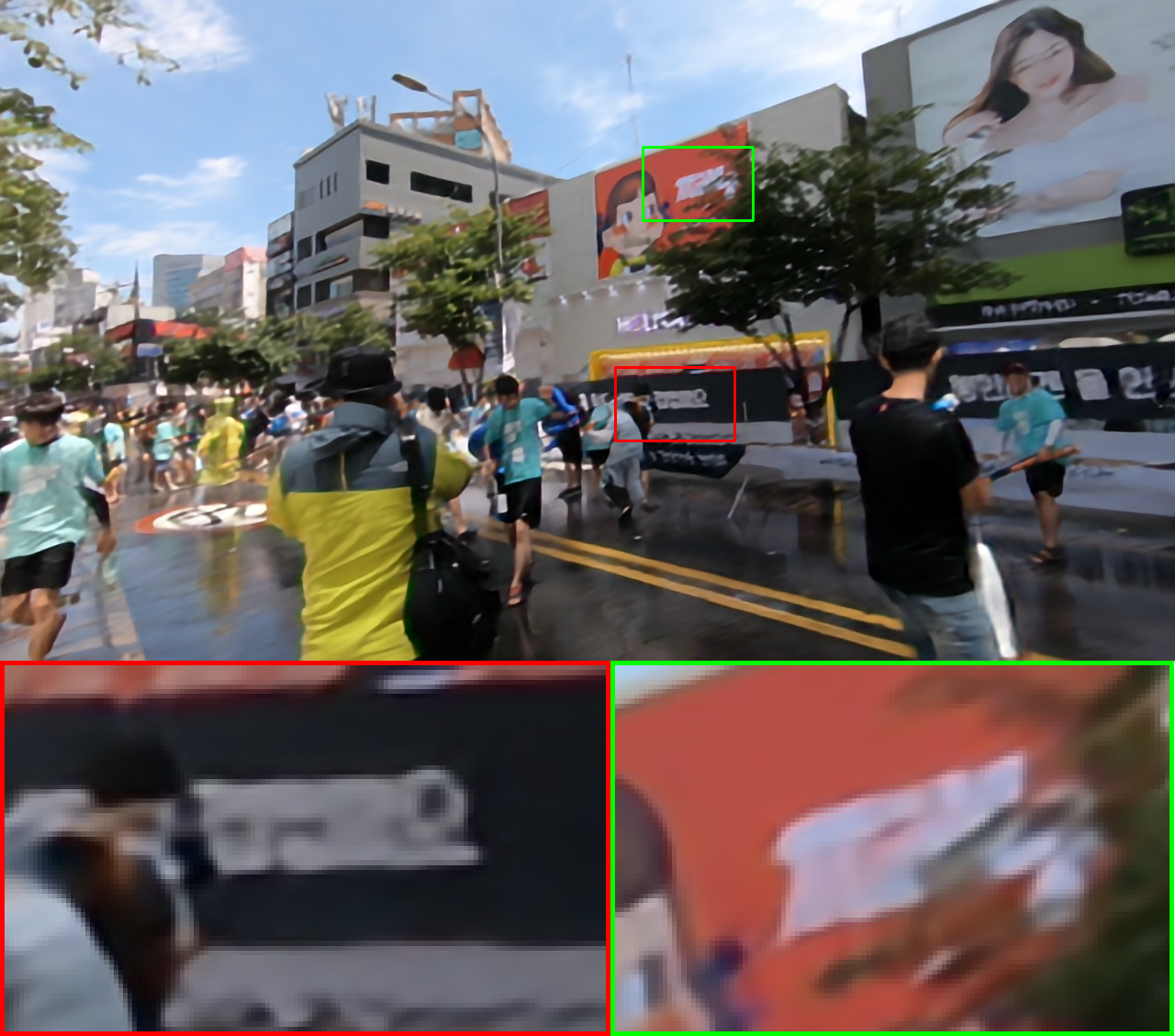}}
      \centerline{EDVR}
    \end{minipage}
  \hfill
    \begin{minipage}{0.152\linewidth}
      \centerline{\includegraphics[width=1.1\linewidth]{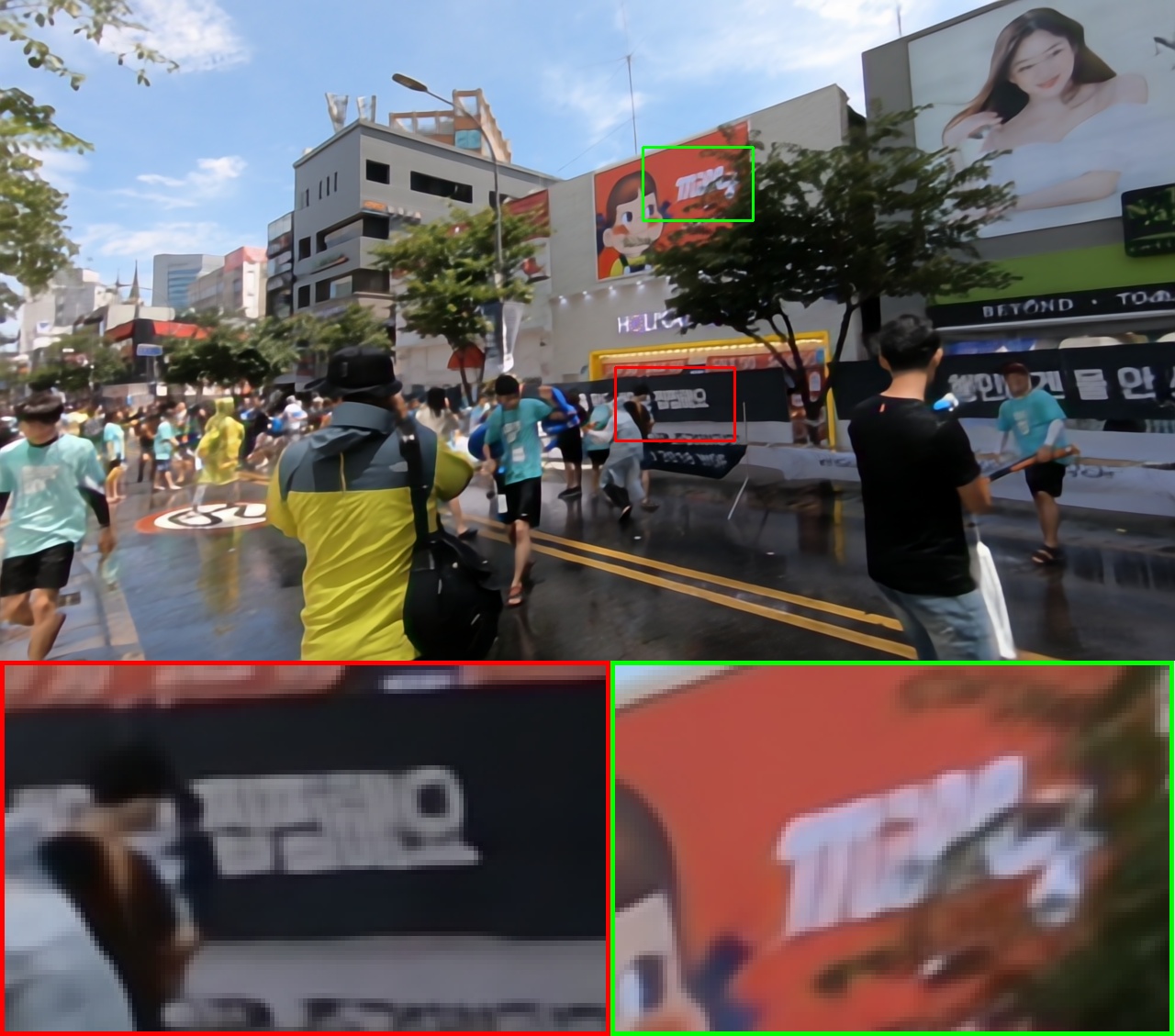}}
      \centerline{\textbf{EDPN (Ours)}}
    \end{minipage}

    \end{minipage}
    \end{center}
    \vspace{-0.5cm}
    \caption{Qualitative comparisons on the test set of REDS for Track 1 (Low Resolution). Please zoom in for better visualization.}
    \label{fig_compare3}
  \end{figure*}

\begin{figure*}[!t]
    \fontsize{7}{9.6}\selectfont
    \begin{center}
    \begin{minipage}{0.98\linewidth}

    \begin{minipage}{0.152\linewidth}
      \centerline{\includegraphics[width=1.1\linewidth]{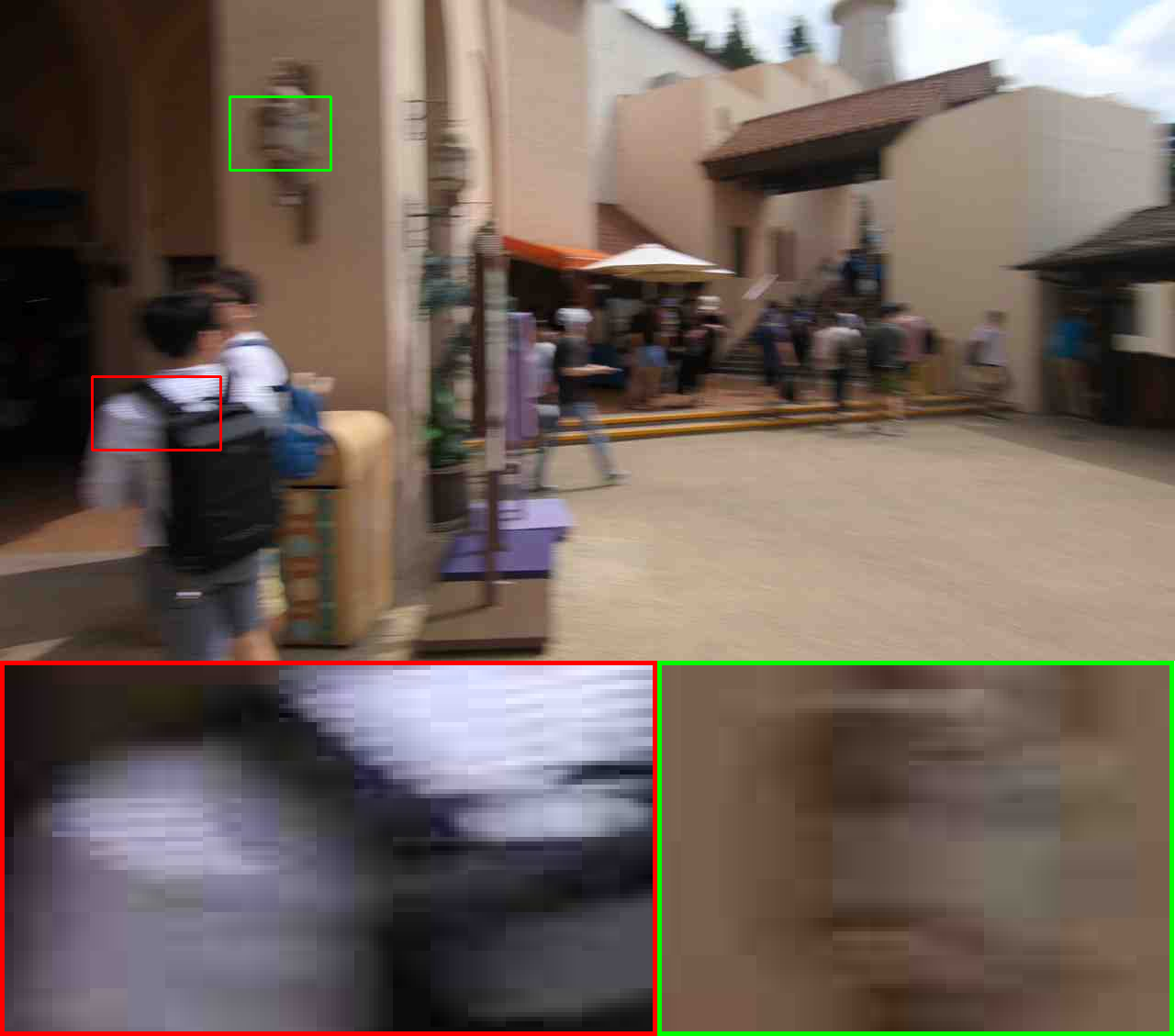}}
      \centerline{Input}
    \end{minipage}
  \hfill   
    \begin{minipage}{0.152\linewidth}
      \centerline{\includegraphics[width=1.1\linewidth]{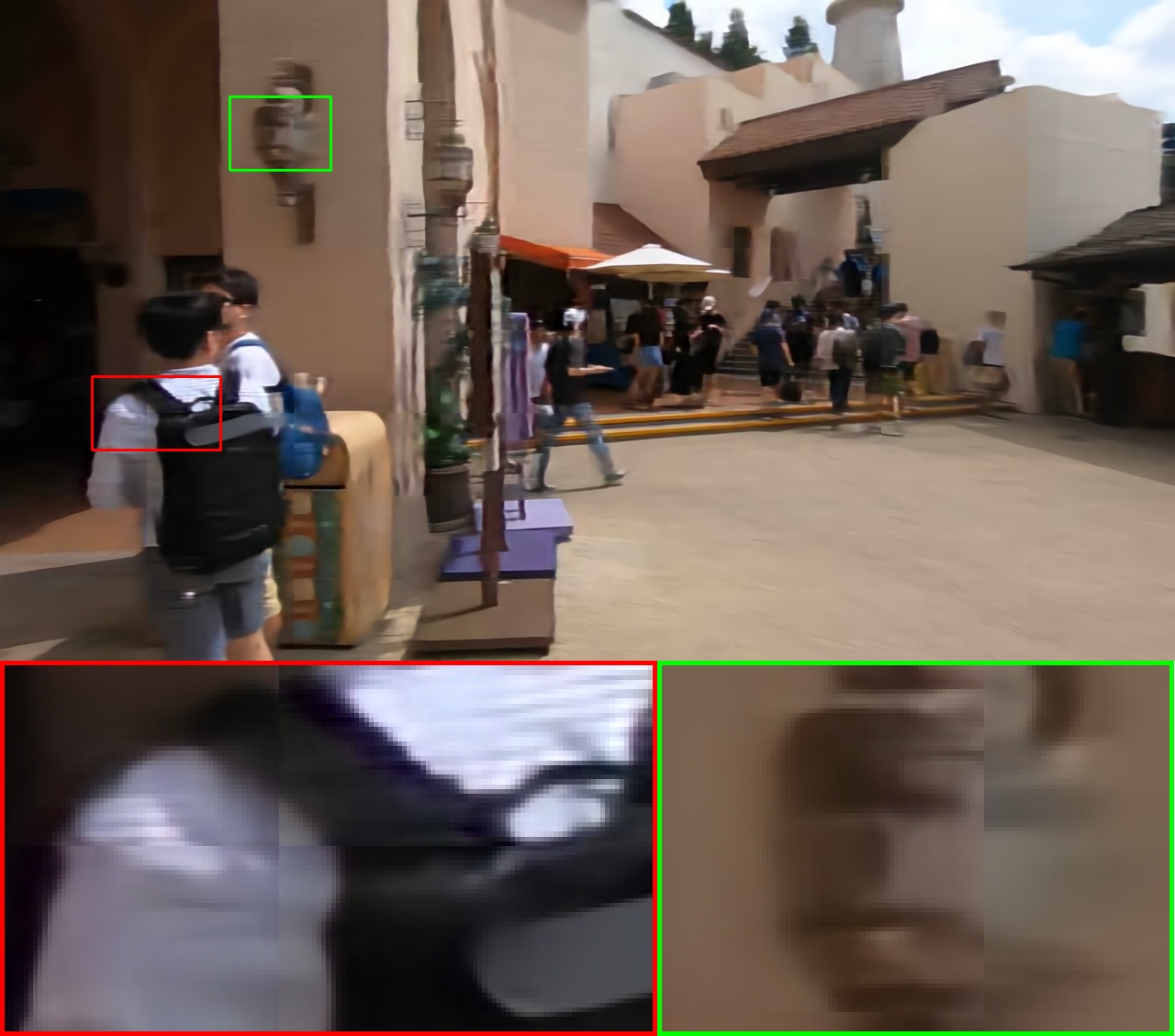}}
      \centerline{RNAN}
    \end{minipage}
  \hfill
    \begin{minipage}{0.152\linewidth}
      \centerline{\includegraphics[width=1.1\linewidth]{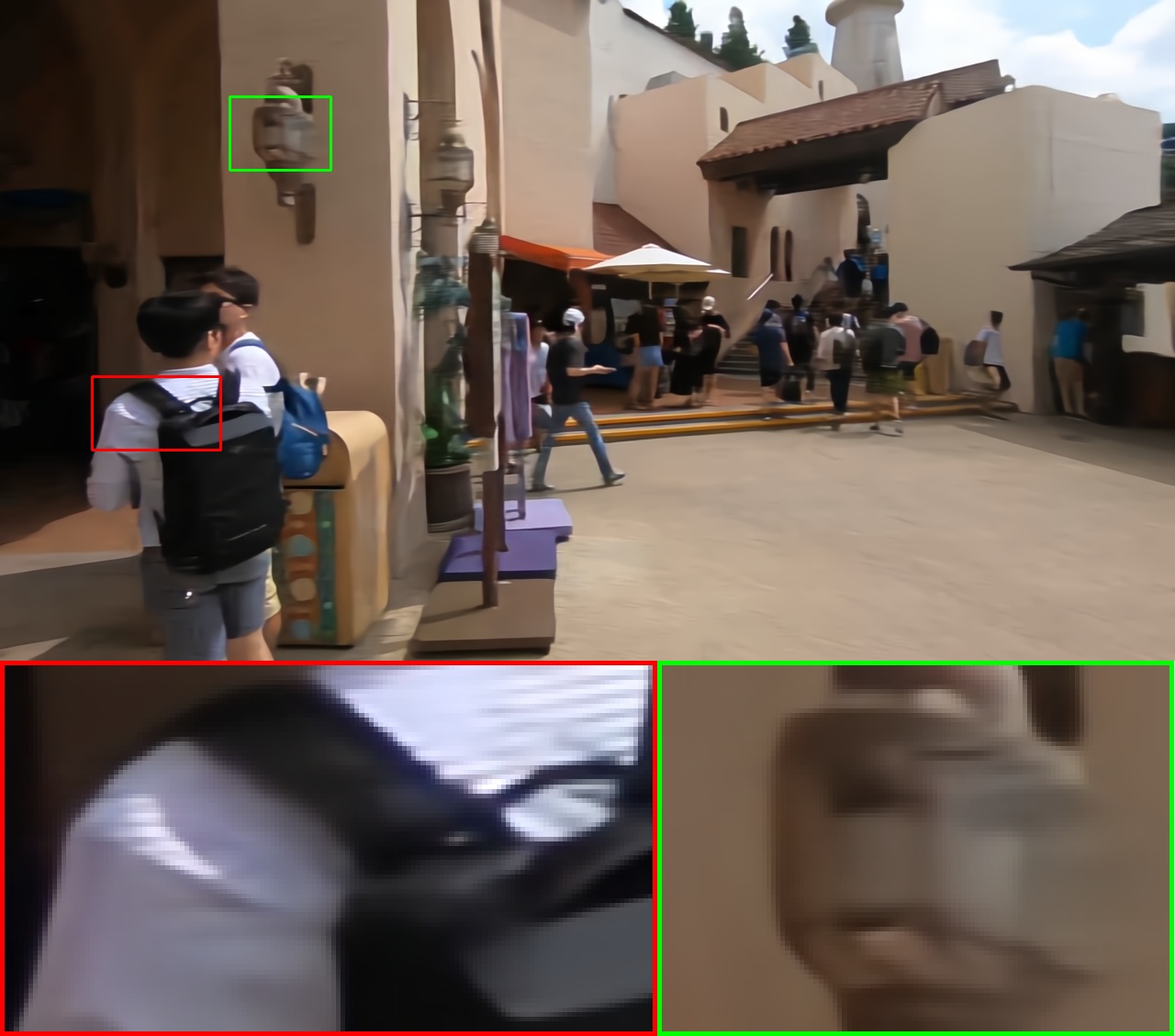}}
      \centerline{MPRNet}
    \end{minipage}
  \hfill
    \begin{minipage}{0.152\linewidth}
      \centerline{\includegraphics[width=1.1\linewidth]{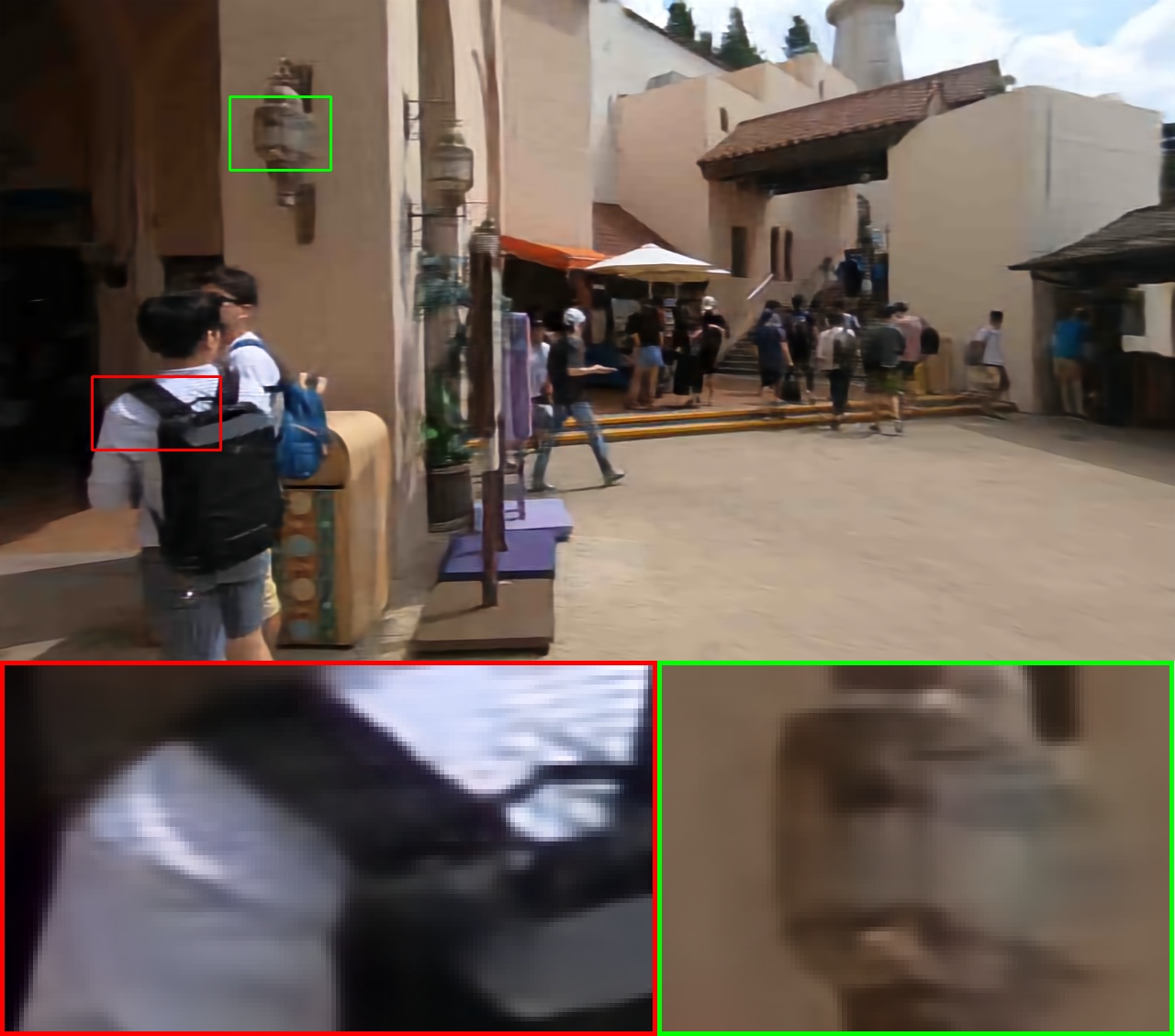}}
      \centerline{SRN}
    \end{minipage}
  \hfill
  \begin{minipage}{0.152\linewidth}
      \centerline{\includegraphics[width=1.1\linewidth]{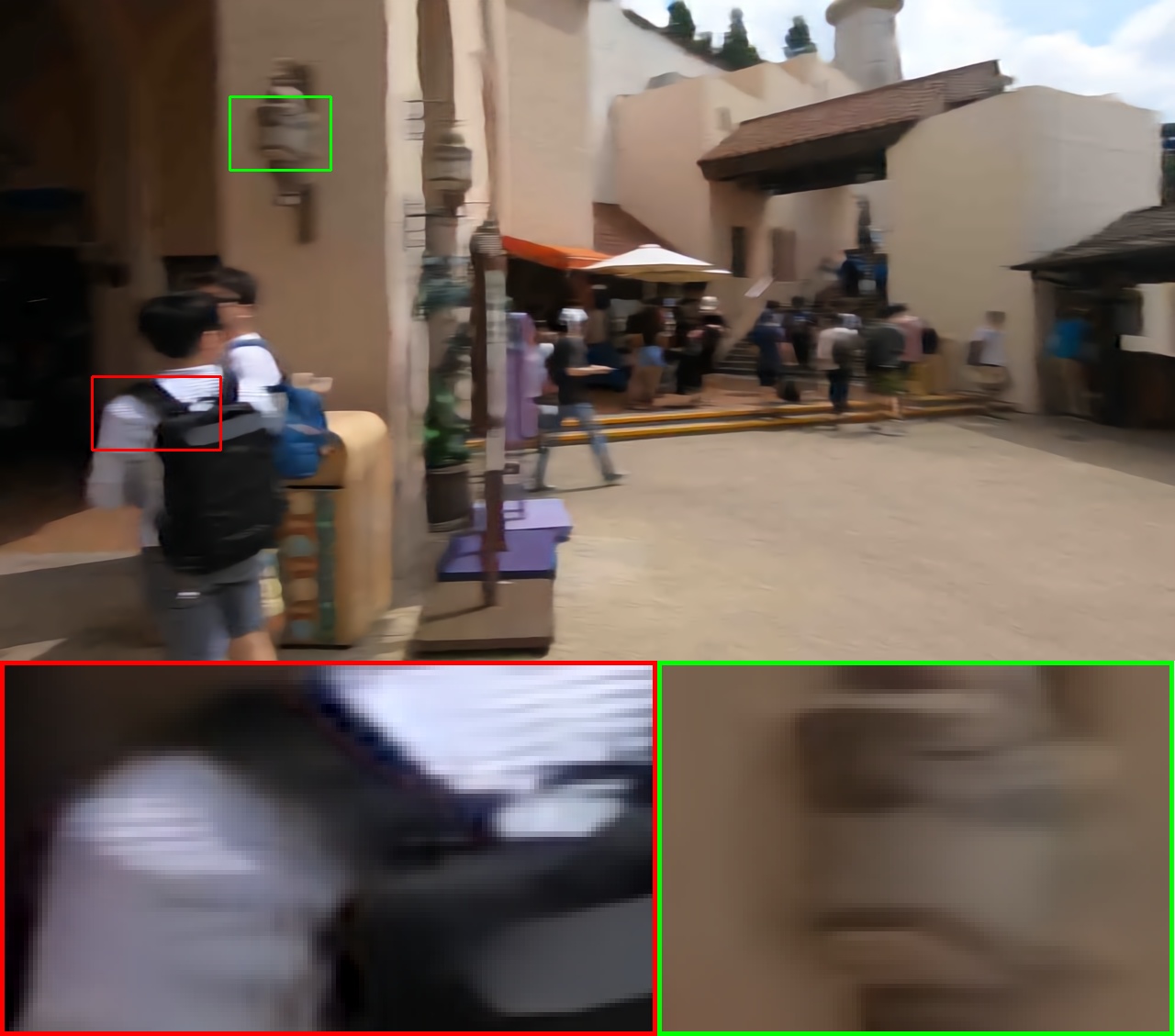}}
      \centerline{EDVR}
    \end{minipage}
  \hfill
    \begin{minipage}{0.152\linewidth}
      \centerline{\includegraphics[width=1.1\linewidth]{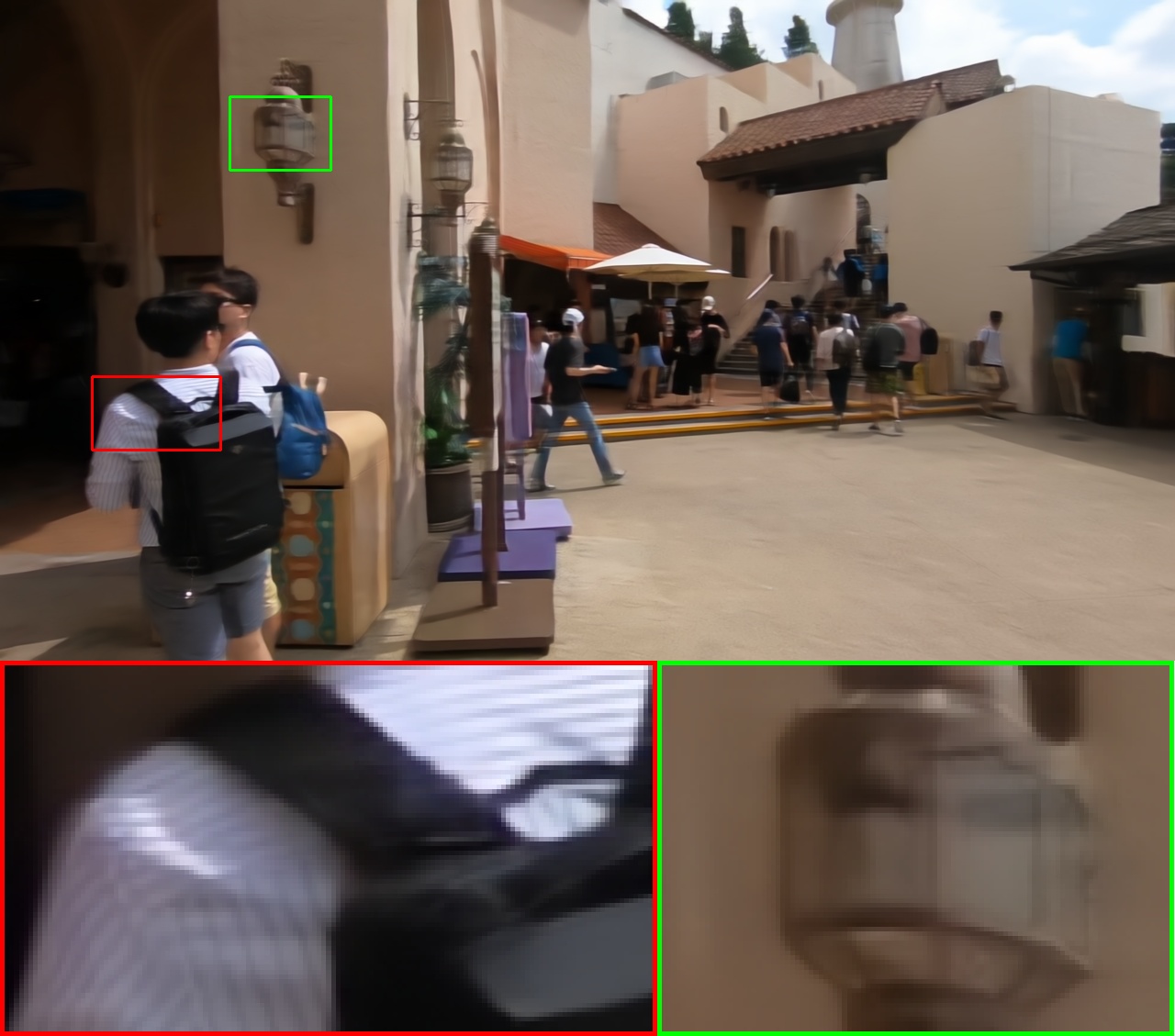}}
      \centerline{\textbf{EDPN (Ours)}}
    \end{minipage}

  \vfill

   \begin{minipage}{0.152\linewidth}
      \centerline{\includegraphics[width=1.1\linewidth]{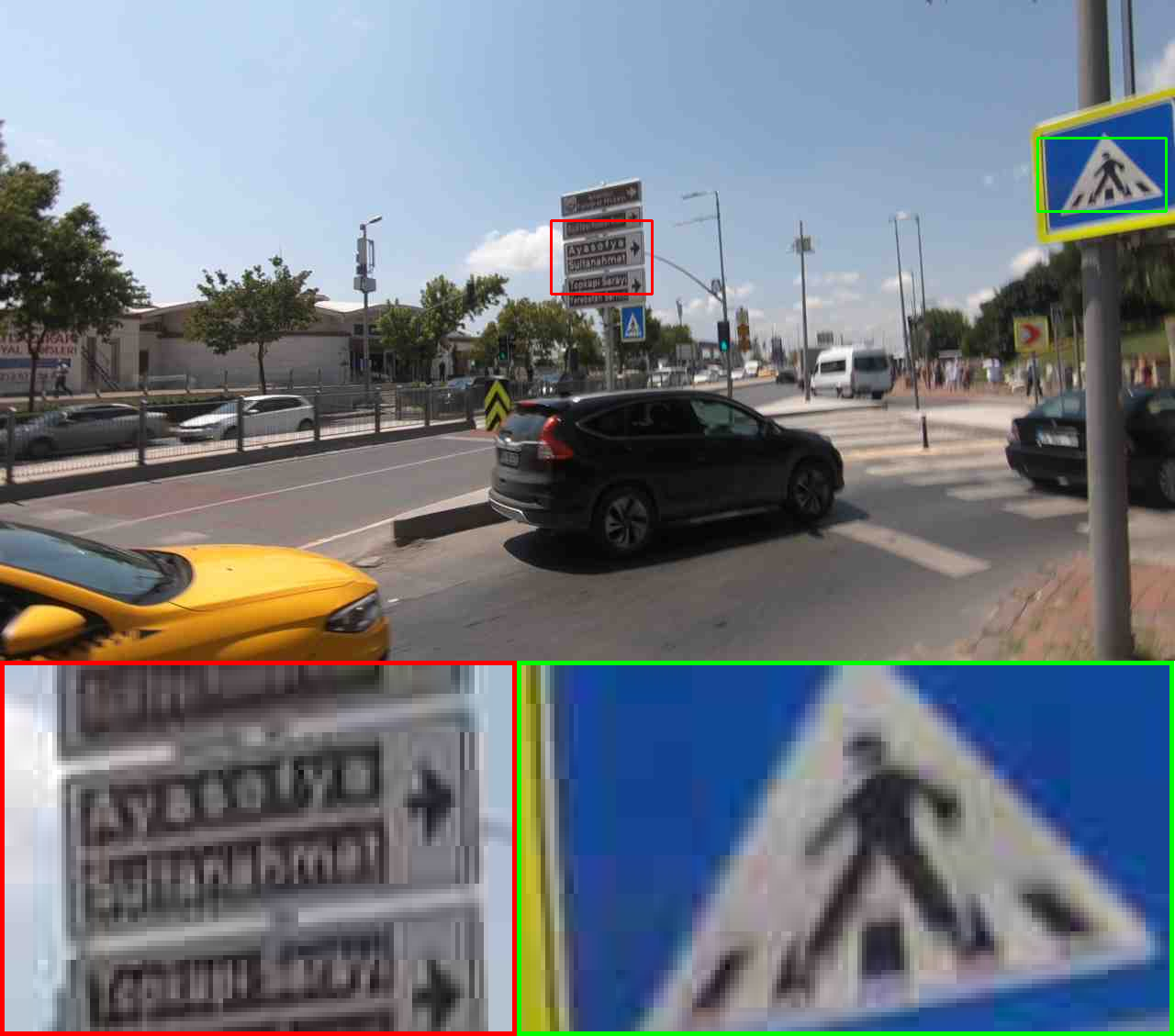}}
      \centerline{Input}
    \end{minipage}
  \hfill   
    \begin{minipage}{0.152\linewidth}
      \centerline{\includegraphics[width=1.1\linewidth]{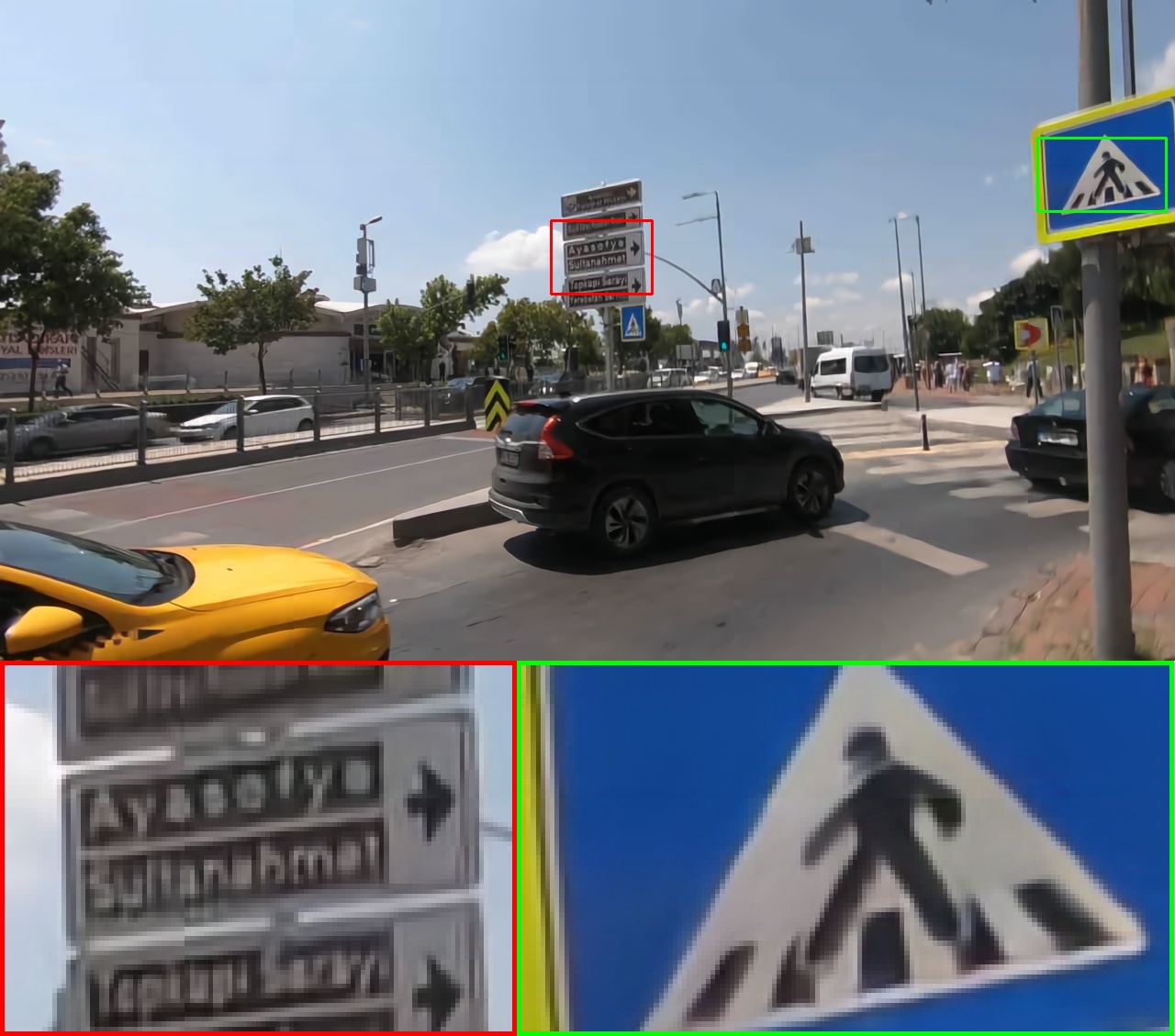}}
      \centerline{RNAN}
    \end{minipage}
  \hfill
    \begin{minipage}{0.152\linewidth}
      \centerline{\includegraphics[width=1.1\linewidth]{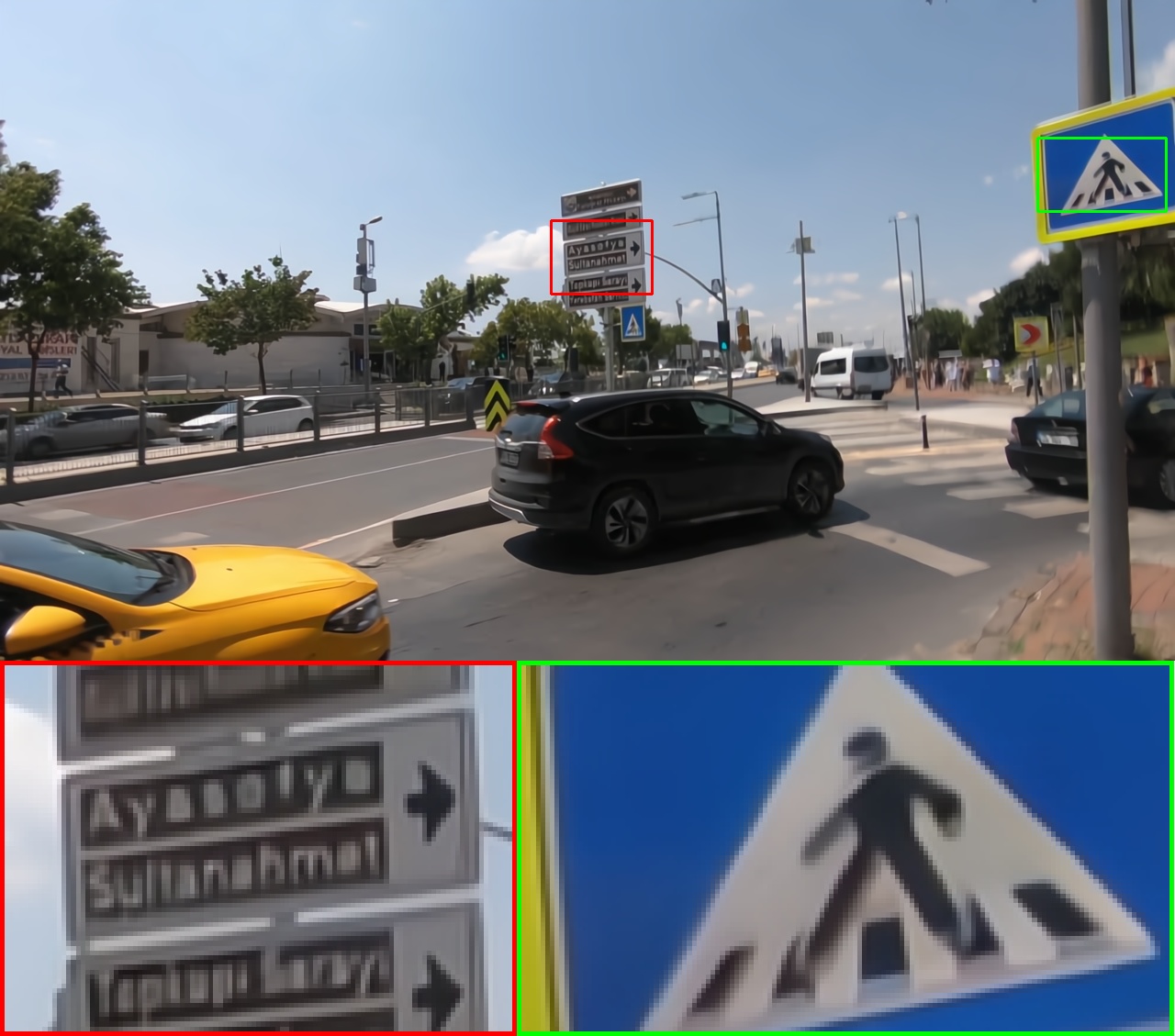}}
      \centerline{MPRNet}
    \end{minipage}
  \hfill
    \begin{minipage}{0.152\linewidth}
      \centerline{\includegraphics[width=1.1\linewidth]{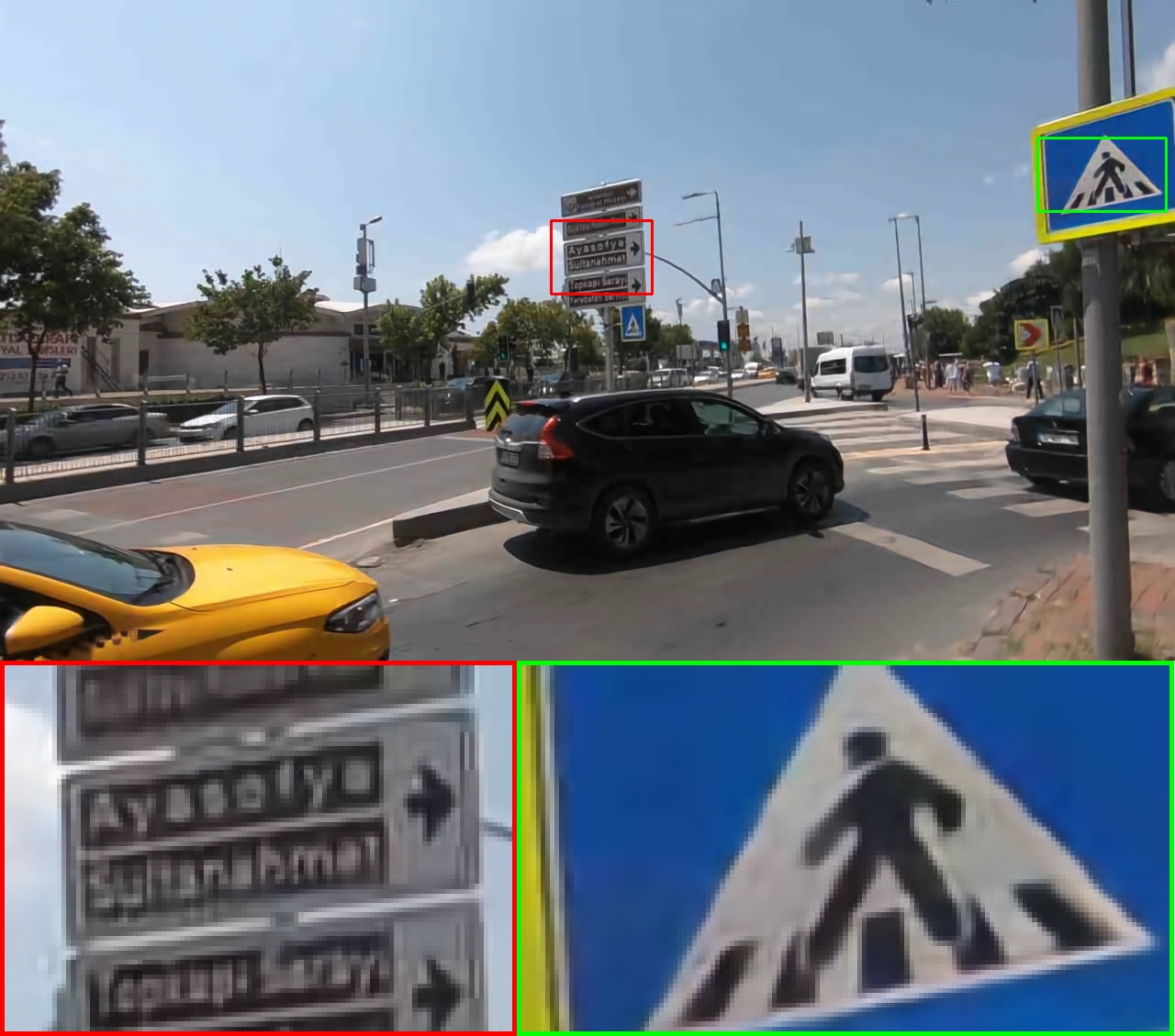}}
      \centerline{SRN}
    \end{minipage}
  \hfill
  \begin{minipage}{0.152\linewidth}
      \centerline{\includegraphics[width=1.1\linewidth]{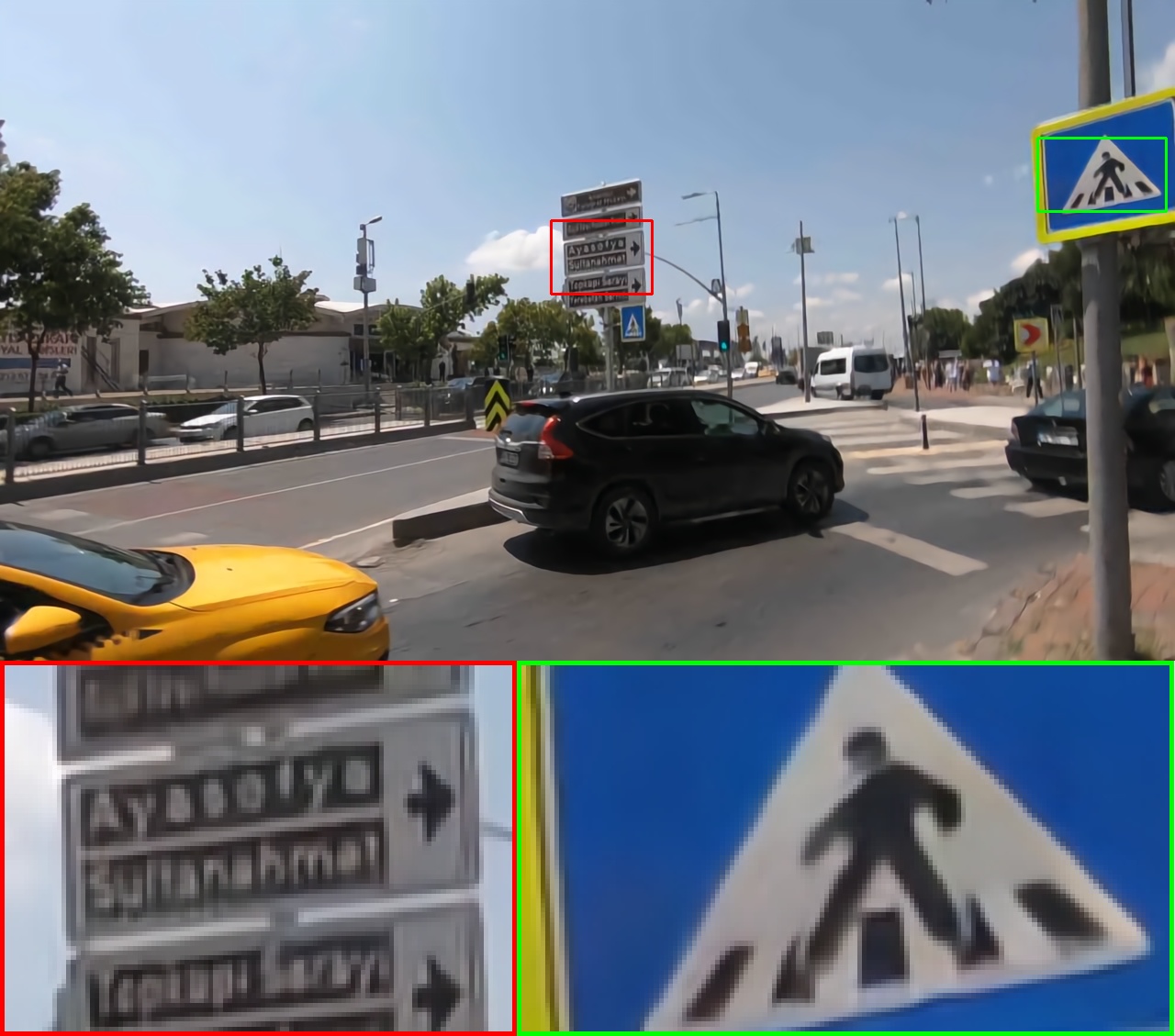}}
      \centerline{EDVR}
    \end{minipage}
  \hfill
    \begin{minipage}{0.152\linewidth}
      \centerline{\includegraphics[width=1.1\linewidth]{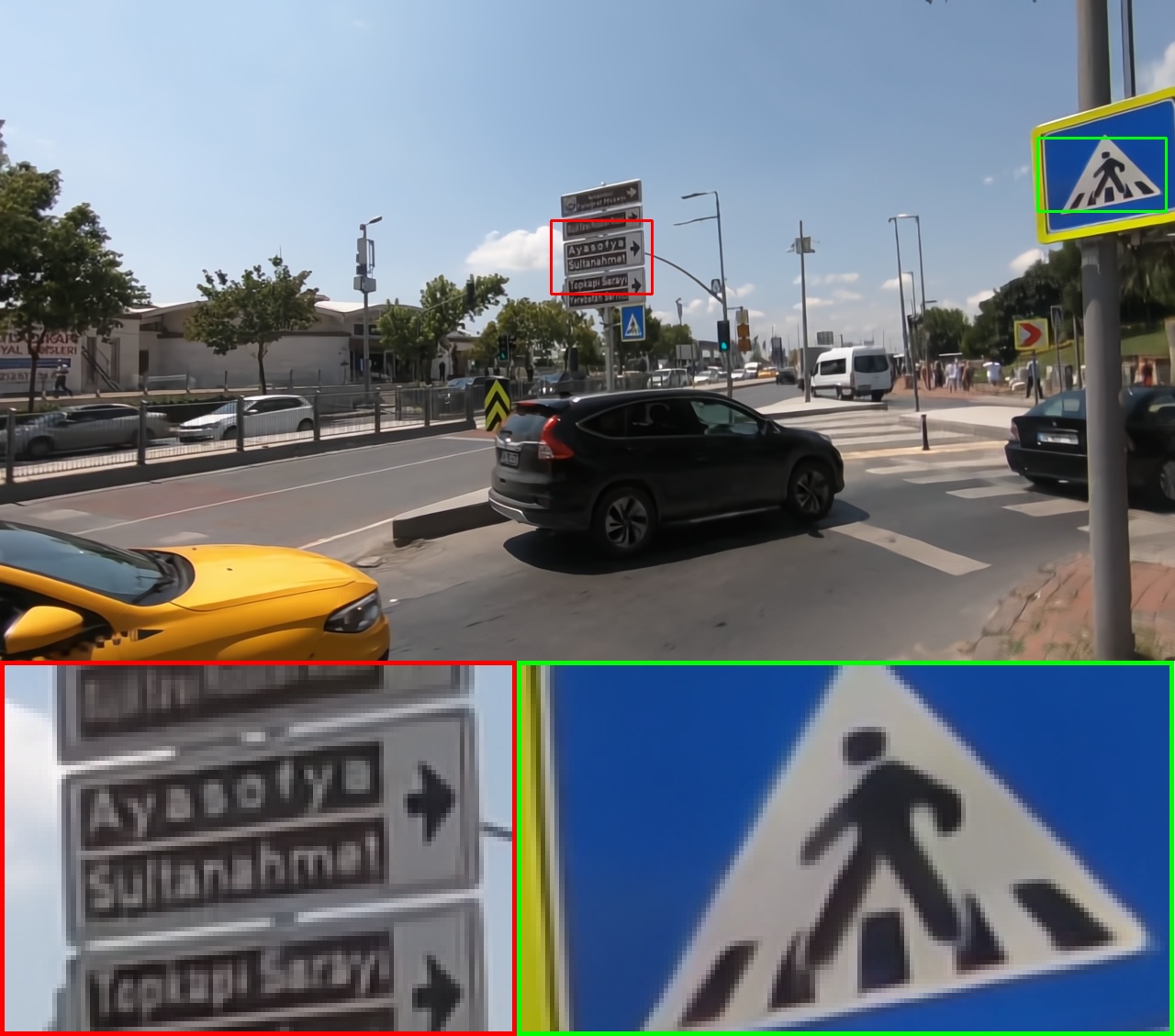}}
      \centerline{\textbf{EDPN (Ours)}}
    \end{minipage}

    \end{minipage}
    \end{center}
    \vspace{-0.6cm}
    \caption{Qualitative comparisons on the test set of REDS for Track 2 (JPEG Artifacts). Please zoom in for better visualization.}
   \vspace{-0.25cm}
    \label{fig_compare4}
  \end{figure*}

To evaluate the perceptual quality, we show two examples of restored results in Figure~\ref{fig_compare1} and Figure~\ref{fig_compare2} for the BISR and BID tasks on the REDS validation set, respectively.
Two more examples on the REDS test set are given in Figure~\ref{fig_compare3} and Figure~\ref{fig_compare4}.
It can be seen that our EDPN provides better qualitative results than other methods with more accurate details in both tasks.
Specifically, the edge regions of the restored images from EDPN are notably shaper and clearer while other methods are only able to address small blur.

\subsection{Challenge Results}
In the NTIRE 2021 Image Deblurring Challenge, EDPN achieves the best PSNR/SSIM/LPIPS scores in Track 1 (Low Resolution) and the best SSIM/LPIPS scores in Track 2 (JPEG Artifacts).
We list the results from the top 10 teams on the final test set in Table~\ref{tab:reds}.
Compared with the second best method, our EDPN achieves 0.13 dB increase in PSNR, 0.017 increase in SSIM and 0.0172 decrease in LILPS in Track 1.
In Track 2, our EDPN achieves 0.0162 increase in SSIM and 0.008 decrease in LILPS.
According to the challenge report~\cite{Nah_2021_CVPR_Workshops_Deblur}, our EDPN is among the most efficient solutions in both tracks.

\vspace{-0.1cm}
\section{Conclusion}
\vspace{-0.1cm}
In this paper, we propose a blurry image restoration network named EDPN, which is designed to address multiple degradations, \eg, blurry image super-resolution and blurry image deblocking.
The proposed two core modules of EDPN, the PPT module and the PSA module, are proved to be highly effective for the above two tasks.
We believe EDPN also has potential to advance other image/video restoration tasks, especially those with multiple degradations.

\vspace{-0.1cm}
\section*{Acknowledgement}
\vspace{-0.1cm}
  We acknowledge funding from National Key R\&D Program of China under Grant 2017YFA0700800, and National Natural Science Foundation of China under Grant 61901435.


{\small
\bibliographystyle{ieee}
\bibliography{egbib}

\begin{thebibliography}{10}\itemsep=-1pt

\bibitem{abuolaim2020defocus}
Abdullah Abuolaim and Michael~S Brown.
\newblock Defocus deblurring using dual-pixel data.
\newblock In {\em ECCV}, 2020.

\bibitem{abuolaim2021NTIRE}
Abdullah Abuolaim, Michael~S Brown, and Radu Timofte.
\newblock Ntire 2021 challenge for defocus deblurring using dual-pixel images:
  Methods and results.
\newblock In {\em CVPRW}, 2021.

\bibitem{caballero2017real}
Jose Caballero, Christian Ledig, Andrew Aitken, Alejandro Acosta, Johannes
  Totz, Zehan Wang, and Wenzhe Shi.
\newblock Real-time video super-resolution with spatio-temporal networks and
  motion compensation.
\newblock In {\em CVPR}, 2017.

\bibitem{chakrabarti2016neural}
Ayan Chakrabarti.
\newblock A neural approach to blind motion deblurring.
\newblock In {\em ECCV}, 2016.

\bibitem{chang2013reducing}
Huibin Chang, Michael~K Ng, and Tieyong Zeng.
\newblock Reducing artifacts in jpeg decompression via a learned dictionary.
\newblock {\em IEEE Transactions on Image Processing}, 62(3):718--728, 2013.

\bibitem{chen2019camera}
Chang Chen, Zhiwei Xiong, Xinmei Tian, Zheng-Jun Zha, and Feng Wu.
\newblock Camera lens super-resolution.
\newblock In {\em CVPR}, 2019.

\bibitem{dai2017deformable}
Jifeng Dai, Haozhi Qi, Yuwen Xiong, Yi Li, Guodong Zhang, Han Hu, and Yichen
  Wei.
\newblock Deformable convolutional networks.
\newblock In {\em ICCV}, 2017.

\bibitem{dai2019second}
Tao Dai, Jianrui Cai, Yongbing Zhang, Shu-Tao Xia, and Lei Zhang.
\newblock Second-order attention network for single image super-resolution.
\newblock In {\em CVPR}, 2019.

\bibitem{dai2021attentional}
Yimian Dai, Fabian Gieseke, Stefan Oehmcke, Yiquan Wu, and Kobus Barnard.
\newblock Attentional feature fusion.
\newblock In {\em WACV}, 2021.

\bibitem{dong2015compression}
Chao Dong, Yubin Deng, Chen~Change Loy, and Xiaoou Tang.
\newblock Compression artifacts reduction by a deep convolutional network.
\newblock In {\em ICCV}, 2015.

\bibitem{dong2014learning}
Chao Dong, Chen~Change Loy, Kaiming He, and Xiaoou Tang.
\newblock Learning a deep convolutional network for image super-resolution.
\newblock In {\em ECCV}, 2014.

\bibitem{dong2018learning}
Jiangxin Dong, Jinshan Pan, Deqing Sun, Zhixun Su, and Ming-Hsuan Yang.
\newblock Learning data terms for non-blind deblurring.
\newblock In {\em ECCV}, 2018.

\bibitem{fan2018decouple}
Qingnan Fan, Dongdong Chen, Lu Yuan, Gang Hua, Nenghai Yu, and Baoquan Chen.
\newblock Decouple learning for parameterized image operators.
\newblock In {\em ECCV}, 2018.

\bibitem{fergus2006removing}
Rob Fergus, Barun Singh, Aaron Hertzmann, Sam~T Roweis, and William~T Freeman.
\newblock Removing camera shake from a single photograph.
\newblock In {\em SIGGRAPH}. 2006.

\bibitem{foi2007pointwise}
Alessandro Foi, Vladimir Katkovnik, and Karen Egiazarian.
\newblock Pointwise shape-adaptive dct for high-quality denoising and
  deblocking of grayscale and color images.
\newblock {\em IEEE Transactions on Image Processing}, 16(5):1395--1411, 2007.

\bibitem{fu2019jpeg}
Xueyang Fu, Zheng-Jun Zha, Feng Wu, Xinghao Ding, and John Paisley.
\newblock Jpeg artifacts reduction via deep convolutional sparse coding.
\newblock In {\em ICCV}, 2019.

\bibitem{gao2019dynamic}
Hongyun Gao, Xin Tao, Xiaoyong Shen, and Jiaya Jia.
\newblock Dynamic scene deblurring with parameter selective sharing and nested
  skip connections.
\newblock In {\em CVPR}, 2019.

\bibitem{gu2015convolutional}
Shuhang Gu, Wangmeng Zuo, Qi Xie, Deyu Meng, Xiangchu Feng, and Lei Zhang.
\newblock Convolutional sparse coding for image super-resolution.
\newblock In {\em ICCV}, 2015.

\bibitem{guo2016building}
Jun Guo and Hongyang Chao.
\newblock Building dual-domain representations for compression artifacts
  reduction.
\newblock In {\em ECCV}, 2016.

\bibitem{guo2017one}
Jun Guo and Hongyang Chao.
\newblock One-to-many network for visually pleasing compression artifacts
  reduction.
\newblock In {\em CVPR}, 2017.

\bibitem{hore2010image}
Alain Hore and Djemel Ziou.
\newblock Image quality metrics: Psnr vs. ssim.
\newblock In {\em ICPR}, 2010.

\bibitem{hu2018squeeze}
Jie Hu, Li Shen, and Gang Sun.
\newblock Squeeze-and-excitation networks.
\newblock In {\em CVPR}, 2018.

\bibitem{jancsary2012loss}
Jeremy Jancsary, Sebastian Nowozin, and Carsten Rother.
\newblock Loss-specific training of non-parametric image restoration models: A
  new state of the art.
\newblock In {\em ECCV}, 2012.

\bibitem{kappeler2016video}
Armin Kappeler, Seunghwan Yoo, Qiqin Dai, and Aggelos~K Katsaggelos.
\newblock Video super-resolution with convolutional neural networks.
\newblock {\em IEEE Transactions on Computational Imaging}, 2(2):109--122,
  2016.

\bibitem{kim2016accurate}
Jiwon Kim, Jung~Kwon Lee, and Kyoung~Mu Lee.
\newblock Accurate image super-resolution using very deep convolutional
  networks.
\newblock In {\em CVPR}, 2016.

\bibitem{krishnan2009fast}
Dilip Krishnan and Rob Fergus.
\newblock Fast image deconvolution using hyper-laplacian priors.
\newblock In {\em NeurlPS}, 2009.

\bibitem{krishnan2011blind}
Dilip Krishnan, Terence Tay, and Rob Fergus.
\newblock Blind deconvolution using a normalized sparsity measure.
\newblock In {\em CVPR}, 2011.

\bibitem{kupyn2018deblurgan}
Orest Kupyn, Volodymyr Budzan, Mykola Mykhailych, Dmytro Mishkin, and
  Ji{\v{r}}{\'\i} Matas.
\newblock Deblurgan: Blind motion deblurring using conditional adversarial
  networks.
\newblock In {\em CVPR}, 2018.

\bibitem{kupyn2019deblurgan}
Orest Kupyn, Tetiana Martyniuk, Junru Wu, and Zhangyang Wang.
\newblock Deblurgan-v2: Deblurring (orders-of-magnitude) faster and better.
\newblock In {\em ICCV}, 2019.

\bibitem{lee2019deep}
Junyong Lee, Sungkil Lee, Sunghyun Cho, and Seungyong Lee.
\newblock Deep defocus map estimation using domain adaptation.
\newblock In {\em CVPR}, 2019.

\bibitem{li2018multi}
Juncheng Li, Faming Fang, Kangfu Mei, and Guixu Zhang.
\newblock Multi-scale residual network for image super-resolution.
\newblock In {\em ECCV}, 2018.

\bibitem{li2017iterative}
Tao Li, Xiaohai He, Linbo Qing, Qizhi Teng, and Honggang Chen.
\newblock An iterative framework of cascaded deblocking and superresolution for
  compressed images.
\newblock {\em IEEE Transactions on Multimedia}, 20(6):1305--1320, 2017.

\bibitem{lim2017enhanced}
Bee Lim, Sanghyun Son, Heewon Kim, Seungjun Nah, and Kyoung Mu~Lee.
\newblock Enhanced deep residual networks for single image super-resolution.
\newblock In {\em CVPRW}, 2017.

\bibitem{nah2019ntire}
Seungjun Nah, Sungyong Baik, Seokil Hong, Gyeongsik Moon, Sanghyun Son, Radu
  Timofte, and Kyoung Mu~Lee.
\newblock Ntire 2019 challenge on video deblurring and super-resolution:
  Dataset and study.
\newblock In {\em CVPRW}, 2019.

\bibitem{nah2017deep}
Seungjun Nah, Tae Hyun~Kim, and Kyoung Mu~Lee.
\newblock Deep multi-scale convolutional neural network for dynamic scene
  deblurring.
\newblock In {\em CVPR}, 2017.

\bibitem{Nah_2021_CVPR_Workshops_Deblur}
Seungjun Nah, Sanghyun Son, Suyoung Lee, Radu Timofte, and Kyoung~Mu Lee.
\newblock Ntire 2021 challenge on image deblurring.
\newblock In {\em CVPRW}, 2021.

\bibitem{pan2016blind}
Jinshan Pan, Deqing Sun, Hanspeter Pfister, and Ming-Hsuan Yang.
\newblock Blind image deblurring using dark channel prior.
\newblock In {\em CVPR}, 2016.

\bibitem{pan2019single}
Liyuan Pan, Yuchao Dai, and Miaomiao Liu.
\newblock Single image deblurring and camera motion estimation with depth map.
\newblock In {\em WACV}, 2019.

\bibitem{quan2020collaborative}
Yuhui Quan, Jieting Yang, Yixin Chen, Yong Xu, and Hui Ji.
\newblock Collaborative deep learning for super-resolving blurry text images.
\newblock {\em IEEE Transactions on Computational Imaging}, 6:778--790, 2020.

\bibitem{ramakrishnan2017deep}
Sainandan Ramakrishnan, Shubham Pachori, Aalok Gangopadhyay, and Shanmuganathan
  Raman.
\newblock Deep generative filter for motion deblurring.
\newblock In {\em ICCVW}, 2017.

\bibitem{richardson1972bayesian}
William~Hadley Richardson.
\newblock Bayesian-based iterative method of image restoration.
\newblock {\em JoSA}, 62(1):55--59, 1972.

\bibitem{schmidt2014shrinkage}
Uwe Schmidt and Stefan Roth.
\newblock Shrinkage fields for effective image restoration.
\newblock In {\em CVPR}, 2014.

\bibitem{shan2008high}
Qi Shan, Jiaya Jia, and Aseem Agarwala.
\newblock High-quality motion deblurring from a single image.
\newblock {\em Acm Transactions on Graphics (TOG)}, 27(3):1--10, 2008.

\bibitem{Son_2021_CVPR_Workshops_VSR}
Sanghyun Son, Suyoung Lee, Seungjun Nah, Radu Timofte, and Kyoung~Mu Lee.
\newblock Ntire 2021 challenge on video super-resolution.
\newblock In {\em CVPRW}, 2021.

\bibitem{su2017deep}
Shuochen Su, Mauricio Delbracio, Jue Wang, Guillermo Sapiro, Wolfgang Heidrich,
  and Oliver Wang.
\newblock Deep video deblurring for hand-held cameras.
\newblock In {\em CVPR}, 2017.

\bibitem{sun2015learning}
Jian Sun, Wenfei Cao, Zongben Xu, and Jean Ponce.
\newblock Learning a convolutional neural network for non-uniform motion blur
  removal.
\newblock In {\em CVPR}, 2015.

\bibitem{svoboda2016compression}
Pavel Svoboda, Michal Hradis, David Barina, and Pavel Zemcik.
\newblock Compression artifacts removal using convolutional neural networks.
\newblock In {\em WSCG}, 2016.

\bibitem{tao2017detail}
Xin Tao, Hongyun Gao, Renjie Liao, Jue Wang, and Jiaya Jia.
\newblock Detail-revealing deep video super-resolution.
\newblock In {\em ICCV}, 2017.

\bibitem{tao2018scale}
Xin Tao, Hongyun Gao, Xiaoyong Shen, Jue Wang, and Jiaya Jia.
\newblock Scale-recurrent network for deep image deblurring.
\newblock In {\em CVPR}, 2018.

\bibitem{tian2020tdan}
Yapeng Tian, Yulun Zhang, Yun Fu, and Chenliang Xu.
\newblock Tdan: Temporally-deformable alignment network for video
  super-resolution.
\newblock In {\em CVPR}, 2020.

\bibitem{vaswani2017attention}
Ashish Vaswani, Noam Shazeer, Niki Parmar, Jakob Uszkoreit, Llion Jones,
  Aidan~N Gomez, Lukasz Kaiser, and Illia Polosukhin.
\newblock Attention is all you need.
\newblock In {\em NeurlPS}, 2017.

\bibitem{wang2020jpeg}
Menglu Wang, Xueyang Fu, Zepei Sun, and Zheng-Jun Zha.
\newblock Jpeg artifacts removal via compression quality ranker-guided
  networks.
\newblock In {\em IJCAI}, 2020.

\bibitem{wang2019edvr}
Xintao Wang, Kelvin~C.K. Chan, Ke Yu, Chao Dong, and Chen~Change Loy.
\newblock Edvr: Video restoration with enhanced deformable convolutional
  networks.
\newblock In {\em CVPRW}, 2019.

\bibitem{wang2018non}
Xiaolong Wang, Ross Girshick, Abhinav Gupta, and Kaiming He.
\newblock Non-local neural networks.
\newblock In {\em CVPR}, 2018.

\bibitem{weiner1949smoothing}
N Weiner and Interpolation Extrapolation.
\newblock Smoothing of stationary time series: With engineering applications,
  1949.

\bibitem{wieschollek2017learning}
Patrick Wieschollek, Michael Hirsch, Bernhard Scholkopf, and Hendrik Lensch.
\newblock Learning blind motion deblurring.
\newblock In {\em ICCV}, 2017.

\bibitem{xiao2021spacetime}
Zeyu Xiao, Xueyang Fu, Jie Huang, Zhen Cheng, and Zhiwei Xiong.
\newblock Space-time distillation for video super-resolution.
\newblock In {\em CVPR}, 2021.

\bibitem{xiao2020spacetime}
Zeyu Xiao, Zhiwei Xiong, Xueyang Fu, Dong Liu, and Zheng-Jun Zha.
\newblock Space-time video super-resolution using temporal profiles.
\newblock In {\em ACM MM}, 2020.

\bibitem{xiong2009image}
Zhiwei Xiong, Xiaoyan Sun, and Feng Wu.
\newblock Image hallucination with feature enhancement.
\newblock In {\em CVPR}, 2009.

\bibitem{xiong2013example}
Zhiwei Xiong, Dong Xu, Xiaoyan Sun, and Feng Wu.
\newblock Example-based super-resolution with soft information and decision.
\newblock {\em IEEE Transactions on Multimedia}, 15(6):1458--1465, 2013.

\bibitem{xu2010two}
Li Xu and Jiaya Jia.
\newblock Two-phase kernel estimation for robust motion deblurring.
\newblock In {\em ECCV}, 2010.

\bibitem{xu2013unnatural}
Li Xu, Shicheng Zheng, and Jiaya Jia.
\newblock Unnatural l0 sparse representation for natural image deblurring.
\newblock In {\em CVPR}, 2013.

\bibitem{xu2017learning}
Xiangyu Xu, Deqing Sun, Jinshan Pan, Yujin Zhang, Hanspeter Pfister, and
  Ming-Hsuan Yang.
\newblock Learning to super-resolve blurry face and text images.
\newblock In {\em ICCV}, 2017.

\bibitem{xue2019video}
Tianfan Xue, Baian Chen, Jiajun Wu, Donglai Wei, and William~T Freeman.
\newblock Video enhancement with task-oriented flow.
\newblock {\em International Journal of Computer Vision}, 127(8):1106--1125,
  2019.

\bibitem{yang2020deblurring}
Chao-Hsun Yang and Long-Wen Chang.
\newblock Deblurring and super-resolution using deep gated fusion attention
  networks for face images.
\newblock In {\em ICASSP}, 2020.

\bibitem{yoo2018image}
Jaeyoung Yoo, Sang-ho Lee, and Nojun Kwak.
\newblock Image restoration by estimating frequency distribution of local
  patches.
\newblock In {\em CVPR}, 2018.

\bibitem{yu2018super}
Xin Yu, Basura Fernando, Richard Hartley, and Fatih Porikli.
\newblock Super-resolving very low-resolution face images with supplementary
  attributes.
\newblock In {\em CVPR}, 2018.

\bibitem{zamir2021multi}
Syed~Waqas Zamir, Aditya Arora, Salman Khan, Munawar Hayat, Fahad~Shahbaz Khan,
  Ming-Hsuan Yang, and Ling Shao.
\newblock Multi-stage progressive image restoration.
\newblock In {\em CVPR}, 2021.

\bibitem{zhang2020joint}
Dongyang Zhang, Zhenwen Liang, and Jie Shao.
\newblock Joint image deblurring and super-resolution with attention dual
  supervised network.
\newblock {\em Neurocomputing}, 412:187--196, 2020.

\bibitem{zhang2018dynamic}
Jiawei Zhang, Jinshan Pan, Jimmy Ren, Yibing Song, Linchao Bao, Rynson~WH Lau,
  and Ming-Hsuan Yang.
\newblock Dynamic scene deblurring using spatially variant recurrent neural
  networks.
\newblock In {\em CVPR}, 2018.

\bibitem{zhang2006edge}
Lei Zhang and Xiaolin Wu.
\newblock An edge-guided image interpolation algorithm via directional
  filtering and data fusion.
\newblock {\em IEEE Transactions on Image Processing}, 15(8):2226--2238, 2006.

\bibitem{zhang2018gated}
Xinyi Zhang, Hang Dong, Zhe Hu, Wei-Sheng Lai, Fei Wang, and Ming-Hsuan Yang.
\newblock Gated fusion network for joint image deblurring and super-resolution.
\newblock In {\em BMVC}, 2018.

\bibitem{zhang2018deep}
Xinyi Zhang, Fei Wang, Hang Dong, and Yu Guo.
\newblock A deep encoder-decoder networks for joint deblurring and
  super-resolution.
\newblock In {\em ICASSP}, 2018.

\bibitem{zhang2018dmcnn}
Xiaoshuai Zhang, Wenhan Yang, Yueyu Hu, and Jiaying Liu.
\newblock Dmcnn: Dual-domain multi-scale convolutional neural network for
  compression artifacts removal.
\newblock In {\em ICIP}, 2018.

\bibitem{zhang2018image}
Yulun Zhang, Kunpeng Li, Kai Li, Lichen Wang, Bineng Zhong, and Yun Fu.
\newblock Image super-resolution using very deep residual channel attention
  networks.
\newblock In {\em ECCV}, 2018.

\bibitem{zhang2019residual}
Yulun Zhang, Kunpeng Li, Kai Li, Bineng Zhong, and Yun Fu.
\newblock Residual non-local attention networks for image restoration.
\newblock In {\em ICLR}, 2019.

\bibitem{zoran2011learning}
Daniel Zoran and Yair Weiss.
\newblock From learning models of natural image patches to whole image
  restoration.
\newblock In {\em ICCV}, 2011.

\end{thebibliography}
}

\end{document}